\documentclass[11pt]{article}

\usepackage[letterpaper, margin=0.98in, top=1in, bottom=1in]{geometry}

\usepackage{microtype}
\usepackage{graphicx}
\usepackage{subfigure}
\usepackage{booktabs} 
\usepackage{hyperref}

\newcommand{\removed}[1]{}
\usepackage[textsize=tiny]{todonotes}

\newcommand{\marko}[1]{{\color{olive} [Marko: #1]}}

\newcommand{\zhiyuan}[1]{{\color{red} [ZL:#1]}}
\newcommand{\kaifeng}[1]{{\color{orange} [KL:#1]}}

\newcommand{\SL}{{\ensuremath{\mathcal{L}_{\mathrm{ST}}}}}
\newcommand{\SLceil}{{\ensuremath{\mathcal{L}_{\mathrm{ST}}^{\mathrm{ceil}}}}}
\newcommand{\TL}{\ensuremath{\mathcal{L}_{\mathrm{TE}}}}
\newcommand{\TU}{\ensuremath{\mU_{\mathrm{TE}}}}

\newcommand{\MS}{\ensuremath{M_{\mathrm{ST}}}}
\newcommand{\MT}{\ensuremath{M_{\mathrm{TE}}}}
\newcommand{\PGR}{\ensuremath{\mathrm{PGR}}}

\newcommand{\email}[1]{\thanks{Email: \texttt{#1}}}

\usepackage{amsmath}
\usepackage{amssymb}
\usepackage{mathtools}
\usepackage{amsthm}
\usepackage[capitalize,noabbrev]{cleveref}
\usepackage{bbm}
\usepackage{natbib}
\usepackage{comment}

\usepackage{aliascnt}
\usepackage[capitalize,noabbrev]{cleveref} 

\theoremstyle{plain}
\newtheorem{theorem}{Theorem}[section]

\newaliascnt{lemma}{theorem}
\newtheorem{lemma}[lemma]{Lemma}
\aliascntresetthe{lemma}
\crefname{lemma}{lemma}{lemmas}
\Crefname{lemma}{Lemma}{Lemmas}

\newaliascnt{proposition}{theorem}
\newtheorem{proposition}[proposition]{Proposition}
\aliascntresetthe{proposition}
\crefname{proposition}{proposition}{propositions}
\Crefname{proposition}{Proposition}{Propositions}

\newaliascnt{corollary}{theorem}
\newtheorem{corollary}[corollary]{Corollary}
\aliascntresetthe{corollary}
\crefname{corollary}{corollary}{corollaries}
\Crefname{corollary}{Corollary}{Corollaries}

\theoremstyle{definition}

\newaliascnt{definition}{theorem}
\newtheorem{definition}[definition]{Definition}
\aliascntresetthe{definition}
\crefname{definition}{definition}{definitions}
\Crefname{definition}{Definition}{Definitions}

\newaliascnt{assumption}{theorem}
\newtheorem{assumption}[assumption]{Condition}
\aliascntresetthe{assumption}
\crefname{assumption}{Condition}{Conditions}
\Crefname{assumption}{Condition}{Conditions}

\newaliascnt{condition}{theorem}

\aliascntresetthe{condition}
\crefname{condition}{Condition}{Conditions}
\Crefname{condition}{Condition}{Conditions}

\newaliascnt{example}{theorem}

\aliascntresetthe{example}
\crefname{example}{example}{examples}
\Crefname{example}{Example}{Examples}

\newaliascnt{notation}{theorem}

\aliascntresetthe{notation}
\crefname{notation}{notation}{notations}
\Crefname{notation}{Notation}{Notations}

\newaliascnt{model}{theorem}
\newtheorem{model}[model]{Model}
\aliascntresetthe{model}
\crefname{model}{model}{models}
\Crefname{model}{Model}{Models}

\theoremstyle{remark}
\newaliascnt{remark}{theorem}
\newtheorem{remark}[remark]{Remark}
\aliascntresetthe{remark}
\crefname{remark}{remark}{remarks}
\Crefname{remark}{Remark}{Remarks}

\usepackage{enumitem}
\setlist[enumerate,1]{leftmargin=0.6cm}
\setlist[itemize,1]{leftmargin=0.4cm}
\usepackage{amsmath,amsfonts,bm}

\def\1{\bm{1}}

\def\eps{{\epsilon}}

\def\vzero{{\bm{0}}}

\def\vphi{{\bm{\phi}}}

\def\va{{\bm{a}}}
\def\vb{{\bm{b}}}

\makeatletter
\newcommand{\ve}{\@ifnextchar\bgroup{\velong}{{\bm{e}}}}
\newcommand{\velong}[1]{{\bm{#1}}}
\makeatother

\def\vf{{\bm{f}}}
\def\vg{{\bm{g}}}

\def\vt{{\bm{t}}}
\def\vu{{\bm{u}}}
\def\vv{{\bm{v}}}
\def\vw{{\bm{w}}}
\def\vx{{\bm{x}}}
\def\vy{{\bm{y}}}
\def\vz{{\bm{z}}}

\def\mA{{\bm{A}}}
\def\mB{{\bm{B}}}

\def\mD{{\bm{D}}}

\def\mG{{\bm{G}}}
\def\mH{{\bm{H}}}
\def\mI{{\bm{I}}}

\def\mS{{\bm{S}}}

\def\mU{{\bm{U}}}
\def\mV{{\bm{V}}}

\def\mX{{\bm{X}}}
\def\mY{{\bm{Y}}}

\def\mPhi{{\bm{\Phi}}}

\DeclareMathAlphabet{\mathsfit}{\encodingdefault}{\sfdefault}{m}{sl}
\SetMathAlphabet{\mathsfit}{bold}{\encodingdefault}{\sfdefault}{bx}{n}

\def\gD{{\mathcal{D}}}

\def\gH{{\mathcal{H}}}

\def\gK{{\mathcal{K}}}
\def\gL{{\mathcal{L}}}

\newcommand{\E}{\mathbb{E}}
\newcommand{\R}{\mathbb{R}}

\DeclareMathOperator*{\argmin}{arg\,min}

\DeclareMathOperator{\sign}{sign}

\newcommand{\cD}{\mathcal{D}}

\newcommand{\cG}{\mathcal{G}}
\newcommand{\cH}{\mathcal{H}}

\newcommand{\cK}{\mathcal{K}}
\newcommand{\cL}{\mathcal{L}}

\newcommand{\cT}{\mathcal{T}}
\newcommand{\cU}{\mathcal{U}}

\newcommand{\cX}{\mathcal{X}}
\newcommand{\cY}{\mathcal{Y}}

\newcommand{\sphS}{\mathbb{S}}

\newcommand{\vbeta}{{\bm \beta}}

\newcommand{\N}{\mathbb{N}}

\newcommand{\norm}[1]{\|#1\|}
\newcommand{\lnorm}[1]{\left\|#1\right\|}
\newcommand{\abs}[1]{\lvert #1 \rvert}

\newcommand{\vspan}{\mathrm{span}}
\newcommand{\colspan}{\mathrm{colspan}}
\newcommand{\range}{\mathrm{range}}
\newcommand{\inner}[2]{\left\langle{#1}, {#2}\right\rangle}
\newcommand{\inne}[2]{\langle{#1}, {#2}\rangle}
\newcommand{\rank}{\mathrm{rank}}

\newcommand{\poly}{\mathrm{poly}}

\newcommand{\dd}{\mathrm{d}}

\makeatletter
\renewcommand{\paragraph}{%
  \@startsection{paragraph}{4}%
  {\z@}{0ex}{-1em}%
  {\normalfont\normalsize\bfseries}%
}
\makeatother

\Crefname{assumption}{Condition}{Conditions}

\title{Weak-to-Strong Generalization Even in Random Feature Networks, Provably}
\usepackage{authblk}
\makeatletter
\renewcommand\AB@affilsepx{\qquad\protect\Affilfont}
\makeatother

\renewcommand{\thefootnote}{\fnsymbol{footnote}}

\author[1,*,\email{\{last name\}@uchicago.edu}]{Marko Medvedev}
\author[2,*]{Kaifeng Lyu}
\author[3]{Dingli Yu}
\author[4]{Sanjeev Arora}
\author[5]{Zhiyuan Li}
\author[5]{Nathan Srebro}
\affil[1]{University of Chicago}
\affil[2]{Simons Institute, UC Berkeley}
\affil[3]{Microsoft Research}
\affil[4]{Princeton University}
\affil[5]{Toyota Technological Institute at Chicago}

\date{}

\begin{document}
\footnotetext[1]{Equal Contribution}
\maketitle

\renewcommand{\thefootnote}{\arabic{footnote}}

\begin{abstract}
Weak-to-Strong Generalization \citep{burns2024weaktostrong} is the phenomenon whereby a strong student, say GPT-4, learns a task from a weak teacher, say GPT-2, and ends up significantly outperforming the teacher. We show that this phenomenon does not require a complex and pre-trained learner like GPT-4, and can arise even in simple non-pretrained models, simply due to the size advantage of the student. But, we also show that there are inherent limits to the extent of such weak-to-strong generalization. 

We consider students and teachers that are random feature models, described by two-layer networks with a random and fixed bottom layer and a trained top layer. A ‘weak’ teacher, with a small number of units (i.e.~random features), is trained on the population, and a ‘strong’ student, with a much larger number of units (i.e.~random features), is trained only on labels generated by the weak teacher. We demonstrate, prove, and understand how the student can outperform the teacher, even though trained only on data labeled by the teacher. We also explain how such weak-to-strong generalization is enabled by early stopping.  

We then show the quantitative limits of weak-to-strong generalization in this model, and in fact in a much broader class of models, for arbitrary teacher and student feature spaces and a broad class of learning rules, including when the student features are pre-trained or otherwise more informative. In particular, we show that in such models the student's error can only approach zero if the teacher's error approaches zero, and a strong student cannot ``boost'' a slightly-better-than-chance teacher to obtain a small error.

\removed{\\
We discuss implications for {\em super-alignment and scalable oversight}, which are the intended key applications of W2SG.}
\end{abstract}

\section{Introduction}

\begin{figure}
    \centering
    \includegraphics[width=1.\linewidth]{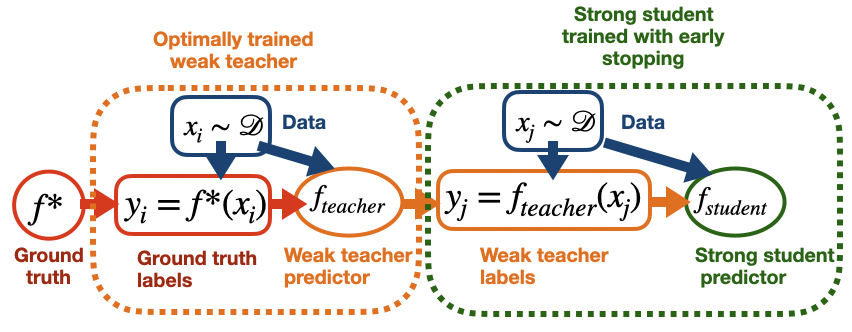}
    \caption{Illustration of our setup for weak-to-strong generalization. The teacher model of smaller size is trained on ground truth labels, while the student model of larger size is trained with early stopping on labels produced by the teacher model. }
    \label{fig:w2s-sketch}
\end{figure}

\cite{sauel59studies}, in the first ever paper introducing the term ``machine learning'', emphasized how his learned system outperformed its human teacher.  \removed{More broadly, the idea of a good student outperforming their teacher is much older, as epitomized, e.g.~by a quote widely attributed to Leonardo de Vinci.\footnote{``Poor is the pupil who does not surpass his master'', though this quote does not appear in de Vinci's surviving writing and its source is unclear to us.}}More broadly, the idea of a good student outperforming their teacher is much older, as epitomized by the quote ``{\em Poor is the pupil who does not surpass his master}'', widely attributed to Leonardo de Vinci.\footnote{The quote does not appear in de Vinci's surviving writing and its source is unclear to us} Recently, \citet{burns2024weaktostrong} discussed the concept in the context of ``superalignment'' (weak humans controlling strong machines), and suggested studying (strong) student machines outperforming their (weaker) machine teachers as a surrogate model.  \citeauthor{burns2024weaktostrong} demonstrated the phenomenon by fine-tuning a ``weak'' teacher model, namely a pre-trained GPT-2\removed{~\citep{radford2019language}}, on task-specific data with true labels, and then using this ``weak'' teacher model to annotate data for the same task, and fine-tuning a ``strong'' student model, namely a pre-trained GPT-4\removed{~\citep{achiam2023gpt}}, only using labels generated by the ``weak'' teacher (see Figure \ref{fig:w2s-sketch}).  In these experiments, the ``strong'' students indeed outperformed their ``weak'' teachers, and \citeauthor{burns2024weaktostrong} asked for an understanding of this {\em weak-to-strong generalization} phenomenon.

There are several mechanisms by which a student can outperform their teacher.  A student can do a more thorough test-time search, e.g.~looking ahead more moves or thinking of more ways to solve a problem.  Taking a step further, on problems where such a search can be helpful, a student can also learn to internalize the search, learning a better value function or next-action (or token) predictor through self-play or training-time search (as in \citeauthor{sauel59studies}'s checker learner, and modern RL-based training methods for LLMs).  A student might also be able to correct a teacher by having better inductive bias, perhaps obtained through pre-training on similar or related tasks.  For example, a Chinese-speaking student can outperform their non-Chinese-speaking teacher when being taught to pronounce the names of Chinese colleagues.  

But can a student outperform their teacher simply by being ``bigger'', having more units and a larger representation, without more specialized or specific inductive bias, nor more or better pre-training?  Does this require sophisticated models with emergent capabilities, such as GPT-4-scale deep transformers, or can it also arise in much simpler models? If the model sizes are further scaled up, will weak-to-strong generalization be enhanced or diminished?

\removed{Without understanding its underlying principles, relying on weak-to-strong generalization for superalignment may be risky.}

In the first part of this paper, we analyze and prove how weak-to-strong generalization (i.e.~a student outperforming their student) can happen even in models as simple as ``random feature models'', i.e., two-layer networks with a random bottom layer and only the top layer trained.  More specifically, we study the setup where the teacher and student models are two-layer networks with $\MT$ and $\MS$ (hidden, random) units, respectively. The student model is significantly stronger than the teacher, in the sense that it has many more units than the teacher, $\MS \gg \MT$.
In both models, we sample the bottom layer weights from the same spherical distribution, and then train only the top layer weights. As in~\citet{burns2024weaktostrong} and \Cref{fig:w2s-sketch}, the teacher model is trained with true labels, and the student model is trained only with labels generated by the teacher.  We further assume that the teacher model is trained so well that it attains the minimum possible population loss. Thus, the only remaining source of error is the approximation error due to the finite number of units.  

We consider two different random feature models, corresponding to two-layer networks with different activation functions and input distributions: (a) ReLU activations with an isotropic input distribution; and (b) linear activations with a simple anisotropic input distribution.  Denoting the teacher and student predictive mean squared errors by $\TL$ and $\SL$ respectively (normalized such that the null loss is $1$), in both cases we show that {\bf the student can infinitely outperform its teacher, in the sense that the ratio $\SL/\TL$ can be arbitrarily small.}  Thus, {\bf the {\em Performance Gap Recovered (PGR)}} introduced by \citet[][see definition in Section \ref{sec:setup} \Cref{eq:pgr}]{burns2024weaktostrong} {\bf can be arbitrarily close to $1$}.  Furthermore, we show an even stronger gap, in the sense that the student learns polynomially better: {\bf for ReLU networks we show that we can have $\SL=\tilde{O}\left( \TL^{1.49} \right)$} (\Cref{thrm:main:2layerrelu}, relying on the Gassian Universality Ansatz), and for {\bf linear networks with anisotropic inputs we show that we can have $\SL = \tilde{O}\left( \TL^2 \right)$} (\Cref{thrm:main:diagfeatcov}).

As in~\citet{burns2024weaktostrong}, we use {\em early stopping} when training the student---it is easy to see how in our setting, early stopping is essential: Since the student is more expressive than the teacher, without early stopping, they would just replicate the teacher's mistakes.  In Section \ref{sec:howdoesw2shappen} we explain how such early stopping allows the weak-to-strong generalization described above, shedding light on its empirically observed role.

Complementing these results, we also study the limits of weak-to-strong generalization.  In \Cref{subsec:lowerbound} we show that the quadratic gap described above is the largest possible, and in any random feature model we would always have $\SL = \Omega(\TL^2)$.  In fact, this limit extends well beyond random feature models, to models with arbitrary teacher and student feature spaces (even if the student features are highly specialized or pre-trained), with early stopped gradient descent (starting from zero initialization) or with other convex regularizers.  In particular, this limit implies that in such settings, we can have vanishing student error $\SL\rightarrow 0$ only when the teacher error also vanishes $\TL\rightarrow 0$, and that if the teacher error is only slightly better than null, the student error will also be close to null---we cannot ``boost'' a teacher with a tiny edge to a near-perfect student.

\paragraph{Asymptotic Notation.} We use $\Theta_c$, $\tilde{O}_c$, and $\Omega_c$, where $c$ is one or more variables, to indicate a multiplicative factor that depends on $c$.  E.g., for any two quantities $\textit{exp}_1,\textit{exp}_2$, that could depend on various variables including $d$ and $k$, writing $\textit{exp}_1 = \tilde{O}_{d,k}(\textit{exp}_2)$ means that there exists some function $f(d,k)$ and exponent $r\geq 0$ s.t.~it always holds that $\textit{exp}_1 \leq f(d,k) \cdot\textit{exp}_2\cdot\log^r (\textit{exp}_2)$.



\section{Two Layer Networks, Random Features, and the Weak-to-Strong Setup}\label{sec:setup}

We consdier learning a predictor $f:\cX\to \cY$, where $\cX=\R^d$ and $\cY=\R$, aiming for small population loss $\cL(f) := \E_{\vx\sim\cD}[ (f(\vx) - f^*(\vx))^2]$, where $\cD$ is some input distribution and $f^*:\cX \rightarrow \cY$ our target function.
Learning is done using two-layer networks with $m$ units of the form
\begin{align}\label{eq:2layernet}
    f_{\vw,\vu}(\vx)=\sum_{i=1}^{m} w_i \sigma(\langle \vu_i,\vx\rangle),
\end{align}
where $\sigma:\R\rightarrow\R$ is some fixed activation function (e.g.~ReLU or linear), the {\bf bottom layer weights $\vu_i\in \R^d$ are randomly drawn}\removed{\footnote{Different distribution $\vu_i\sim \mathcal{U}$ can also be considered, but in our setup we can take the bottom layer weights to be spherical Gaussians.}}
\begin{equation}\label{eq:u}
    \vu_i \overset{\textrm{i.i.d. }}{\sim} \mathcal{U}
\end{equation}from some fixed distribution $\mathcal{U}$ over $\R^d$ (e.g.~spherical Gaussian or similar)
and then \textbf{fixed during training}, and {\bf only the top layer weights $w_i\in \R$ are trained}. Such models, which amount to learning a linear predictor over the {\em random features} defined by the outputs of the hidden units, have been extensively studied as finite approximations of kernel methods~\citep{vempala2006proj,rahimi2007}, as models and limits of deep learning in certain regimes~\citep{mei2021concentration} and recently also for deriving scaling laws and optimal tradeoffs~\citep{lee2024scaling,paquette2024scaling}.

\paragraph{Weak-to-Strong Setup.} 
We train two networks, a `weak' teacher network $f_{\text{teacher}}$ with $\MT$ units, and a `strong' student network $f_{\text{student}}$ with $\MS \gg \MT$ units.  First, we train the teacher network $f_{\text{teacher}}:=f_{\vw_{\mathrm{t}},\vu_{\mathrm{t}}}$ on the true target labels $(x,f^*(x))$ using very large amount of data.  We will model this by training the teacher on population. That is, to train the teacher we draw the bottom layer weights $\{\vu_{\mathrm{t},i}\}_{i=1}^{\MT}$ at random as in \eqref{eq:u}, and for most results in this paper we assume that the top layer weights achieves the population loss minimization~(\Cref{ass:teacher_minimizer}):
\begin{assumption}\label{ass:teacher_minimizer}
    The teacher model attains the minimum loss among all possible models with the given bottom layers:
\begin{equation}\label{eq:teacher_definition} 
\vw_{\mathrm{t}} = \argmin_{\vw\in\R^{\MT}} \E_{\vx\sim \cD}\left[\left(f_{\vw,\vu_{\mathrm{t}}}(\vx)-f^*(\vx)\right)^2 \right].
\end{equation}
\end{assumption}
We denote the teacher's population loss (which is the minimal possible loss using its $\MT$ random units) by
\begin{equation}
\TL := \E_{\vx\sim\cD}\left[ \left(f_{\text{teacher}}(\vx) - f^*(\vx)\right)^2\right].
\end{equation} 

We then train the student on inputs $x \sim \cD$ but only using labels $f_{\text{teacher}}(x)$ generated by the teacher.  To train the student, we draw the bottom layer at random according to \eqref{eq:u}, initialize the top layer weights to zero, $\forall_i \vw_{\mathrm{s},i}(0)=0$, and train these weights using SGD on i.i.d.~samples $\vx(n) \sim \cD$ labeled with by the teacher:
\begin{align}\label{eq:stud-SGD}
\begin{aligned}
w_{\mathrm{s},i}(n+1)
    = w_{\mathrm{s},i}(n) - \eta \cdot \frac{\partial}{\partial w_{\mathrm{s},i}}\!\left(f_{\vw_{\mathrm{s}}(n),\vu_{\mathrm{s}}}(\vx(n)) - f_{\text{teacher}}(\vx(n))\right)^2.\\
    \end{aligned}
\end{align}



\removed{
We denote the resulting network, after $N$ steps of SGD, by $f_{\text{student}} = f_{\vw(N),\vu_S}$.

We assume the student has unrestricted access to draws $x\sim\cD$ from the input distribution and teacher labels $f_{\text{teacher}}(x)$ for these points. Similarly, we will assume that the student has unrestricted access to the teacher, so it is trained on a large amount of teacher-labeled examples using SGD with $N$ steps. Let $f_{\vw_s,\vu_s}^{S}:=f_{\text{student}}$ be the parametrization of the student network:
\begin{align}\label{eq:studentparam}
    f_{\vw_s,\vu_s}^{S}=\sum_{i=1}^{\MS} w_{s,i} \sigma(\langle \vu_{s,i},\vx\rangle)
\end{align}
where $\vw_s\in \R^m,\vu_s\in \R^{d\times m}$ are the parameters of the student. The student is initialized with $\vw_{s,i}(0)=0$ for all $i$ and trained with SGD with stepsize $\eta$ and $N$ steps on $\cD$ where the target labels are generated by the teacher $f_{\text{teacher}}$. At step $n$ we do the following update on the weights $\vw_{s}$
} 

\paragraph{Gradient Flow Limit.} We assume that the student has unrestricted access to the examples labeled by the teacher, and so can take an arbitrarily small learning rate $\eta$ and an appropriately large number of steps $N$. Accordingly, we study the behavior with an infinitesimally small learning rate $\eta$ where the student's dynamics \eqref{eq:stud-SGD} converge to the gradient flow trajectory:
\begin{equation}\label{eq:gradflow}
\begin{aligned}
    \frac{\dd w_{\mathrm{s},i}(t)}{\dd t}
    =-\frac{1}{\MS}\frac{\partial}{\partial w_{\mathrm{s},i}} \E_{\vx \sim \cD}\!\!\left[\left(f_{\vw_{\mathrm{s}}(t),\vu_{\mathrm{s}}}(\vx) - f_{\text{teacher}}(\vx)\right)^2\right]
\end{aligned}
\end{equation}
where we replaced the step number $n$ with training time $t=n\eta\MS$. We add an additional factor of $\MS$ here to ensure the gradient flow has a proper limit when $\MS\to\infty$, which will be useful in the analysis of the rest of the paper. Gradient flow has been shown to approximate the trajectory of gradient descent well, both for 2-layer networks \cite{alnur2019earlystopping} and for deep networks \cite{elkabetz2021optimization}.  
\removed{where we replaced the step number $n$ with training time $t=n\eta \MS$. We add an additional factor of $\MS$ here to ensure the gradient flow has a proper limit when $\MS\to\infty$, which will be useful in the analysis of the rest of the paper. We rescale $f_{\vw_S(t),\vu_s}$ with an additinal $\frac{1}{\sqrt{\MS}}$ as $f_{\vw_S(t),\vu_s} = \frac{1}{\MS}\sum_{i=1}^{\MS} w_{s,i}(t) \sigma(\langle \vu_{s,i} ,\vx \rangle)$. We do this to ensure that there is a proper limit when $\MS\to\infty$, which will be useful for the rest of the paper.}In our analysis, we consider a student trained with the gradient flow dynamics \eqref{eq:gradflow} with stopping time $T$ that needs to be specified and denote the resulting student predictor
\begin{equation}\label{eq:studentparam}
f_{\text{student}} = f_{\vw_{\mathrm{s}}(T),\vu_{\mathrm{s}}}.
\end{equation}

\removed{\footnote{Formally, when $\MS\to \infty$, the parameters of the stduent $(\vw_s(t),\vu_s)$ are replaced with the density over the parameter space $\rho(t)$.}}

\paragraph{Infinitely Large Student.} Furthermore, for simplicity we consider an infinitely large student $\MS \to \infty$ (see formal mathematical definitions in \Cref{sec:general}). All of our infinite-width results can be extended to the finite-width case for sufficiently large width. \removed{Importantly, all of our results can be rewritten to hold for finite $\MS$, as long as it is polynomially large enough.}\removed{ Roughly, all of our results will hold for a finite width $\MS$ student trained with gradient flow with an additional error term of $O\left(\frac{\text{poly}(\MT)}{\MS}\right)$. For simplicity, we will state the results considering $\MS\to \infty$. }

\removed{For a formal statement that shows that our results extend to a finite width student network, see \Cref{app:gradflow}.}
\removed{In \Cref{app:gradflow}, we show that gradient flow approximates the case of a student with finite width $M_{ST}$ trained with SGD with a corresponding number of steps as long as $\MS$ is significantly larger than $\MT$.}
\removed{Importantly, all of our results with translate to the case of a finite width student $\MS$. \marko{reword this to: We note that all our results can be rewritten to hold for finite $\MS$.}} 

\paragraph{Generalization Gap.} We evaluate the student on the ground truth labels (which they do not have access to!), that is 
\begin{equation}\SL := \E_{\vx\sim\cD}[ f_{\text{student}}(\vx) - f^*(\vx))^2].
\end{equation}
We ask how the student error $\SL$ compares with the teacher error $\TL$ and, in particular, whether and how it can be smaller even though it only sees examples labeled by the teacher.

To quantify the extent to which the student can best its teacher,  \citet{burns2024weaktostrong} suggested the \emph{Performance Gap Recovered} (PGR), which is defined as\footnote{This definition in terms of loss is equal to \citeauthor{burns2024weaktostrong}'s definition in terms of accuracy.}:
\begin{align}\label{eq:pgr}
    \PGR = \frac{\TL-\SL}{\TL-\SLceil} \geq \frac{\TL-\SL}{\TL} 
\end{align}
where $\SLceil \geq 0$ is the ``ceiling performance''\citep{burns2024weaktostrong} of the student (the loss the student model could potentially have attained with direct access to the real labels).  We do not carefully define or analyze this ceiling performance, since we provide lower bounds on the right-hand side in \eqref{eq:pgr}, and hence on the $\PGR$ regardless of the ceiling.

\paragraph{Specific Activation and Data Models.} We study two different variants of the random feature model (aka 2-layer network with random and fixed bottom layer) introduced above, which differ in the choice of activation function and input distribution:
\begin{model}[$2$-layer ReLU Network]
\label{ass:relu}
    The activation functions $\sigma(z)=\max(z,0)$ is a standard ReLU function, the bottom layer weights $\vu$ are uniform over the sphere, i.e.~$\mathcal{U}=\text{Unif}(\mathbb S^{d-1})$ in \eqref{eq:u}, and the inputs $\vx$ are also uniformly distributed on the sphere, i.e.~$\cD = \text{Unif}(\mathbb S^{d-1})$.  In this model, we will consider target functions $f^*(x)$ which are sums of linear functions and even polynomials.
\end{model}

\begin{model}[Linear Network]
\label{ass:linearnetwork}
    The activation function $\sigma(z)=z$ is linear (i.e.~this is a linear neural net), and the bottom layer weights $\vu$ are standard Gaussian
    , i.e.~$\mathcal{U}=N(0,\mI_d)$ in \eqref{eq:u}.  This time the inputs $\vx$ are non-isotropic and are drawn from a Gaussian $\cD = N(0,\Psi)$ with diagonal covariance $\Psi = \text{diag}(\psi_1,\psi_2,\dots,\psi_d)$ with $\psi_i$ decreasing. 
      Here, we will consider linear targets supported on the first top few coordinates. 
\end{model}

\Cref{ass:linearnetwork} is fairly general and was studied by e.g.~\citet{lee2024scaling} as a model for studying scaling laws. It can also be thought of as capturing a model where each unit outputs  $\langle \vu,\phi(\vx)\rangle$, where $\phi(\cdot)$ captures what happens in additional lower layers, which creates a non-isotropic representation at the top layer.


\section{Quantifying Weak-to-Strong Generalization}\label{sec:quantifying-w2s}

We are now ready to establish weak-to-strong generalization, starting with the ReLU \ref{ass:relu}:

\begin{theorem}[Weak-to-Strong generalization with $2$-layer ReLU Network]\label{thrm:main:2layerrelu}
    Consider the ReLU \Cref{ass:relu} and the weak-to-strong setup of Section \ref{sec:setup}, for any dimension $d$ and teacher size $\MT$.  Consider a target $f^*$ that is an even polynomial\footnote{Or more generally, a sum of a linear function and an even polynomial.} of degree at most $k$ normalized s.t.~$\E[{f^*}^2]=1$,   with any even degree $k$. If $\MT \ge \Theta(d^{2k})$ and the student is trained with some stopping time $T=\Theta_{d,k}( \log \MT)$, then we have that with probability at least $1- \frac{3}{\MT}$\removed{$-d^k\exp(-c\frac{\MT}{d^k})$},
    \begin{align*}
        \SL\le  O_{d,k}\left( \frac{\log^2 \MT+\log^2 \TL}{\sqrt{\MT}} \, \TL \right),
    \end{align*}
    and so $\SL/\TL\to 0$ and $\PGR\to 1$ as $\MT$ increases.

    Furthermore, under the Gaussian Universality Ansatz\footnote{The Gaussian Universality Ansatz states that when sampling $\vx$, the Gaussian universality holds for the eigenfunctions in the sense that the expected risk remains unchanged if we replace them with Gaussian with appropriate parameters \cite{simon2021}. In our setting, this can be differently stated as the expected risk for \Cref{ass:relu} will remain unchanged if we replace it with an infinitely dimensional \Cref{ass:linearnetwork} with the appropriate covariance. Formally, the lower bound on $\TL$ and an upper bound on $T$ are conditional on the eigenframework. This assumption has been empirically verified in a number of settings \citep{simon2021, canatar2021spectral,wei:2022-more-than-a-toy,misiak2024}.}, for\footnote{This choice of dimension is arbitrary. We can get close to $\SL=\tilde{O}_{d,k}(\TL^{1.5})$ with a larger $d$.} $d>200$, $\TL = \Omega_{d,k}( \frac{1}{\MT^{1.02}})$, and so as $\MT\to\infty$ we have that 
    \begin{align*}
        \SL = \tilde{O}_{d,k}\left( \TL^{1.49} \right).
    \end{align*}
\end{theorem}

For a more detailed statement spelling out the dependence on the degree $k$ and stopping time $T$, see \Cref{thrm:main:2layerReLUgeneral} and for the proof see \Cref{app:2layer}.

\begin{figure}[ht]
    \centering
    \subfigure[$\SL/\TL$ v.s. $t$]{\includegraphics[width=0.49\linewidth]{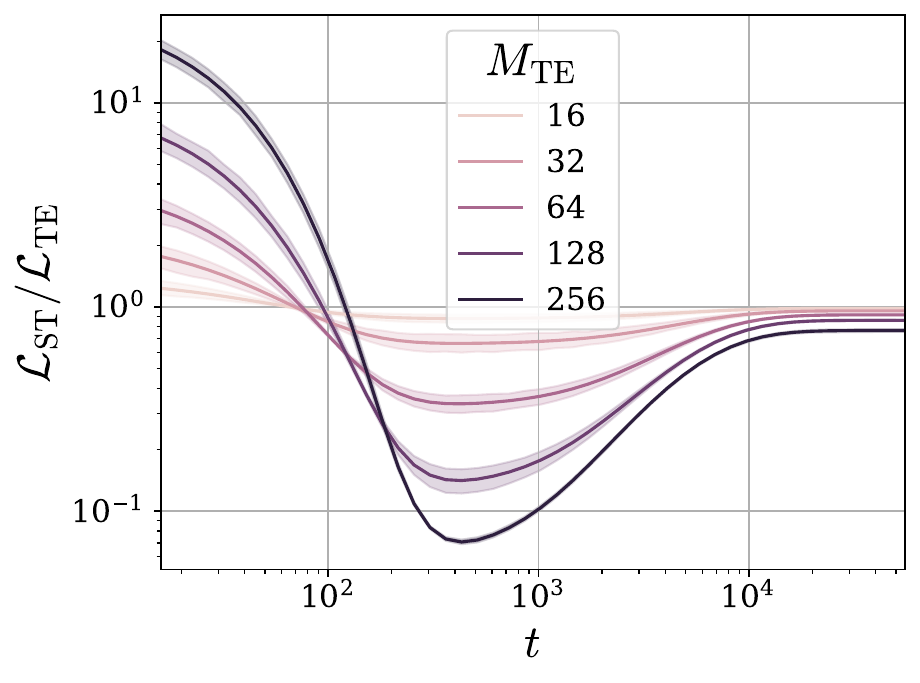}\label{fig:exp:2layer-loss-ratio}}
    \subfigure[$\SL$ v.s. $\TL$]{\includegraphics[width=0.49\linewidth]{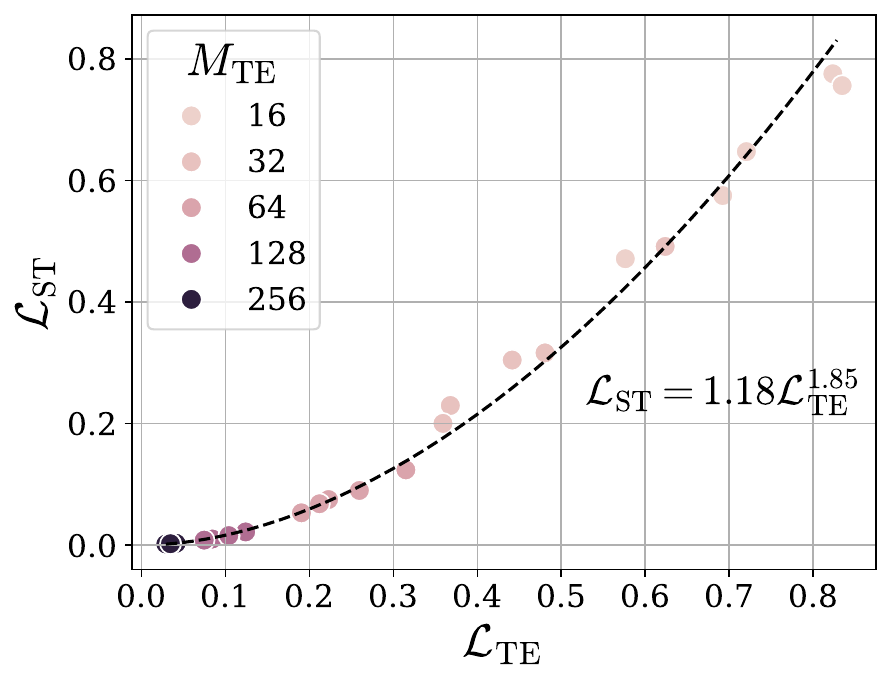}\label{fig:exp:2layer-loss-fitting}}
    \caption{Weak-to-strong generalization happens in ReLU random feature networks (\Cref{ass:relu}) with input dimension $d=32$, student size $\MS=16384$, and teacher size $\MT\in\{16, \ldots, 256\}$. We consider a linear target function $f^*(\vx)= \langle \beta, \vx\rangle$ for unit norm some $\beta$.  \Cref{fig:exp:2layer-loss-ratio} plots the ratio between student loss $\SL$ and teacher loss $\TL$, with varying teacher size $\MT$ and gradient flow training time $t$. With appropriate stopping time, we see a significant weak-to-strong generalization gain.  This gain diminishes with overtraining and running gradient flow to convergence, the student mimics the teacher, has the same error, and does not excite weak-to-strong generalization.  In \Cref{fig:exp:2layer-loss-fitting}, we fit the minimal student loss $\SL$ (at the optimal stopping time for each teacher size) as a power law function of the student loss $\TL$, confirming Theorem \ref{thrm:main:2layerrelu}. See \Cref{sec:exp} for simulation details.}
    \label{fig:exp:2layer-main}
\end{figure}

\Cref{thrm:main:2layerrelu} already establishes that in \Cref{ass:relu}, weak-to-strong generalization provably happens for a large set of targets, and for large teachers, we can get a performance gap recovered (PGR) arbitrarily close to $1$.  We see also that the PGR is not a fine enough measure of weak-to-strong generalization, since the separation is obtained when the teacher error is small, and the PGR does not tell us how small the student error $\SL$ can be as a function of the teacher error $\TL$: having $PGR\rightarrow 1$ only says that the student error $\SL$ goes to zero superlinearly faster than the teacher error $\TL$, but not how much faster.  Indeed, we can see that even in the ReLU Model \ref{ass:relu}, the student error is {\em polynomially} smaller, by an {\em exponent} of at least $1.49$. This last result relies on the widely verified Gaussian Universality Ansatz, and is also verified in our experiments: in \Cref{fig:exp:2layer-loss-ratio}, we show that when the target $f^*$ is linear, with proper early stopping time, the loss ratio $\SL/\TL$ decreases when $\MT$ grows. Furthermore, in \Cref{fig:exp:2layer-loss-fitting}, we observe that the student loss $\SL$ is polynomially smaller than the teacher loss $\TL$ with an estimated exponent even above our bound.  Indeed, we expect our bound on the behavior of ReLU Random Feature Networks is loose, and it might be possible to establish an even stronger separation.

In order to demonstrate a stronger separation theoretically, and also avoid the Gaussian Universality Ansatz, we turn to the Linear Network \Cref{ass:linearnetwork}:


\begin{theorem}[Weak-to-Strong Generalization with Linear Network]\label{thrm:main:diagfeatcov}
    Consider the linear network \Cref{ass:linearnetwork}, with $\Psi_d = (1,\dots,1,\psi_*,\ldots,\psi_*)$, where $1$ is repeated $k$ times and $\psi_*=(d-k)^{-2/3}$ is repeated $d-k$ times, and a linear target $f^*(\vx) = \inne{\beta}{\vx}$ supported by the first $k$ coordinates, i.e.~$\beta_i=0$ for $i>k$ that is normalized $\E[f^{*2}]=1$. In the weak-to-strong setting of Section \ref{sec:setup}, for any $k$ and $d>k+Ck^{3}$(where $C$ is an absolute constant), and a teacher of size $\MT=(d-k)^{2/3}$, if the student is trained until time $T=\Theta(\log \MT)$, we have that with probability at least $0.99$ for all $\MT$,
    \begin{align*}
        \SL\le \tilde{O}(k\TL^2). 
    \end{align*}
    In particular, for any fixed $k$, as $d,\MT\rightarrow\infty$, we have that  $\PGR\to 1$. 
\end{theorem}

\begin{figure}[ht]
    \centering
    \subfigure[$\SL/\TL^2$ v.s. $t$]{\includegraphics[width=0.49\linewidth]{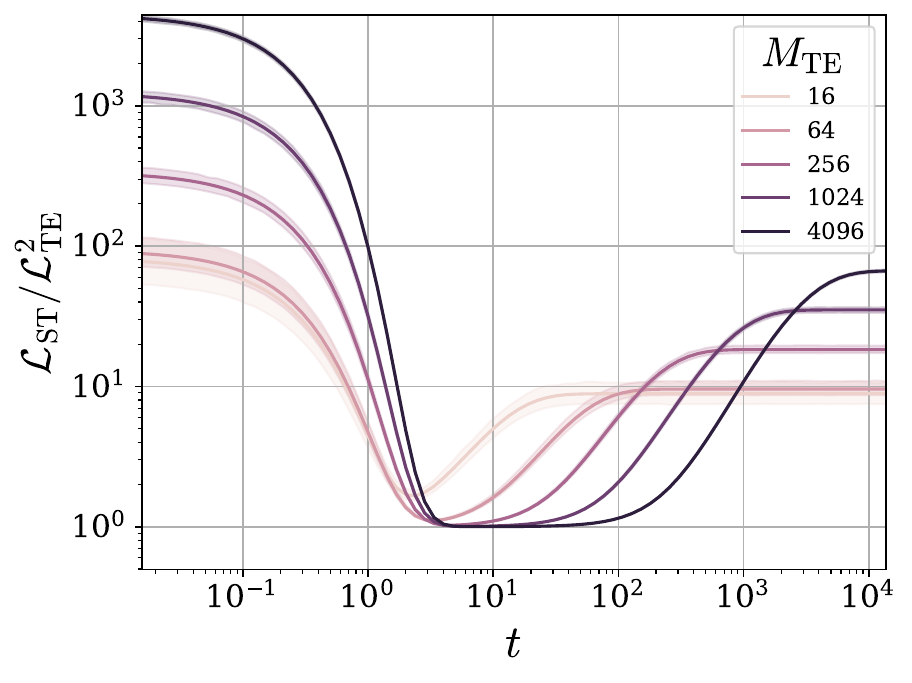}\label{fig:exp:linear-loss-ratio}}
    \subfigure[$\SL$ v.s. $\TL$]{\includegraphics[width=0.49\linewidth]{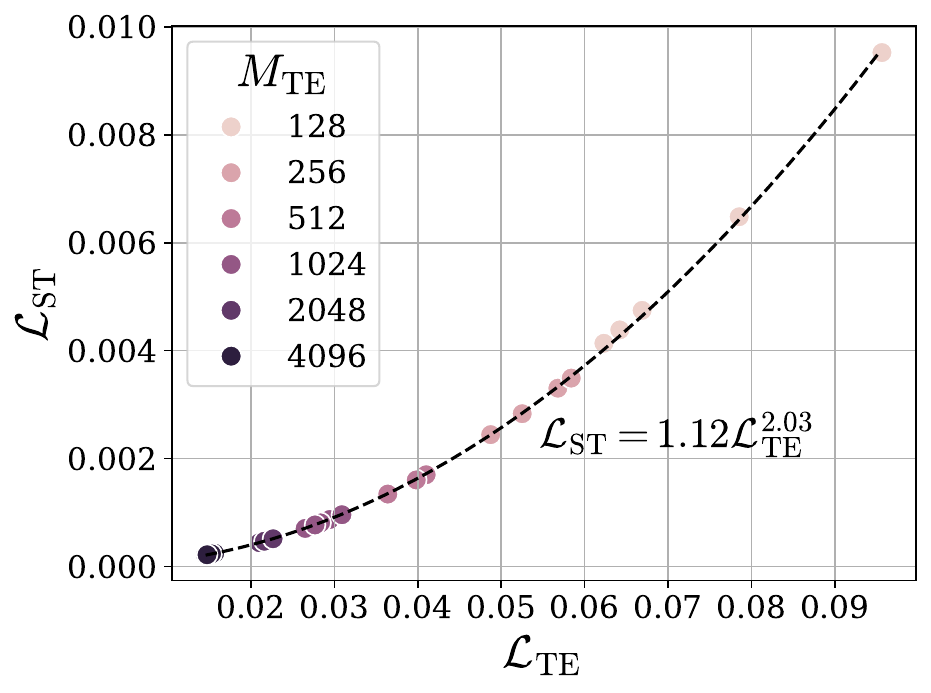}\label{fig:exp:linear-loss-fitting}}
    \caption{Weak-to-Strong generalization happens in random linear feature networks (\Cref{ass:linearnetwork}). 
 Here we used an input distribution as in \Cref{thrm:main:diagfeatcov},  with $k=1$ and a target function $f^*=\langle e_1, \vx\rangle$ where $e_1$ is the first standard basis vector. \Cref{fig:exp:linear-loss-ratio} plots the ratio between the student loss $\SL$ and squared teacher loss $\TL^2$, with varying teacher size $\MT$, and where the dimensionality $d=\MT^{3/2}$ as set as in the scaling of \Cref{thrm:main:diagfeatcov}, as a function of the gradient flow  time $t$. With proper early stopping time $\SL/\TL^2$ converges to approximately $1$ as $\MT$ grows, confirming that for large $\MT$ we have $\SL \propto \TL^2$ as in \Cref{thrm:main:diagfeatcov}.  This is also confirmed in \Cref{fig:exp:linear-loss-fitting}, where we fit the student loss $\SL$ as a power law function of teacher loss $\TL$, and recover an excellent fit with an exponent very close to $2$.  We again see that overtraining diminishes weak-to-strong generalization.  See \Cref{sec:exp} for simulation details.}
    \label{fig:exp:linear-main}
\end{figure}

\removed{\Cref{thrm:main:diagfeatcov} shows that the weak-to-strong generalization happens in \Cref{ass:linearnetwork}. Here, we not only have a performance gap recovered (PGR) going to $1$, but we also have a greater separation between the student and the teacher, in the sense that the student tends to $0$ quadratically faster.} A more general statement for \Cref{ass:linearnetwork} that holds for any covariance matrix $\Psi$ and target is given in \Cref{thrm:main:diagfeatgeneral}. For the proof, see \Cref{app:diagfeat}.  We indeed observe the behavior described by the Theorem in simulation experiments, as presented in \Cref{fig:exp:linear-main}, where we can see the student loss $\SL$ is indeed quadratically smaller than the teacher loss $\TL$, already for moderately sized networks.

\paragraph{$\Theta(1)$-error asymptotics}

\Cref{thrm:main:diagfeatcov} shows a quadratic gap between the student and teacher errors, but both errors vanish as the model size $\MT$ increases. Perhaps a more interesting and relevant scaling is a proportional scaling, where with larger model sizes we can handle increasingly more complex problems while still ensuring bounded error. A more specific variant of the separation result shows that for any asymptotic error, we can have a quadratic error gap even as model size and dimensionality jointly increase:

\begin{theorem}[$\Theta(1)$-error asymptotics for Linear Networks]\label{thrm:main:thetaone}
For any $\alpha>0$ there exists $d_{\alpha} \in \mathbb N$ such that for all $d>d_\alpha$, \Cref{ass:linearnetwork} with  $f^* = x_1$,
$\Psi_{d} =(1,\sqrt{\frac{\alpha}{d-1}},\dots,\sqrt{\frac{\alpha}{d-1}})$, and $\MT = \sqrt{\frac{d-1}{\alpha}}$, probability at least $0.99$ we have
\begin{align*}
    \TL > 
        0.98\frac{\alpha}{1+\alpha}~\text{  and  }~ \SL\le 100 \TL^2.
\end{align*}
\removed{Furthermore, for any $\varepsilon>0$ we have $\lim_{m\to\infty}\mathbb P\left( \TL\ge 0.98 \frac{\alpha}{1+\alpha}-\varepsilon\right)=1$.}
\end{theorem}

The proof can be found in \Cref{app:additionalproofslin}.

\removed{On the other hand, for the ReLU \Cref{ass:relu} in fixed dimension $d$ we don't have $\Theta(1)$-error asymptotics for any target function that is a sum of an even and a linear function (i.e. that can be represented by \Cref{ass:relu}) under Gaussian Universality.

\begin{theorem}[No $\Theta(1)$-error asymptotics for ReLU ]\label{thrm:main:nothetaoneforrelu}
    Under Gaussian Universality, if $f^*$ is spanned by the spherical harmonics of order that is either even and at most $k$ or $1$, for \Cref{ass:relu} in fixed dimension~$d$ we have that
    \begin{align*}
        \TL = o(1),
    \end{align*}
    i.e. we have to have $\TL \to 0$ as $\MT\to \infty$.
\end{theorem}

For the proof of \Cref{thrm:main:nothetaoneforrelu} see \Cref{app:gaussianthetaone}.}

\removed{We can only get $\Theta(1)$-error asymptotics in the case of increasing dimension. 

\begin{theorem}[]
    Let $f^*$ be spanned by spherical harmonics of order at most $k$. Then, for the ReLU network \Cref{ass:relu}, if we consider a sequence of problems of learning $f^*$ with $d$ increasing and $\MT=d^{2k}$, then with probability at least $1-$ there exists $C_L>0$ such that
    \begin{align*}
       \liminf_{m\to\infty} \TL \ge C_L
    \end{align*}
\end{theorem}}

\section{The Limit of Weak to Strong Generalization}\label{subsec:lowerbound}

In the previous Section we saw how a student can significantly outperform its teacher, even in random feature models.  In particular, we showed that a quadratic reduction in error is possible. One might ask if in some cases an even greater improvement is possible.  Perhaps a cubic or higher-order polynomial improvement?  In this Section, we show that the quartatic improvement of \Cref{thrm:main:diagfeatcov} is the largest possible in {\em any} random feature model.  No matter the feature distribution and target, and how related they are, if the teacher has error $\TL$, a student, even with many more features, cannot obtain error better than $\SL=\Omega(\TL)$.  

In fact, we show that this limitation holds much more broadly, not only in {\em random} feature models, but for any teacher optimal on some set of features and for any student trained with early stopped gradient flow on any other set of features.  The Random Feature models are a special case where the teacher and student models are small and large random subsets from the same feature distribution, but the limitation holds even if the student features are pre-trained or otherwise cleverly selected.  

Furthermore, the limitation applies to a broad class of training methods, which go beyond just early stopped gradient flow.  All we require is that the learned predictor is {\em shrinkage optimal} in the following sense:

\newcommand{\ftarget}{f_{\textrm{train}}}

\begin{definition}[Shrinking Optimality]\label{defi:shrinkage_optimality}
    Given two functions $\hat{f},\ftarget:\mathcal{X}\removed{\mathbb{R}^{d}}\to \mathbb{R}$, we say $\hat{f}$ is \emph{shrinnking-optimal} with respect to $\ftarget$ if for any $0\le \alpha\le 1$, $\E_{\vx}(( \hat{f}(\vx)-\ftarget(\vx))^2)\le \E_{\vx}((\alpha \hat{f}(\vx)-\ftarget(\vx))^2)$.
\end{definition}

That is, all we require is that the training method returns a predictor $\hat{f}$ which cannot improve its training objective, simply by scaling down, or shrinking toward the origin.  Shrink optimality is satisfied by loss minimization on a linear subspace (i.e.~our learning rule for the teacher), or more generally any convex class that includes the origin.  It is also satisfied by early stopped gradient flow over any feature space or with any kernel, and with any stopping time (including $T=\infty$, which corresponds to just loss minimization):

\begin{lemma}[Shrinkage Optimality of Gradient Flow Solutions]\label{lemma:gf_shrinkage_optimal}
    For any feature map $\phi(x)$ and any target $\ftarget(x)$ consider gradient flow $\dot{\vw} = -\nabla_\vw \E_{\vx}\left[(\langle\vw,\phi(\vx)-\ftarget(\vx))^2\right]$ initialized at $\vw(0)=0$.  Then at any time $0\leq T\leq\infty$, the predictor $f_T(\vx)=\langle\vw(T),\phi(\vx)\rangle$ is shrink optimal w.r.t.~$\ftarget$.
\end{lemma}

\Cref{lemma:gf_shrinkage_optimal} shows that in the Random Feature setup considered in \Cref{sec:setup}, the teacher and the student both satisfy the shrinkage optimality \Cref{defi:shrinkage_optimality}, with respect to target and the teacher predictor respectively. 

With these definitions in place, we are ready to state our main lower bound on weak-to-strong generalization:

\begin{theorem}[Genereal Limitation of Weak-to-Strong Generalization]\label{thrm:main:generallimit}
    For any target $f^*$ normalized s.t.~$\E_{\vx}\left[f^*(\vx)^2\right]\leq 1$, for any teacher  $f_{\text{teacher}}$ that is shrinking-optimal to $f^*$ and any student $f_{\text{student}}$ that is shrinking-optimal to $f_{\text{teacher}}$, it holds that
    \begin{align*}
        \SL \ge  \frac{\left(\sqrt{1+3\TL}-\sqrt{1-\TL}\right)^2}{4} \ge \frac{3}{4} \TL^2. 
    \end{align*}
\end{theorem}

\begin{corollary}[Limit of Random Feature Weak to Strong Generalization]\label{thrm:main:lowerbound}
    For any Random Feature Network Weak-to-Strong training as in \Cref{sec:setup}, any normalized target $\E[{f^*}^2]=1$, and any student stopping time $T$, we have that
    \begin{align*}
        \SL \ge  \frac{\left(\sqrt{1+3\TL}-\sqrt{1-\TL}\right)^2}{4} \ge \frac{3}{4} \TL^2. 
    \end{align*}
\end{corollary}

For the proof of \Cref{lemma:gf_shrinkage_optimal}, \Cref{thrm:main:generallimit}, and \Cref{thrm:main:lowerbound}, see \Cref{app:limitation}.

\Cref{thrm:main:generallimit} and \Cref{thrm:main:lowerbound} establish significant limits on the extent of weak-to-strong generalization.  We see that although the PGR can be arbitrarily close to one, and the student error can be quadratically smaller than the teacher error, it cannot be any smaller than that.  We cannot have arbitrarily high teacher error and arbitrarily low student error.  In particular, we cannot have a situation in which the teacher error is very close to the null risk $\E[{f^*}^2]$, i.e.~the teacher is barely learning, while the student error is low (if $\TL$ is closer to one, then $\SL$ must also be close to one).  That is, weak-to-strong learning {\em cannot} significantly ``boost'' a teacher that is only slightly better than chance. And, for the student error to go to zero, {\em the teacher error must also go to zero}.

These limitations hold not only in random feature models, but based on \Cref{thrm:main:generallimit}, also much more broadly.  First of all, they hold with {\em any} student and teacher feature space---these need not be random.  We saw in \Cref{sec:quantifying-w2s} that a quadratic weak-to-strong improvement is possible even with random features, with the same student and teacher feature distributions.  But what \Cref{thrm:main:generallimit} tells us is that even if the student features are much more specialized than the teacher features, even if they are more aligned with the target, or even if they are pre-trained, we cannot get any larger improvement.  Furthermore, this holds for a very general class of learning rules.  Beyond early stopped gradient flow, using any type of convex regularizer (minimized at the origin) or constraining to any convex hypothesis class (which includes the origin) also ensures shrink-optimality.  Any such learning rule is still subject to the limitation of \Cref{thrm:main:generallimit}.  

The main caveat here is that the limitation applies only with gradient flow {\em initialized at the origin}, with convex hypothesis classes {\em containing the origin} or with convex regularizers {\em minimized at the origin}.  Concretely, consider a student that uses a pre-trained feature space, but trains the top layer weights from scratch (starting from the origin).  This student is subject to the limitation of \Cref{thrm:main:generallimit}.  In contrast, consider a student that fine-tunes a pre-trained top layer, starting from the pre-trained initialization (or alternatively, explicitly regularizes to be close to the pre-trained model).  Such a student is {\em not} shrink optimal and is {\em not} subject to the limitation of \Cref{thrm:main:generallimit}.  Indeed, if the pre-trained model is already very good (at an extreme---if the pre-trained model already happens to be better on the task than the teacher), the student can be arbitrarily better than its teacher (similar to the Chinese-speaking student outperforming its non-Chinese-speaking teacher in Chinese).  In the setting of \cite{charikar2024quantifying}, the difference is whether the teacher and student classes contain the origin (in which case shrink optimality holds and \Cref{thrm:main:generallimit} applies), or do not contain the origin.   One can think of this distinction in terms of whether the inductive bias specified by the feature space, learning rule, initialization, and hyopthesis class is {\em generic}, i.e.~extends from the origin, prefering certain directions but not a particular point in predictor space, or {\em specialized}, i.e.~prefering not only particular directions of features, but a very specific bias.

\paragraph{Bootstrapping.} A natural question to ask is whether extending the weak-to-strong setup described in \Cref{sec:setup} to use a chain of students can help break the limitations of \Cref{thrm:main:lowerbound}.  That is, training a teacher on the target, the first student on the teacher, the second student on the first student, and so on in a bootstrapped game of ``telephone'', then finally asking if the final student can be much better than the teacher.  The lower bound of \Cref{thrm:main:generallimit} and \Cref{thrm:main:lowerbound} do not apply here as-is, because shrink-optimality is not transitive.  That is, if $f_2$ is shrink-optimal w.r.t.~$f_1$, and $f_1$ is shrin-optimal w.r.t.~$f_\textrm{teacher}$, this does not imply $f_2$ is shrink-optimal w.r.t.~$f_\textrm{teacher}$.  Instead, we provide a slightly weaker lower bound against a more general property, which {\em is} transitive:

\begin{theorem}[Limitation of Weak-to-Strong Generalization with a Bounded Student]\label{thrm:main:normboundedlimit}
     If the target $f^*$ is normalized, $\E_{\vx\sim\cD}[ (f^*(\vx))^2]=1$, then for any teacher $f_{\text{teacher}}$ shrinking-optimal w.r.t. $f^*$ and any student $f_{\text{student}}$ s.t. $\E_\vx\left[f_\text{student}(\vx)^2\right] \leq \E_\vx\left[f_\text{teacher}(\vx)^2\right]$, we have that
    \begin{align*}
        \SL \ge \left(1-\sqrt{1-\TL} \right)^2  \ge \frac{1}{4}\TL^2
    \end{align*}
\end{theorem}

It is easy to verify that the shrink optimality implies $\E_\vx\left[\hat{f}(\vx)^2\right] \leq \E_\vx\left[\ftarget(\vx)^2\right]$, and the transitivity of the inequality now implies:

\begin{corollary}[Limitation of Weak-to-Strong Generalization with Bootstrapping]\label{thrm:main:boostinglimit}
    If the target $f^*$ is normalized, $\E_{\vx\sim\cD}[ (f^*(\vx))^2]=1$, then for any teacher $f_{\text{teacher}}$ shrinking-optimal w.r.t. $f^*$ and any chain of student $f^{(i)}_{\text{student}}$, $i=0,2,\ldots,\infty$, where $f^{(i)}_{\text{student}} = f_{\text{teacher}}$ and $f^{(i+1)}_{\text{student}}$ is shrinking-optimal w.r.t. $f^{(i)}_{\text{student}}$, we have that for all $i\ge 1$
    \begin{align*}
        \cL_{\mathrm{ST},i} \ge \left(1-\sqrt{1-\TL} \right)^2  \ge \frac{1}{4}\TL^2,
    \end{align*}
where $\cL_{\mathrm{ST},i}$ is the test loss of the $n$-th student.
In particular, this includes the case where the student $f_{\text{student}}^{(i)}$ is trained by gradient flow w.r.t. $f_{\text{student}}^{(i-1)}$ for arbitrary stopping time $T_i$, and using arbitrarily (random or determnistic) features.
\end{corollary}

For the proof of \Cref{thrm:main:normboundedlimit} and \Cref{thrm:main:boostinglimit} see \Cref{app:limitation}.

\section{How Does Weak-to-Strong Generalization Happen?}\label{sec:howdoesw2shappen}


\paragraph{Linear Network.} To understand how the strong student can correct their weak teacher and obtain a lower error, let us first consider linear networks as in  \Cref{ass:linearnetwork}. In this case, both weak and strong models, $f_{\text{teacher}}(\vx)$ and $f_{\text{student}}(\vx)$, are linear functions of $\vx$, and thus can be written as $f_{\vw,\vu}(\vx)=\inne{\beta}{\vx}$ where $\beta=\mU^\top \vw$.
First, let us understand the teacher predictor. The teacher has a small number of features $\MT$, so they are not aligned with coordinate directions $x_i$. Instead, the teacher can only learn $\vbeta_{\textrm{teacher}} \in \text{span}(\TU)$, and so learns $\vbeta_{\textrm{teacher}}$ to be the projection of $f^*$ to $\text{span}(\TU)$. So, even when the ground truth $f^*=\langle \beta^*, \vx \rangle$ is spanned by the first few coordinate directions (say there is $K^*$ so that $\beta^*_i=0$ for all $i>K^*$), the teacher predictor will have nonzero components in all directions $x_i$. That is, $f_{\textrm{teacher}} = \langle \beta_{\textrm{teacher}}, \vx\rangle$ with $\beta_{\textrm{teacher}}\neq 0$ for all $i$. Comparing the ground truth $f^*$ with the teacher predictor $f_{\textrm{teacher}}$ in directions $x_i$, we see that since $\beta^*_i=0$ for all $i>K^*$, all of the teacher predictor components in directions $x_i$ for $i>K^*$ are noise. Therefore, the student can decrease error over the teacher by decreasing these directions of the teacher predictor $f_{\textrm{teacher}}$ while keeping the projections to $x_i$ for $i\le K^*$ close to the teacher. To understand how this can happen, let us understand the student's learning rule, i.e. early-stopped gradient descent of a random features network. Let $ f_{\vw, \vu}(\vx)$ be the learned predictor whose first layer has random features $\sigma(\langle \vu_{i},\vx\rangle)$. Learning a predictor $f$ with an early-stopped gradient descent shrinks $f-f_{\vw, \vu}$ in the direction of the model's random features $\sigma(\langle \vu_{i},\vx\rangle)=(\mU \vx)_i$ proportional to the variance of each of the features. For a very wide student, i.e. with $\MS\to \infty$, this amounts to the student's features effectively being the coordinate directions $x_i$, so the shrinkage of $f-f_{\vw, \vu}$ in directions $x_i$ will be proportional to the variance $\psi_i$ of $x_i$. This is because as $\MS\to \infty$, we have that $\mU_{\text{ST}}^\top \mU_{\text{ST}} \to \MS I$. This makes learning directions $x_i$ with larger variance $\psi_i$ (i.e. for smaller $i$) significantly faster than the ones with smaller variance $\psi_i$ (i.e. for larger $i$). Now we can understand why the student's learning rule reduces error for the teacher predictor and how weak-to-strong happens. Since the student's learning rule shrinks $f_{\textrm{teacher}}-f_{\textrm{student}}$ along directions $x_i$, the student predictor $f_{\textrm{student}} = \langle \beta_{\textrm{student}}, \vx\rangle$ will have $\beta_{\textrm{student},i}$ between the initialized valued and $\beta_{\textrm{teacher},i}$ depending on how early the gradient descent is stopped. So, because the student is initialized at $0$, the student improves over the teacher's error by keeping small variance directions close to initialization, i.e, effectively "zeroing out" small variance directions by having $\beta_{\textrm{student},i}\approx 0$ for all large $i>K^*$. This happens with situably chosen stopping time $T$ if there is a sudden drop in variance $\psi_i$ at index $K$, i.e. $\psi_K\gg \psi_{K+1}$ (as long as $K>K^*$), because the time to learn any direction $x_i$ for $i\le K$ will be much smaller than the time to learn any of the directions $x_i$ for $i>K$. So, in this case, we can choose a stopping time $T$ to be such that all large variance directions $x_i$ are learned, that is $(f_{\textrm{teacher}}-f_{\textrm{student}})_i$ is very small for the first $K$ direction $x_i$. In other words, $(f_{\textrm{student}})_i \approx f_{\textrm{teacher},i}$ for large variance directions (i.e. small index) $x_i$, but all the small variance directions (i.e. large index) $x_i$ of $f_{\textrm{student},i}$ are close to initialization, zero. In this way, the student predictor $f_{\text{student}}(\vx)$ effectively learned the high variance (i.e. small index $i$) directions $x_i$ of the teacher predictor $f_{\text{teacher}}$ and zeroed out the small variance (i.e. large index $i$) directions $x_i$ of the teacher predictor $f_{\text{teacher}}$. All of this can be explicitly seen in the closed form solution of gradient flow in function space, \Cref{eq:gradflow}. \removed{ Let us now turn to the teacher. Although the signal $f^*$ (i.e. the target of GF) is entirely in the first few high variance coordinates of $\vx$, say $\{x_i\}_{i\le K}$, if we have a small number of teacher nodes $\MT$, they will not be directly represented in the hidden layer, i.e. in $\TU \vx$. Instead, the teacher can only learn $\vbeta_{\mathrm{TE}} \in \text{span}(\TU)$, and so learns $\vbeta_{\mathrm{TE}}$ to be the projection of $f^*$ to $\text{span}(\TU)$. This means that $f_{\text{teacher}}(\vx)$ has some energy (non-zero coefficients) in all the coordinates.}The student shrinkage will reduce the noise (i.e. teacher signal uncorrelated to the ground truth) in low variance coordinates by zeroing out the coefficients along those directions. Therefore, the weak-to-strong improvement comes exactly from improvement along the noise directions (i.e. directions uncorrelated with the ground truth). Note that if we train for too long, the effect will be the same in all directions, so there will be no weak-to-strong generalization. So early stopping is crucial in this setup.

As student training progresses, we fit more of the teacher's signal (i.e. the target of GF), but the shrinkage effect reduces. This can be seen in Figures \ref{fig:exp:2layer-main} and \ref{fig:exp:linear-main}, where overtraining hurts weak-to-strong generalization.  At the limit, as training time goes to infinity, the student would converge to exactly mimicking the teacher, and thus converge to the teacher error, without any weak-to-strong benefit, as can be clearly seen in Figure \ref{fig:exp:2layer-main}.

The difficulty of applying this reasoning\removed{in linear networks, \Cref{ass:linearnetwork}, and ReLU networks, \Cref{ass:relu},} is that we need to establish that a significant portion of the teacher error is actually present in the noise direction. In \Cref{thrm:main:2layerrelu} and \Cref{thrm:main:diagfeatcov}, we not only show that a significant portion of the teacher's error $\TL$ will be along the low energy directions but also that when the number of teacher's units $\MT$ is large, most of the teacher's error lies along the low energy, i.e. noise directions. Our main result, \Cref{thm:general-main}, handles this difficulty. Proving that most of the teacher error $\TL$ is in the low energy directions allows us to conclude that the $\text{PGR}$ goes to $1$ as the number of teacher units $\MT$ increases, instead of just being positive, and establish a quantitative relationship between the student and teacher errors. 

\paragraph{$2$-layer ReLU.} For ReLU, we have a similar situation except that the relevant directions for a wide student will be spherical harmonics. Learning a predictor $f_{\vw,\vu_s}$ using $f$ as a target with early stopped gradient descent will shrink $f-f_{\vw,\vu_s}$ in the direction of the student's random features $\sigma(\langle \vu_{s,i}, \vx \rangle) =\text{ReLU}(\langle \vu_{s,i}, \vx \rangle) $, similarly to the case of linear networks. The relevant basis for a wide student, i.e. in the $\MS\to \infty$ case, is not the coordinates of the input $\vx_i$ but rather spherical harmonics $\phi_{i,k}(\vx)$. The difference comes from the fact that our probability distribution is now $\cD = \text{Unif}(\mathbb S^{d-1})$, and the spherical harmonics $\{\{ \phi_{i,k}\}_{i=1}^{N_k}\}_{k=1}^{\infty}$ are the relevant basis here. In this basis, the teacher predictor is given by $f_{\text{teacher}}(\vx) = \sum_{i,k} \beta_{\textrm{teacher},i,k} \phi_{i,k}(\vx)$. The student will learn the teacher's coefficients $\beta_{\textrm{teacher},i,k}$ for $\phi_{i,k}$ faster for the spherical harmonics of lower order $k$. Since the covariance of the features is given by the square of the coefficients in the decomposition of the ReLU function in the basis $\phi_{i,k}$, $\sigma_k^2$, they will naturally have jumps for certain indices $k$. Similarly, for any one of these jumps at, say, $k=K$, i.e. $\sigma_{K}\gg \sigma_{K+2}$ (odd index $\sigma_i$ are zero for $i>1$ for the ReLU function), the gradient descent with a wide student will learn the directions $\phi_{i,k}(\vx)$ for $k\le K$ significantly faster than $\phi_{i,k}(\vx)$ for $k> K$. So, again, we can choose an appropriate stopping time $T$ such that the student predictor $f_{\textrm{student}}$ learns the large variance directions, $\phi_{i,k}$ for $k\le K$, of the teacher predictor $f_{\text{teacher}}$ almost completely, but for small variance directions, $\phi_{i,k}$ for $k>K$, is very close to the initialization, i.e. zero. This means that the student can denoise the direction of $\phi_{i,k}$ for $k>K$. So, in particular, if all of the signal $f^*$ is aligned with the directions of the first $k_0$ order of spherical harmonics, then everything the teacher learns in the directions $\phi_{i,k}$ for $k>k_0$ is the incorrectly learned noise. Zeroing out this incorrectly learned noise by stopping the gradient flow at the right stopping time enables weak-to-strong generalization.

To understand the inner workings of weak-to-strong generalization in this setup further, we turn to the dynamics of GF (\Cref{eq:gradflow}) in \Cref{sec:general}. This will be simplified if we consider the student's network as a kernel.  

\section{General Setup and Results}\label{sec:general}

In \Cref{subsec:general-mult-w2s}, we show a general result that describes weak-to-strong improvement of a student described by a kernel trained with gradient flow and an optimally trained teacer with some fixed set of features, which is \Cref{thm:general-main}. In \Cref{subsec:w2s-rf}, we show that if the teacher and student are described by ReLU and linear random feature networks, i.e. \Cref{ass:relu} and \Cref{ass:linearnetwork}, then the weak-to-strong benefit from \Cref{thm:general-main} will be significant with high probability, which are \Cref{thrm:main:2layerNNgeneral} and \Cref{thrm:main:diagfeatgeneral}.



\subsection{General Multiplicative Weak-to-Strong Improvement}\label{subsec:general-mult-w2s}

\paragraph{Gradient Flow Dynamics Recap.} We can rewrite \Cref{eq:gradflow} in function space from parameter space in order to analyze the infinite-width limit $\MS\to\infty$ in the same space. To do that, it will be convenient to consider the student to be described by its kernel 
\begin{align}\label{eq:inducedkernel}
\begin{aligned}
    \gK(\vx, \vx') := \E_{\vu\sim\text{unif}(\mathbb S^{d-1})}\left[\sigma(\langle \vu,\vx\rangle)\sigma(\langle \vu,\vx' \rangle) \right].
    \end{aligned}
\end{align}

 The kernel $\gK$ can be decomposed as $\gK(\vx, \vx') = \sum_{k \ge 1} \lambda_k e_k(\vx) e_k(\vx')$,
where $\{\lambda_i \}_{i=1}^{\infty}$ are the eigenvalues of the associated kernel operator in descending order, and $\{e_i(\vx)\}_{i=1}^{\infty} $ are orthonormal eigenfunctions in the inner product with respect to $\cD$, which we denote $\inne{\cdot}{\cdot}_{\cD}$, i.e. $\inne{e_i}{e_j}_{\cD}=\delta_{ij}$. The inner product is defined as $\langle f,g\rangle_{\cD}:=\E_{\vx\sim\cD}\left[ f(\vx)g(\vx)\right]$. 
We can derive (see \Cref{app:gradflow}) the following closed-form expression for the gradient flow dynamics of the student~(\Cref{eq:gradflow}) at time $t$ in the $\MS\to\infty$ limit: 
\begin{equation} \label{eq:student-dynamics}
    f_{t} = \sum_{k \ge 1} \left(1 - e^{-\lambda_k t}\right) \inne{f_{\text{teacher}}}{e_k}_{\cD}e_k
\end{equation}
We note that given data distribution $\cD$ (and thus the inner-product $\inne{\cdot}{\cdot}_\cD$), the above solution of gradient flow at time $t$ is only a function of target $f_{\textrm{teacher}}$, time $t$, and the kernel $\mathcal{K}$, which we denote by 
\begin{equation}\label{eq:studen-predictor}
f_t= \cT^{\cK}_t(f_{\text{teacher}}), \quad \text{where} \quad \cT^\cK_t\triangleq \mathtt{id} - e^{-t\cK},
\end{equation}   
where $\mathtt{id}$ is the identity mapping. In other words, if different two-layer networks have different bottom layers but with the same induced kernel, they lead to same training behavior under gradient flow. For an outline of the derivation of gradient flow dynamics, see \Cref{app:gradflow}. 

\paragraph{Student-Teacher Improvement Bound for Early Stopped Gradient Flow.}First, in \Cref{thrm:main:early_stop_loss}, we will derive a bound on the error of a student trained with an early stopped gradient flow in the weak-to-strong setup. Consider a generalized 2-stage learning process of weak-to-strong setup from \Cref{sec:setup}, where the student is trained with gradient flow for time $t$ with kernel $\cK$ on population given by the teacher predictor $f_{\textrm{teacher}}$, i.e. as given by \Cref{eq:studen-predictor,eq:student-dynamics} and denoted with $f_{t}=\cT^{\cK}_t(f_{\text{teacher}})$, but the teacher predictor can be an arbitrary square integrable function in the eigenspace of $\cK$. Furthermore, we require that the student's kernel $\gK$ correctly captures the prior of the target function $f^*$ in the sense that $f^*$ is supported on the top-$K$ eigenspace of $\gK$ with nonzero eigenvalues:
\begin{assumption}[Ground Truth]
    \label{ass:fstar}
    $f^*$ is supported on the top-$K$ eigenspace of $\gK$ corresponding to nonzero eigenvalues, i.e., if $\lambda_k=0$ for $k\le K$ then $\inne{f^*}{e_k}_{\cD}=0$ and
    $\forall k > K: \inne{f^*}{e_k}_{\cD} = 0$.
\end{assumption}
Then, we can bound the students error at time $T$ with respect to the ground truth $f^*$ in terms of the spectrum of the kernel $\lambda_i$.

\begin{lemma}\label{thrm:main:early_stop_loss}    Under~\Cref{ass:fstar} for any kernel $\cK$, any function $f_{\textrm{teacher}}$ in the eigenspace of $\cK$, all stopping times $T > 0$, and any index $S \ge K$, we have that for the predictor $f_{t}=\cT^{\cK}_t(f_{\text{teacher}})$
    \begin{align*}
        \cL(f_T) \le 
        \cL(f_{\text{teacher}})
        + \frac{e^{-\lambda_K T}}{2 - e^{-\lambda_K T}} \norm{f^*}_{\cD}^2
        -(1 - \lambda_{S+1}^2 T^2) \sum_{k \ge S + 1} \inne{f_{\text{teacher}}}{e_k}_{\cD}^2.
    \end{align*}
\end{lemma}

\Cref{thrm:main:early_stop_loss} is an intermediate lemma that shows that the error of an early stopped GF solution with respect to the grond truth $f^*$ can be bounded in terms of the target $f_{\textrm{teacher}}$ and the projection of the target $f_{\textrm{teacher}}$ into the eigenspace of order $\ge S+1$, $\sum_{k \ge S + 1} \inne{f_{\text{teacher}}}{e_k}_{\cD}^2$. So, to get the final weak-to-strong upper bound (i.e. \Cref{thm:general-main}), we show in \Cref{thrm:main:random-feature-error} that for the optimally trained teacher, the proportion of error in eigendirections of order $\ge S+1$ is lower bounded by a quantity we call \emph{teacher-student feature alignment}. 

\paragraph{Teacher-Student Feature Alignment.} The key quantity in our characterization of weak-to-strong generalization gap is \emph{teacher-student feature alignment}, $\kappa_S$. We will introduce it here.

Let $\MT=m$ and let $\{e_i\}_{i=1}^{\infty}$ be the eigenbasis of the student's kernel $\cK$. Let $\vw\in \R^m$ be the teacher's weights, and let the teacher be parametrized as $f_{\text{teacher}} = f_{w} = \sum_{i=1}^{m} w_i g_i$, where $\{g_i\}_{i=1}^{m}$ are teacher's units, i.e. features. Let the student's predictor at time $T$ be $f_T$. For $S \ge K$, let $\mA_S, \mB_S \in \R^{m\times m}$ be the Gram matrices of teacher's features projected onto the top-$S$ eigenspace of $\gK$, {\it i.e.}, $A_{S,i,j} := \sum_{k = 1}^{S} \inne{g_i}{e_k}_{\cD} \inne{g_j}{e_k}_{\cD}$,
$B_{S,i,j} := \sum_{k \ge S + 1} \inne{g_i}{e_k}_{\cD} \inne{g_j}{e_k}_{\cD}$. Then we define the teacher-student feature alignment $\kappa_S$ as  
\begin{align} \label{eq:def-kappa}
    \kappa_S := 
    \frac{1}{1+\lambda_1\!\left((\sqrt{\mA_S})^+ \mB_S (\sqrt{\mA_S})^+\right)}.
\end{align}
$\kappa_S$ measures the alignment of the teacher's features with the top-$S$ eigenspace of the student kernel and is a completely deterministic quantity. \removed{It represents the cosine of the maximal principal angle between the top-$S$ eigenspace and teacher features, $\kappa_S = \inf_{\va\in \textrm{span}(\{ e_i\}_{i=1}^{S}), \| \va \|=1} \sup_{\vb \in \textrm{span}(\{g_i\}_{i=1}^{m}), \| \vb \|=1} \langle \va,\vb \rangle^2$.}As explained in \Cref{sec:howdoesw2shappen}, this misalignment is crucial for enabling weak-to-strong generalization in this setup.

We will refer to $\hat{f}_{w}$ the optimally trained teacher. We will require that no two units in the first layer of the teacher will be linearly dependent among the first $K$ directions $\{e_i\}_{i=1}^{K}$.
\begin{assumption}\label{ass:rankA}
    $\rank(\mA_K) = K$.
\end{assumption}

\begin{lemma}[Error Ratio of Weak-to-Strong Predictor in High Order Eigendirections]\label{thrm:main:random-feature-error}
    In the random feature model in \Cref{eq:2layernet}, under~\Cref{ass:fstar,ass:rankA},
    if $\gL(\hat{f}_{\vw}) = \min_{\vw' \in \R^m} \gL(\hat{f}_{\vw'})$, then
    \begin{align*}
        \sum_{k \ge S + 1} \inne{\hat{f}_{\vw}}{e_k}_{\cD}^2
        \ge \kappa_{S} \cdot \cL(\hat{f}_{\vw}).
    \end{align*}
\end{lemma}
\Cref{thrm:main:random-feature-error} shows that if the signal is spanned by first $S$ eigendirections then at least $\kappa_{S}$ portion of error of the random feature predictor in \Cref{eq:2layernet}
will be in the eigendirections of order $\ge S+1$.

Using \Cref{eq:student-dynamics}, \Cref{thrm:main:early_stop_loss}, and \Cref{thrm:main:random-feature-error} we can understand how weak-to-strong happens from \Cref{sec:howdoesw2shappen} more clearly. In this notation, in \Cref{sec:howdoesw2shappen}, $\langle f_{\text{teacher}},e_k\rangle_{\cD}$ is what we call $\beta_i$ and $e_i(\vx)$ are $x_i$ in \Cref{ass:linearnetwork} and spherical harmonics $\phi_{i,k}(\vx)$ in \Cref{ass:relu}. Note that since $\lambda_k$ are decreasing, $1-e^{-\lambda_k t}$ is closer to $1$ for larger eigenvalues $\lambda_i$, i.e. smaller indices $i$. In the special case that there is $K$ for which $\lambda_{K+1}$ is significantly smaller than $\lambda_{K}$, then for stopping time $T \in (\frac{1}{\lambda_K}, \frac{1}{\lambda_{K+1}})$, all $e^{-\lambda_iT}$ all going to be close to $0$ for $i\le K$, but $e^{-\lambda_{i}T}$ for $i\ge K+1$ will be close to $1$, effectively zeroing out $f_{\text{teacher}}$ in directions $e_{\ge K+1}$. This means that in \Cref{thrm:main:early_stop_loss}, the second term $\frac{e^{-\lambda_K T}}{2 - e^{-\lambda_K T}} \norm{f^*}_{\cD}^2$ will be small. Taking $S=K+1$, the thirm term will be close to $\kappa_{K+1}$ portion of the error by \Cref{thrm:main:random-feature-error}, which \Cref{thrm:main:early_stop_loss} shows that we zero out. This is the intuition behind \Cref{thm:general-main}.

\paragraph{Multiplicative Error Improvement in Weak-to-Strong Generalization.} Our main result establishes multiplicative error improvement in the weak-to-strong setup for a 2-layer network teacher with any features, \emph{even non-random}, and a student described by a kernel $\cK$.

\begin{theorem}[Weak-to-Strong Multiplicative Error Improvement]\label{thm:general-main}
    For any groundtruth $f^*$, p.s.d. kernel $\cK$ and positive integer $K$ satisfying \Cref{ass:fstar} and any teacher model $f_{\text{teacher}}$ satisfying \Cref{ass:rankA} and \Cref{ass:teacher_minimizer} ($f_{\text{teacher}}$ attaining minimum loss), then for any time $T > 0$,
    then the test loss of student $f_{\text{student}}\triangleq \cT^\cK_T(f_{\text{teacher}})$ has the following upper bound:
    \begin{align*}
        \SL &\le 
        \inf_{S \ge K} \bigg\{\left(
        1- (1 - \lambda_{S+1}^2 T^2) \kappa_S
        \right)
        \TL
        + \frac{e^{-\lambda_K T}}{2 - e^{-\lambda_KT}} \norm{f^*}_{\cD}^2\bigg\}.
    \end{align*}
\end{theorem}


\Cref{thm:general-main} shows that the student's error is roughly $(1-\kappa_S)$ fraction of teacher's if we can choose the stopping time $T$ and index $S$ so that $\lambda_{S+1}^2T^2$ and $\frac{e^{-\lambda_K T}}{2-e^{-\lambda_KT}}$ are both small. For each $S$, the optimal $T$ is such that $\lambda_{K}T$ is large but $\lambda_{S+1}T$ is small. Taking $T=\frac{1}{\lambda_K}\log \frac{1}{\delta}$, we have $e^{-\lambda_K T} = \delta$ and $\lambda_{S+1}T = \frac{\lambda_{S+1}}{\lambda_K}\log \frac{1}{\delta}$. If $\lambda_{S+1}<\lambda_K$ is sufficiently small, we can take a very small $\delta$, so that both terms $\frac{e^{-\lambda_K T}}{2 - e^{-\lambda_KT}} \norm{f^*}_{\cD}^2$ and $\lambda_{S+1}T$ are small, so the upper bound will be dominated by the $(1-\kappa_S)\TL$ term. In particular, this can be done if there is a big eigengap at some index $S>K$, i.e. $\lambda_{S+1}\ll \lambda_{S}$, because then $\frac{e^{-\lambda_K T}}{2-e^{-\lambda_KT}} <\delta$ and $\lambda_{S+1}T = \frac{\lambda_{S+1}}{\lambda_K} \log \frac{1}{\delta}<\frac{\lambda_{S+1}}{\lambda_S} \log \frac{1}{\delta}\ll \log \frac{1}{\delta}$. Note that \Cref{thm:general-main} is completely deterministic and holds both for random and non-random teachers. For the proof of \Cref{thrm:main:early_stop_loss}, \Cref{thrm:main:random-feature-error}, and \Cref{thm:general-main}, and the explanation for when equalities hold, see  \Cref{app:proof:general-main}.

\subsection{Weak-to-Strong Improvement in Random Feature Networks}\label{subsec:w2s-rf}

In this section, we provide detailed versions of \Cref{thrm:main:2layerrelu,thrm:main:diagfeatcov}, namely \Cref{thrm:main:2layerReLUgeneral} and \Cref{thrm:main:diagfeatgeneral}, respectively. Further, we show that the result for ReLU networks can be generalized to any activation function.

\paragraph{2-layer Network with Arbitrary Activation Function} Further, we show that the result for ReLU \Cref{ass:relu} generalizes to a $2$-layer network with features drawn uniformly and any activation function, which we define in \Cref{ass:2layer}.

\begin{model}[$2$-layer Network]\label{ass:2layer}
    Consider a $2$-layer network with an activation function $\sigma$ that is bounded on $[-1,1]$. Let the distribution $\cD$ be $\vx \sim \text{Unif}(\mathbb S^{d-1})$. The first layer parameters $\vu_i$ are initialized randomly isotropically, i.e. $\vu_i\sim \text{Unif}(\mathbb S^{d-1})$. Let $\{\sigma_i\}_{i=1}^{\infty}$ be the nonzero coefficients of the activation function $\sigma(\langle \vx, \vt\rangle)$ in the basis of spherical harmonics, i.e. $\sigma(\langle \vx, \vt \rangle) = \sum_{i,k} \sigma_i \phi_{i,k}(\vx) \phi_{i,k}(\vt)$, where the sum goes over only the nonzero $\sigma_i$. Here $\phi_{i,k}$ is the $i$-th spherical harmonic of order $k$ and is also the eigenbasis of the induced kernel~(\Cref{eq:inducedkernel})\footnote{The existence of such a decomposition follows from the Funk-Hecke formula}. 
\end{model}


\begin{theorem}[Weak-to-Strong Generalization with 2-layer NN]\label{thrm:main:2layerNNgeneral}
    Under \Cref{ass:2layer}, let \Cref{ass:fstar} be satisfied for groundturth $f^*$, population kernel $\cK$ (\Cref{eq:inducedkernel}) and positive integer $K$. If $\MT\ge K$ and the teacher attains the minimum loss~(\Cref{ass:teacher_minimizer}), then  for the student trained to stopping time $T$, with probability at least $1- \frac{2}{\MT}-K\exp(-\frac{c}{2}\frac{\MT}{K})$ over the randomness of teacher features, it holds that
    \begin{align*}
        \SL 
        \le 
\inf_{S\ge K} \Biggl\{
    \frac{ 4(\sum_{k=1}^{S} \sigma_k^2)(\sum_{k=S+1}^{\infty} \sigma_k^2)  \log^2 \MT}{\sigma_{S}^4 \MT}\,\TL
    + \sigma_{S+1}^4T^2 \,\TL\Biggr\}
    + \frac{e^{-\sigma_{K}^2T}}{2-e^{-\sigma_{K}^2T}} \,\bigl\|f^*\bigr\|_{\cD}^2.
    \end{align*}
\end{theorem}
 As in \Cref{thm:general-main}, we can take the stopping time $T=\frac{1}{\sigma_K^2} \log \frac{1}{\delta_T}$ if there is a large jump in $\sigma_{i}$. For the proof of this theorem, see \Cref{app:2layer}.

\paragraph{2-layer ReLU Networks.}As a corollary, we have the bound for ReLU networks. For ReLU activation in \Cref{ass:relu}, the expansion of $\sigma$ in the basis of spherical harmonics $\phi_{k,i}$ will be $(\sigma_1,\dots,\sigma_1,\sigma_2,\dots,\sigma_2,\dots)$ where each $\sigma_i$ repeats $N_k =  \frac{(2k+d-2)(k+d-3)!}{k!(d-2)!}$ times, which is the order of the $k$-th spherical harmonics in $d$ dimensions. Furthermore, $\sigma_k$ is nonzero only for even $k$ and $k=1$. Therefore, \Cref{ass:fstar} in this case amounts to $f^*$ being a sum of a linear function and an even polynomial of order at most $k$, where $K=N_0+\dots+N_k$.  




\begin{corollary}[Weak-to-Strong Generalization with 2-layer ReLU NN]\label{thrm:main:2layerReLUgeneral}
    Under \Cref{ass:relu}, let \Cref{ass:fstar} be satisfied for $f^*$ for some $K=N_0+\dots+N_k$ and let $k$ be the corresponding degree. If the teacher attains minimum loss (\Cref{ass:teacher_minimizer}), then for the student trained until stopping time $T$ will have with probability at least $1-\frac{2}{\MT}-K\exp(-c\frac{\MT}{K})$, 
    \begin{align*}
        \SL 
        \le 
\inf_{s\ge K} \Biggl\{
    \frac{ 4(\sum_{k=1}^{s} \sigma_k^2N_k)(\sum_{k=s+1}^{\infty} \sigma_k^2N_k)  \log^2 \MT}{\sigma_{s}^4 \MT}\,\TL
    + \sigma_{s+2}^4T^2 \,\TL\Biggr\}
    + \frac{e^{-\sigma_{k}^2T}}{2-e^{-\sigma_{k}^2T}} \,\bigl\|f^*\bigr\|_{\cD}^2.
    \end{align*}
    
    Furthermore, for any $\delta_T \in (0,1)$, if the stopping time is $T=\frac{1}{\sigma_k^2} \log \frac{1}{\delta_T}$, we will have with probability at least $1-\frac{2}{\MT}-K\exp(-c\frac{\MT}{K})$, 
    \begin{align*}
        \SL\le O_d\left(\frac{1}{\sqrt{\MT}} \left(\log \MT \right)^2 \right) \cdot \TL
        +O_d\left(\frac{1}{\sigma_k^4}\frac{1}{\sqrt{\MT}}\right)\cdot \left(\log \frac{1}{\delta_T} \right)^2 \TL+\frac{\delta_T}{2-\delta_T}\|f^*\|_{\cD}^2.
    \end{align*}
    Here $c$ is an absolute constant.
\end{corollary}
For the proof of this theorem, see \Cref{app:sub:2layerReLU}.

\paragraph{Linear Networks}

Finally, we can state the general bound for linear networks, \Cref{ass:linearnetwork}. 

\begin{theorem}[Weak-to-Strong Generalization in Linear Networks]\label{thrm:main:diagfeatgeneral}
    For \Cref{ass:linearnetwork}, if $f^*$ is spanned by the first $K$ coordinates and if the teacher attains minimum loss $\TL=\hat{\cL}_{\min}$, then for the student trained until time $T$ we have with probability at least $1-\frac{4}{m}$ that
    \begin{align*}
        \SL &\le 
\inf_{S \ge K} \Biggl\{
    \frac{\bigl(\sum_{i=1}^{S} \psi_i\bigr)\bigl(\sum_{i=S+1}^{d} \psi_i\bigr)}{\psi_S^2}
    \,\frac{(\log m)^2}{m}\,\TL
    + \psi_{S+1} \, T^2 \,\TL\Biggr\}
    + \frac{e^{-\psi_K T}}{2 - e^{-\psi_K T}} \,\bigl\|f^*\bigr\|_{\cD}^2.
    \end{align*}
\end{theorem}

The relevant feature variances here are given by the input distribution $\{\psi_i\}_{i=1}^{d}$, and we use those to select the stopping time, similarly to \Cref{thrm:main:2layerReLUgeneral}. For the proof, see \Cref{app:diagfeat}.

\subsection{Teacher-Student Feature Alignment in Random Feature Networks.}

In this section, we isolate the results on the random feature models (\Cref{ass:relu} and \Cref{ass:linearnetwork}) from the specific weak-to-strong setups that we consider. We prove bounds on the teacher-student feature alignment quantity defined in \Cref{eq:def-kappa}, which hold for these random feature models more generally, i.e. independently of the weak-to-strong setup. \Cref{thrm:main:feature-misalignment} is the general bound on teacher-student feature misalignment $\kappa_S$ and \Cref{cor:main:misalignment-linear} and \Cref{cor:main:misalignment-relu} give the specific bounds on teacher-student feature misalignment in linear and relu random feature networks.


\begin{lemma}[General Bound on the Teacher-Student Feature Misalignment]\label{thrm:main:feature-misalignment}
    If for the random feature distribution $\cG$ it holds that there exists $B>0$ such that for all $q\in \mathbb N, q\ge 2$, we have that $\mathbb E_{g\sim \cG}\left( \|P_{\le J} g\|_{\cD}^{q}\|P_{\ge J+1} g\|_{\cD}^{q}\right)\le \frac{1}{2} q!B^{q}$, then for $\kappa_S$ defined in \Cref{eq:def-kappa} with probability at least $1-2\delta$ it holds that
    \begin{align*}
        \frac{1}{\kappa_S}-1 \le \frac{mB \left( \log \frac{1}{\delta}\right)^2}{\lambda^{\mA}_{S}},
    \end{align*}
    where $\lambda^{\mA}_{S}$ is the $S$-th largest eigenvalue of $\mA$.
\end{lemma}

Applying \Cref{thrm:main:feature-misalignment} to the particular distributions of ReLU Random Features in \Cref{ass:relu} and Linear Networks in \Cref{ass:linearnetwork}, we get the following bounds on the teacher-student feature misalignment.

\begin{corollary}[Bound on the Teacher-Student Feature Misalignment in Random Linear Networks]\label{cor:main:misalignment-linear}
    For Linear Network \Cref{ass:linearnetwork}, with probability at least $1-2\delta-2\exp(-c_1\frac{1}{\psi_1^4}t_A^2 )$ we have that 
     \begin{align*}
          \frac{1}{\kappa_S}-1\le \frac{8m\left(\sum_{i=1}^{S}\psi_i\right)\left(\sum_{i=S+1}^{\infty}\psi_i\right)\left(\log \frac{1}{\delta}\right)^2}{\psi_S ^2\left( \sqrt{m}-C\sqrt{S}-t_A\right)^4}
     \end{align*}
     where $c_1$ and $C$ are absolute constants.
     \removed{Setting $t_A = \psi_1^2 \log m$, $\delta=\frac{1}{m}, \delta_1 =\frac{1}{m} $, we get that with probability at least $1-\frac{4}{m}$ we have 
     \begin{align*}
         \frac{1}{\kappa_S}-1 \le  \frac{8m\left(\sum_{i=1}^{S}\psi_i\right)\left(\sum_{i=S+1}^{\infty}\psi_i\right)\left(\log m\right)^2}{\psi_S^2 \left( \sqrt{m}-C\sqrt{S}-\psi_1^2\log m\right)^4}.
     \end{align*}}
\end{corollary}

\begin{corollary}[Bound on the Teacher-Student Feature Misalignment in Random ReLU Network]\label{cor:main:misalignment-relu}
    Under ReLU Network \Cref{ass:relu}, if \Cref{ass:fstar} is satisfied for $f^*$ and $K$, let $s$ be such that $N_0+\dots+N_{s-1}+1\le K \le N_0+\dots+N_s$. Then with probability at least $1-2\delta - 2K \exp(-ct_A^2)$
     \begin{align*}
          \frac{1}{\kappa_S}-1\le  \frac{ m (\sum_{k=1}^{s} \sigma_k^2N_k)(\sum_{k=s}^{\infty} \sigma_k^2 N_k)  \left(\log \frac{1}{\delta}\right)^2}{\sigma_{s}^4 \left( \sqrt{m}-t_A \sqrt{K}\right)^4}
     \end{align*}
     where $c$ is an absolute constant and $\delta \in (0,1)$. 
      \removed{We can take $s=m^{\frac{1}{4(d-4)}}+o_m(1)$ ($s$ is closest even integer to $m^{\frac{1}{4(d-4)}}$) so that 
    \begin{align*}
        \frac{1}{\kappa_S}-1 &\le \Theta_d\left( \frac{ m \sqrt{m} \left(\log \frac{1}{\delta}\right)^2}{ \left( \sqrt{m}-t_A \sqrt{K}\right)^4} \right) \\
    \end{align*}
    Note that this $\Theta$ hides absolute constants and $d$ dependence.}
\end{corollary}

\Cref{cor:main:misalignment-linear} and \Cref{cor:main:misalignment-relu} help specify the general weak-to-strong multiplicative improvement bound \Cref{thm:general-main}, which is a deterministic claim that holds for both random and non-random teachers, to the weak-to-strong bounds for ReLU and linear networks, namely \Cref{thrm:main:2layerReLUgeneral} and \Cref{thrm:main:diagfeatgeneral}. These results on random feature models may be of independent interest.  For the proofs, see \Cref{thrm:app:kappalowerbound2layer}.

\section{Related work}

\paragraph{Theoretical Understanding of Weak-to-Strong Generalization.} \citet{lang2024theoretical} propose a theoretical framework that establishes weak-to-strong generalization for classification when the strong student is assumed to learn functions that are robust to mislabeled data among many correctly labeled neighbors. \citet{shin2024datacentric} use this framework to formulate a mechanism for weak-to-strong generalization for data with easy and hard patterns. \citet{charikar2024quantifying} study a general regression setup with squared loss, where a key example is training only the last linear layer of the teacher and student networks. They show that if the strong student's capacity is not large enough to express the weak teacher's function, then the student can outperform the teacher by an amount quantified by how much the student does not fit the teacher's labels. \citet{mulgund2025relating} further generalize this result to a broader class of loss functions, including the cross entropy loss. \citet{wu2024benign} demonstrated that weak-to-strong generalization can happen in an overparametrized spiked covariance model for classification. \cite{ildiz2024highdim} show that in a two-stage learning process in high-dimensional ridgeless regression, a student trained on teacher labels can be better than a student trained on real labels (which is an effect of "distillation" rather than "weak-to-strong").

 \citet{charikar2024quantifying} and \citet{wu2024benign} consider linear models that are most closely related to our setup. \citet{charikar2024quantifying} consider a general regression setup with square loss and \cite{wu2024benign} consider classification with a minimum norm interpolator. In the setup of \citet{charikar2024quantifying}, if the teacher and student classes contain the origin, our lower bound applies to their setting.

\paragraph{Empirical Work.} Following up on \citet{burns2024weaktostrong}, many works explored ways to leverage weak models to improve the performance of strong models. \citet{shin2024datacentric, li2024superfiltering} propose data selection methods to improve W2S generalization. \citet{somerstep2024transfer} propose to use in-context learning to refine weak teacher labels. \citet{yang2024reasoning,bansal2024computeoptimal} show that small models can be effectively used to generate reasoning data with Chain-of-Thought (CoT) for supervising large models - \citet{bansal2024computeoptimal} point out that this can be more compute-optimal as compared to generating data from large models. \citet{sun2024easy} showed that a weak model can be trained as a reward model on easy tasks and then used to guide the training of a strong model on harder tasks. \citet{ji2024aligner} proposed a plug-and-play module for alignment, where a small model learns to redistribute the output of a large model.
\citet{tao2025your} systematically evaluated the performance of aligning a large model with feedback from a small model instead of human feedback. \citet{yang2024deceive} highlighted that the strong model can deceive the weak model by being well aligned only in the areas familiar to the weak model.

\section{Conclusion}

We prove that weak-to-strong generalization can happen even in a simple model of two-layer neural networks with random features. We show that in this case, the PGR (Performance Gap Recovered) can converge to $1$ when teacher loss goes to $0$. Additionally, we show an even stronger separation between the teacher and the student, namely that the student error is polynomially smaller than the teacher error. For the ReLU network, we show that the exponent in this polynomial dependence is at least $1.49$ while for the linear networks, we show that it is $2$. Further, we explain that in his setup, early stopping is crucial for weak-to-strong generalization to occur. We also explain that in this setup, the weak-to-strong improvement comes from the student zeroing out the teacher signal in the high-order eigendirections (which in our setup are noise directions under the condition that the ground truth is spanned by the first few eigendirections). Most interestingly, we show that there is an inherent limitation to weak-to-strong generalization in random feature models trained with gradient descent or gradient flow, and even more generally. Namely, we show that if the target is normalized, the student error can be at most quadratically smaller than the teacher error. Our quadratic lower bound asserts that the student loss cannot be smaller than the square of teacher loss, regardless of the choice of student feature, the early stopping time, and even the number of rounds of bootstraping (i.e., even with multiple students).




\section*{Acknowledgements}
ZL and SA are supported by OpenAI superalignment grant. 
SA is also supported by NSF, DARPA, ONR.
MM would like to thank Theodor Misiakiewicz for providing the reference \citep{defilippis2024} and for a useful discussion about it.


\bibliography{bibliography}

@inproceedings{burns2024weaktostrong,
title={Weak-to-Strong Generalization: Eliciting Strong Capabilities With Weak Supervision},
author={Collin Burns and Pavel Izmailov and Jan Hendrik Kirchner and Bowen Baker and Leo Gao and Leopold Aschenbrenner and Yining Chen and Adrien Ecoffet and Manas Joglekar and Jan Leike and Ilya Sutskever and Jeffrey Wu},
booktitle={Forty-first International Conference on Machine Learning},
year={2024},
url={https://openreview.net/forum?id=ghNRg2mEgN}
}

@article{mei2021concentration,
title = {Generalization error of random feature and kernel methods: Hypercontractivity and kernel matrix concentration},
journal = {Applied and Computational Harmonic Analysis},
volume = {59},
pages = {3-84},
year = {2022},
note = {Special Issue on Harmonic Analysis and Machine Learning},
issn = {1063-5203},
doi = {https://doi.org/10.1016/j.acha.2021.12.003},
url = {https://www.sciencedirect.com/science/article/pii/S1063520321001044},
author = {Song Mei and Theodor Misiakiewicz and Andrea Montanari}
}

@article{petrini2022,
    author ={Leonardo Petrini and Francesco Cagnetta and Eric Vanden-Eijnden and Matthieu Wyart},
    title = {Learning sparse features can lead to overfitting in neural networks},
    journal = {36th Conference on Neural Information Processing Systems (NeurIPS 2022)},
    year = {2022}
}

@inbook{vershynin2011, place={Cambridge}, title={Introduction to the non-asymptotic analysis of random matrices}, booktitle={Compressed Sensing: Theory and Applications}, publisher={Cambridge University Press}, author={Vershynin, Roman}, year={2012}, pages={210–268}}

@article{rahimi2007,
    author = {Ali Rahimi and Benjamin Recht},
    title = {Random Features for Large-Scale Kernel Machines},
    journal = {Advances in Neural Information Processing Systems 20 (NIPS 2007)},
    year = {2007}
}

@article{vempala2006proj,
    author = {Rosa I Arriaga and Santosh Vempala},
    title = {An algorithmic theory of learning: Robust concepts and random projection},
    journal = {Machine Learning},
volume = {63-2},
pages = {161-182},
    year = {2006}
}

@article{paquette2024scaling,
    author = {Elliot Paquette and Courtney Paquette and Lechao Xiao and Jeffrey Pennington},
    title = {4+3 Phases of Compute-Optimal Neural Scaling Laws},
    journal = {38th Conference on Neural Information Processing Systems (NeurIPS 2024)},
    year = {2024} 
}

@article{lee2024scaling,
    author = {Licong Lin and Jingfeng Wu and Sham M. Kakade and Peter L. Bartlett and Jason D. Lee},
    title = {Scaling Laws in Linear Regression: Compute, Parameters, and Data},
    journal = {38th Conference on Neural Information Processing Systems (NeurIPS 2024)},
    year = {2024}
}

@article{pinelis1994,
    author = {Iosif Pinelis},
    title = {Optimum Bounds for the Distributions of Martingales in Banach Spaces},
    journal = {Ann. Probab. 22(4)},
    pages = {1679-1706},
    year = {1994}
}

@inproceedings{shin2024datacentric,
title={Weak-to-Strong Generalization Through the Data-Centric Lens},
author={Changho Shin and John Cooper and Frederic Sala},
booktitle={The Thirteenth International Conference on Learning Representations},
year={2025},
}

@inproceedings{yang2024reasoning,
    title = "Weak-to-Strong Reasoning",
    author = "Yang, Yuqing  and
      Ma, Yan  and
      Liu, Pengfei",
    editor = "Al-Onaizan, Yaser  and
      Bansal, Mohit  and
      Chen, Yun-Nung",
    booktitle = "Findings of the Association for Computational Linguistics: EMNLP 2024",
    month = nov,
    year = "2024",
    address = "Miami, Florida, USA",
    publisher = "Association for Computational Linguistics",
    doi = "10.18653/v1/2024.findings-emnlp.490",
    pages = "8350--8367"
}

@article{bansal2024computeoptimal,
    author = {Hritik Bansal and Arian Hosseini and Rishabh Agarwal and Vinh Q. Tran and Mehran Kazemi},
    title = {Smaller, Weaker, Yet Better: Training LLM Reasoners via Compute-Optimal Sampling},
    journal = {The Thirteenth International Conference on Learning Representations},
    year = {2025}
}

@article{ildiz2024highdim,
    author = {M. Emrullah Ildiz and Halil Alperen Gozeten and Ege Onur Taga and Marco Mondelli and Samet Oymak},
    title = {High-dimensional Analysis of Knowledge Distillation: Weak-to-Strong Generalization and Scaling Laws},
    journal = {The Thirteenth International Conference on Learning Representations},
    year = {2025}
}

@inproceedings{yang2024deceive,
    title={Super(ficial)-alignment: Strong Models May Deceive Weak Models in Weak-to-Strong Generalization},
    author={Wenkai Yang and Shiqi Shen and Guangyao Shen and Wei Yao and Yong Liu and Gong Zhi and Yankai Lin and Ji-Rong Wen},
    booktitle={The Thirteenth International Conference on Learning Representations},
    year={2025},
    url={https://openreview.net/forum?id=HxKSzulSD1}
}

@inproceedings{tao2025your,
title={Your Weak {LLM} is Secretly a Strong Teacher for Alignment},
author={Leitian Tao and Yixuan Li},
booktitle={The Thirteenth International Conference on Learning Representations},
year={2025},
}

@inproceedings{charikar2024quantifying,
 author = {Charikar, Moses and Pabbaraju, Chirag and Shiragur, Kirankumar},
 booktitle = {Advances in Neural Information Processing Systems},
 editor = {A. Globerson and L. Mackey and D. Belgrave and A. Fan and U. Paquet and J. Tomczak and C. Zhang},
 pages = {126474--126499},
 publisher = {Curran Associates, Inc.},
 title = {Quantifying the Gain in Weak-to-Strong Generalization},
 volume = {37},
 year = {2024}
}

@inproceedings{wu2024benign,
title={Provable weak-to-strong generalization via benign overfitting},
author={David Xing Wu and Anant Sahai},
booktitle={The Thirteenth International Conference on Learning Representations},
year={2025},
}

@article{defilippis2024,
    author = {Leonardo Defilippis and Bruno Loureiro and Theodor Misiakiewicz},
    title ={Dimension-free deterministic equivalents and scaling laws for random feature regression},
    journal = {38th Conference on Neural Information Processing Systems (NeurIPS 2024).},
    year = {2024}
}

@article{radford2019language,
  title={Language models are unsupervised multitask learners},
  author={Radford, Alec and Wu, Jeffrey and Child, Rewon and Luan, David and Amodei, Dario and Sutskever, Ilya and others},
  journal={OpenAI blog},
  volume={1},
  number={8},
  pages={9},
  year={2019}
}

@article{achiam2023gpt,
  title={{GPT}-4 technical report},
  author={Achiam, Josh and Adler, Steven and Agarwal, Sandhini and Ahmad, Lama and Akkaya, Ilge and Aleman, Florencia Leoni and Almeida, Diogo and Altenschmidt, Janko and Altman, Sam and Anadkat, Shyamal and others},
  journal={arXiv preprint arXiv:2303.08774},
  year={2023}
}

@article{simon2021,
    author = { Simon, James B. and Dickens, Madeline and Karkada, Dhruva and DeWeese, Michael R.},
    title ={The Eigenlearning Framework: A Conservation Law Perspective on Kernel Regression and Wide Neural Networks},
    journal = {Transactions on Machine Learning Research},
    year = {2023}
}

@article{canatar2021spectral,
    title={Spectral bias and task-model alignment explain generalization in kernel regression and infinitely wide neural networks},
    author={Abdulkadir Canatar and Blake Bordelon and Cengiz Pehlevan},
    volume = {12},
    number = {1},
    pages = {1-12},
    year = {2021},
    journal = {Nature Communications}
}

@inproceedings{wei:2022-more-than-a-toy,
  author    = {Alexander Wei and
               Wei Hu and
               Jacob Steinhardt},
  title     = {More Than a Toy: Random Matrix Models Predict How Real-World Neural
               Representations Generalize},
  booktitle = {International Conference on Machine Learning},
  series    = {Proceedings of Machine Learning Research},
  year      = {2022},
}

@article{misiak2024,
  author       = {Theodor Misiakiewicz and
                  Basil Saeed},
  title        = {A non-asymptotic theory of Kernel Ridge Regression: deterministic
                  equivalents, test error, and {GCV} estimator},
  journal      = {CoRR},
  volume       = {abs/2403.08938},
  year         = {2024},
  url          = {https://doi.org/10.48550/arXiv.2403.08938},
  doi          = {10.48550/ARXIV.2403.08938},
  eprinttype    = {arXiv},
  eprint       = {2403.08938},
  bibsource    = {dblp computer science bibliography, https://dblp.org}
}

@inproceedings{
lang2024theoretical,
title={Theoretical Analysis of Weak-to-Strong Generalization},
author={Hunter Lang and David Sontag and Aravindan Vijayaraghavan},
booktitle={The Thirty-eighth Annual Conference on Neural Information Processing Systems},
year={2024}
}

@inproceedings{somerstep2024transfer,
title={A transfer learning framework for weak to strong generalization},
author={Seamus Somerstep and Felipe Maia Polo and Moulinath Banerjee and Yaacov Ritov and Mikhail Yurochkin and Yuekai Sun},
booktitle={The Thirteenth International Conference on Learning Representations},
year={2025},
}

@inproceedings{ji2024aligner,
 author = {Ji, Jiaming and Chen, Boyuan and Lou, Hantao and Hong, Donghai and Zhang, Borong and Pan, Xuehai and Qiu, Tianyi (Alex) and Dai, Juntao and Yang, Yaodong},
 booktitle = {Advances in Neural Information Processing Systems},
 editor = {A. Globerson and L. Mackey and D. Belgrave and A. Fan and U. Paquet and J. Tomczak and C. Zhang},
 pages = {90853--90890},
 publisher = {Curran Associates, Inc.},
 title = {Aligner: Efficient Alignment by Learning to Correct},
 volume = {37},
 year = {2024}
}

@inproceedings{sun2024easy,
 author = {Sun, Zhiqing and Yu, Longhui and Shen, Yikang and Liu, Weiyang and Yang, Yiming and Welleck, Sean and Gan, Chuang},
 booktitle = {Advances in Neural Information Processing Systems},
 editor = {A. Globerson and L. Mackey and D. Belgrave and A. Fan and U. Paquet and J. Tomczak and C. Zhang},
 pages = {51118--51168},
 publisher = {Curran Associates, Inc.},
 title = {Easy-to-Hard Generalization: Scalable Alignment Beyond Human Supervision},
 volume = {37},
 year = {2024}
}

@article{mulgund2025relating,
  title={Relating Misfit to Gain in Weak-to-Strong Generalization Beyond the Squared Loss},
  author={Mulgund, Abhijeet and Pabbaraju, Chirag},
  journal={arXiv preprint arXiv:2501.19105},
  year={2025}
}

@inproceedings{li2024superfiltering,
    title = "Superfiltering: Weak-to-Strong Data Filtering for Fast Instruction-Tuning",
    author = "Li, Ming  and
      Zhang, Yong  and
      He, Shwai  and
      Li, Zhitao  and
      Zhao, Hongyu  and
      Wang, Jianzong  and
      Cheng, Ning  and
      Zhou, Tianyi",
    editor = "Ku, Lun-Wei  and
      Martins, Andre  and
      Srikumar, Vivek",
    booktitle = "Proceedings of the 62nd Annual Meeting of the Association for Computational Linguistics (Volume 1: Long Papers)",
    month = aug,
    year = "2024",
    address = "Bangkok, Thailand",
    publisher = "Association for Computational Linguistics",
    doi = "10.18653/v1/2024.acl-long.769",
    pages = "14255--14273",
}

@inproceedings{alnur2019earlystopping,
    author = {Alnur Ali and J. Zico Kolter and Ryan J. Tibshirani},
    title = {A Continuous-Time View of Early Stopping for Least Squares},
    booktitle ={Proceedings of the 22nd International Conference on Artificial Intelligence and Statistics (AISTATS) 2019, Naha, Okinawa, Japan},
    year = {2019}
}

@inproceedings{elkabetz2021optimization, author = {Elkabetz, Omer and Cohen, Nadav}, title = {Continuous vs. discrete optimization of deep neural networks}, year = {2021}, isbn = {9781713845393}, publisher = {Curran Associates Inc.}, address = {Red Hook, NY, USA}, booktitle = {Proceedings of the 35th International Conference on Neural Information Processing Systems}, articleno = {378}, numpages = {14}, series = {NIPS '21} }

@ARTICLE{sauel59studies,
  author={Samuel, A. L.},
  journal={IBM Journal of Research and Development}, 
  title={Some Studies in Machine Learning Using the Game of Checkers}, 
  year={1959},
  volume={3},
  number={3},
  pages={210-229},
  doi={10.1147/rd.33.0210}}
\bibliographystyle{icml2025}

\newpage
\appendix
\onecolumn
\section*{Appendix}

\tableofcontents

\section{Gradient Flow Dynamics}\label{app:gradflow}

We first introduce the function space view of gradient flow over the top layer of the student's network \Cref{eq:gradflow}. The main benefit of working in function space from parameter space is allowing us to analyze the infinite-width limit in the same space. We first recall we parametrize the student as:
\begin{align*}
    f_{\text{student}} = f_{\vw_{\mathrm{s}},\vu_{\mathrm{s}}}(\vx)=\sum_{i=1}^{\MS} w_{\mathrm{s},i} \sigma(\langle \vu_{\mathrm{s},i},\vx\rangle).
\end{align*}
\removed{\marko{fix this part.}
In the limit $\MS\to \infty$ we desrcibe the student's predictor with a density $\rho$ over the parameters $\vu_{\mathrm{s}} \in \R^d$
\begin{align}\label{eq:meanfieldpredictor}
    f_{\rho}(\vx) = \int \sigma(\langle \vu_{\mathrm{s}} , \vx \rangle) \rho(\dd \vu_{\mathrm{s}})
\end{align}}
Let $\sigma_{\mathrm{s},i}(\vx) = \sigma(\langle \vu_{\mathrm{s},i},\vx\rangle)$. Then, \Cref{eq:gradflow} in function space is
\begin{align}\label{eq:app:dynamicsfuncspace}
\begin{aligned}
    \frac{\dd}{\dd t} f_{\vw_{\mathrm{s}}(t),\vu_{\mathrm{s}}}(\vx)
    =&
    -\frac{1}{\MS}\sum_{i=1}^{\MS}\mathbb{E}_{\vy\sim \cD}\inne{f_{\vw_{\mathrm{s}}(t),\vu_{\mathrm{s}}}(\vy) - f_{\text{teacher}}(\vy)}{\sigma_{\mathrm{s},i}(\vy)}~\sigma_{\mathrm{s},i}(\vx)\\
    =&-\mathbb{E}_{\vy \sim \cD}\left[K_{\MS}(\vx, \vy) (f_{\vw_{\mathrm{s}}(t),\vu_{\mathrm{s}}}(\vy) - f_{\text{teacher}}(\vy))\right]
    \end{aligned}
\end{align}
where $K_{\MS}(\vx, \vy) = \frac{1}{\MS}\sum_{i=1}^{\MS}\sigma_{\mathrm{s},i}(\vx)\sigma_{\mathrm{s},i}(\vy)$ is the empirical kernel.

Taking $\MS\to \infty$, the solution of \Cref{eq:app:dynamicsfuncspace} converges to the solution of the following equation\removed{, with empirical average replaced by expectation and student parameters $(\vw_{\mathrm{s}}(t),\vu_{\mathrm{s}}(t))$ replaced by the density over the parameter space $\rho(t)$}:
\begin{align}\label{eq:app:kerneldynamics}
\begin{aligned}
    \frac{\dd}{\dd t} f_{t}(\vx)
    = -\E_{\vy\sim\cD}\left[\gK(\vx,\vy)\left( f_{t}(\vy)-f_{\text{teacher}}(\vy) \right) \right],
    \end{aligned}
\end{align}
where
\begin{align}\label{eq:app:inducedkernel}
\begin{aligned}
    \gK(\vx, \vx') := \E_{\vu\sim\text{unif}(\mathbb S^{d-1})}\left[\sigma(\langle \vu,\vx\rangle)\sigma(\langle \vu,\vx' \rangle) \right].
    \end{aligned}
\end{align}


  The kernel $\gK$ can be decomposed as $\gK(\vx, \vx') = \sum_{k \ge 1} \lambda_k e_k(\vx) e_k(\vx')$,
where $\lambda_1 \ge \lambda_2 \ge \lambda_3 \ge\cdots$ are the eigenvalues of the associated kernel operator in descending order, and $e_1(\vx), e_2(\vx), e_3(\vx), \dots $ are orthonormal eigenfunctions, that is, $\inne{e_i}{e_j}_\cD = \delta_{ij}$. This allows us to decompose the gradient flow dynamics~(\Cref{eq:app:dynamicsfuncspace}) into simple ODEs for each eigendirection, which can be easily solved. With this decomposition, we can derive the following closed-form expression for the gradient flow dynamics of the student~(\Cref{eq:gradflow}) at time $t$: 
\begin{equation} \label{eq:app:student-dynamics}
    f_{t} = \sum_{k \ge 1} \left(1 - e^{-\lambda_k t}\right) \inne{f_{\text{teacher}}}{e_k}_{\cD}e_k
\end{equation}

\begin{theorem}[Closed Form Solution to Gradient Flow Dynamics]\label{thrm:app:kernelflow}
    Let $\cK:\mathcal{X}\times\mathcal{X}\to \mathcal{R}$ be a p.s.d. kernel with eigendecomposition $\cK(\vx,\vx')=\sum_{i=1}^{\infty} \lambda_i e_i(\vx)e_i(\vt)$, where $\inne{e_i}{e_j}_\cD =\delta_{ij}$. For any $f^*:\mathcal{X}\to \mathbb{R}$,  the equation of gradient flow for a predictor $f_{t}$
\begin{align}\label{eq:app:flowdiffeq}
\begin{aligned}
    &\frac{\dd}{\dd t} f_{t}(\vx)
    =-\E_{\vy\sim\cD}\left[\gK(\vx,\vy)\left( f_{t}(\vy)-f^*(\vy) \right) \right],
    \end{aligned}
\end{align}
with initial condition given by $f_0=0$ has a closed form solution $f_{t}$
\begin{equation*} 
    f_{t} = \sum_{k \ge 1} \left(1 - e^{-\lambda_k t}\right) \inne{f^*}{e_k}_{\cD}e_k.
\end{equation*}
\end{theorem}
\begin{proof}[Proof of \Cref{thrm:app:kernelflow}]
    To solve the gradient flow equation in \eqref{eq:app:flowdiffeq}, we project it onto the eigenbasis of the kernel operator $\cK$. Let $\beta_t^i = \langle f_t, e_i \rangle_{\cD}$ be the coefficients of $f_t$.
        
    Taking the inner product of both sides of \eqref{eq:app:flowdiffeq} with $e_j$:
    \begin{align*}
        \frac{d}{dt}\beta_t^j &= -\mathbb{E}_{y \sim \cD}\left[\left\langle \gK(\cdot, y), e_j \right\rangle_{\cD}(f_t(y) - f^*(y))\right] \\
        &= -\lambda_j \langle e_j, f_t - f^* \rangle_{\cD} \\
        &= -\lambda_j (\beta_t^j - \langle f^*, e_j \rangle_{\cD})
    \end{align*}
    Solving this ODE with initial condition $\beta_0^j = 0$:
    \begin{align*}
        \beta_t^j = (1 - e^{-\lambda_j t})\langle f^*, e_j \rangle_{\cD}.
    \end{align*}
    Therefore,
    \begin{align*}
        f_t = T^{\mathcal{K}}_t f^* = \sum_{j=1}^{\infty}(1 - e^{-\lambda_j t})\langle f^*, e_j \rangle_{\cD} e_j,
    \end{align*}
    which concludes the proof.
\end{proof}

\section{Different Model Types}\label{app:modeltypes}
We will first introduce a model that is slightly more general than \Cref{ass:linearnetwork}.

\begin{model}[Diagonal Covariate Features]\label{ass:app:daigfeat}
    Let $\{e_i\}_{i=1}^{\infty}$ be a fixed orthonormal basis of $L_2^{\cD}(\cX)$. Take the random features to be given by 
    \begin{align*}
        g = \sum_{i=1}^{\infty}\langle g,e_i\rangle_\cD e_i
    \end{align*}
    where $\{\langle g,e_i\rangle_{\cD}\}_{i=1}^{\infty} \sim N(0,\Lambda)$ where $\Lambda = (\lambda_1,\lambda_2,\dots)$ in decreasing order.
\end{model}

Our results will hold for \Cref{ass:app:daigfeat} and they are more general than \Cref{ass:linearnetwork}.

\begin{proposition}[Linear network as Diagonal Coavariate Features]\label{app:prop:modelequiv}
To get \Cref{ass:linearnetwork} from \Cref{ass:app:daigfeat}, we take $\Lambda = (\psi_1,\dots,\psi_d,0,0,\dots)$, $\langle x\mapsto x^\top u,  x\mapsto x^\top v\rangle_\cD = \sum_{i=1}^d \psi_iu_iv_i$, and $e_i(x) =  x_i /\sqrt{\psi_i}$. Thus if we set $g_i$ in \Cref{ass:app:daigfeat} as $g(x) = \vx^\top \vu_i$, where $\mU\sim N(0,I_d)$, then $\{\langle g,e_i \rangle_{\cD}\}_{i=1}^{\infty} \sim N(0,\Lambda)$.
    
\end{proposition}

\begin{proposition}[ReLU Network as Diagonal Covariate Features]\label{prop:app:reluasdiag}
    If in \Cref{ass:app:daigfeat} we take $\cD = \text{Unif}(\mathbb S^{d-1})$, $\{ e_i \}_{i=1}^{\infty}$ to be the spherical harmonics $\{ \{ \phi_{i,k}\}_{i=1}^{N_k} \}_{k=1}^{\infty}$, and choose 
    \begin{align*}
        \Lambda = (\sigma_1,\dots,\sigma_1,\sigma_2,\dots,\sigma_2,\dots,\sigma_k,\dots,\sigma_k,\dots),
    \end{align*} where $\sigma_k$ repeats $N_k$ times, we recover the same feature covariance as in \Cref{ass:relu}. Under Gaussian Universality assumptions, for this instance of \Cref{ass:app:daigfeat}, the teacher and the student risk behave the same as in \Cref{ass:relu}.
\end{proposition}

\begin{proof}[Proof of \Cref{prop:app:reluasdiag}]
    
    In \Cref{ass:relu}, the kernel induce by the random features is given by
    \begin{align*}
    \gK(\vx, \vx') := \E_{\vu\sim\text{unif}(\mathbb S^{d-1})}\left[\sigma(\langle \vu,\vx\rangle)\sigma(\langle \vu,\vx' \rangle) \right].
\end{align*}
A random feature in the ReLU case \Cref{ass:relu} is given by
\begin{align*}
    \sigma(\inne{\vw}{\vx}) = \sum_{k \ge 0} N_k \sigma_k P_{d,k}(\inne{\vw}{\vx}) = \sum_{k \ge 0} \sigma_k \inne{\vphi_{k}(\vw)}{\vphi_{k}(\vx)}=\sum_{k=0}^{\infty}\sum_{i=1}^{N_k} \sigma_k \phi_{i,k}(\vw) \phi_{i,k}(\vx).
\end{align*}

A random feature in the Diagonal Feature Covariance case \Cref{ass:app:daigfeat} is given by

\begin{align*}
    g(\vx) = \sum_{k=0}^{\infty} \sum_{i=1}^{N_k} g_{i,k} \phi_{i,k}(\vx).
\end{align*}

\begin{align*}
    \cK(\vx,\vx')=\E_{\vu}[\sigma(\inne{\vu}{\vx})\sigma(\inne{\vu}{\vx'})]
    = \E_{\vu}\left[ \sum_{k \ge 0} \sigma_k \inne{\vphi_k(\vu)}{\vphi_k(\vx)}  \cdot \sigma_k \inne{\vphi_k(\vu)}{\vphi_k(\vx')} \right]
    = \sum_{k \ge 0} \sigma_k^2 \inne{\vphi_k(\vx)}{\vphi_k(\vx')}.
\end{align*}
On the other hand, we have that for \Cref{ass:app:daigfeat} the kernel is
    \begin{align*}
        \cK(\vx, \vx') = \E_{g}\left[ g(\vx) g(\vx') \right] =  \E_{g}\left[ \sum_{i,k} \sum_{j,l} g_{i,k} \phi_{i,k}(\vx) g_{j,l} \phi_{j,l}(\vx) \right] = \sum_{k=0}^{\infty}\sigma_k^2 \langle \vphi_{k}(\vx), \vphi_{k}(\vx') \rangle.
    \end{align*}
\end{proof}

\subsection{Diagonalizability of the Feature Covariance in \Cref{ass:linearnetwork}}

In this section we establish the generality of \Cref{ass:linearnetwork}. Namely, we show that one can assume that $\vu$ is isotropic Gaussuian distribution and $\vx$ has diagonal covariance if and only if their covariance matrices are codiagonalizable.

\begin{proposition}[Diagonalizability of the Linear Network Covariance]\label{prop:app:linearnetworkgenerality}
    A linear network $\sigma(z) = z$ with bottom layer weights $\vu $ distributed accoding to a Gaussian distribution $\cU = N(0,\Psi_{\vu})$ and the inputs $\vx$ distributed according to $\cD = N(0,\Psi_{\vx})$ is equivalent to \Cref{ass:linearnetwork} if and only if $\Psi_{\vx}$ and $\Psi_{\vu}$ are codiagonalizable, with $\Psi = \mV^T\mS_{\vu}^{-1/2}\mV^T  \mS_{\vx} \mV\mS_{\vu}^{-1/2}\mV$ where $\mV$ is the shared basis.
\end{proposition}
\begin{proof}[Proof of \Cref{prop:app:linearnetworkgenerality}]
    Note that the distribution in the first case is given by $z\mid \vx \sim \langle \vu ,\vx\rangle\mid \vx \sim N(0,\vx \Sigma_u \vx^T)$. Note further that for \Cref{ass:linearnetwork}  with $\Psi$ we have that $\tilde z \mid \tilde \vx \sim \langle \tilde \vu , \tilde \vx \rangle \mid \tilde \vx \sim N(0,\Psi) $. Now if $\tilde \vx = \mA \vx$ for some $\mA$, then we have that $\Psi = cov(\tilde \vx) = cov(\mA \vx) = \mA^T \Sigma_x \mA = \mA^T \mV_x \mS_x \mV_x \mA $,  where $\mV_x $ is the basis of $\Sigma_x$, i.e. $\Sigma_x=\mV_x^T \mS_x \mV_x$ for $\mS_x$ diagonal. Since $\Psi$ is diagonal. So $\mA^T \mV_x \mS_x \mV_x \mA$ is diagonal if and only if $\mA = \mD \mV_x$ for $\mV_x$ a basis of $\Sigma_x$ and some diagonal matrix $\mD$. So the distributions of $z$ and $\tilde z$ are the same if and only if $\mA = \mD \mV_x$ and $\vx \Sigma_{\vu} \vx^T = \vx \mV_{x} \mD^2 \mV_x \vx^T$, so $\Sigma_{\vu} = \mV_{x} \mD^2 \mV_{x}$.
\end{proof}

\section{Proof of Weak-to-Strong Multiplicative Error Improvement}\label{app:proof:general-main}

Here we present the proof of \Cref{thm:general-main}. We will introduce the following notation. Let $\MT=m$ and let $\{e_i\}_{i=1}^{\infty}$ be the eigenbasis of the student's kernel $\cK$. Let the units, i.e. features, of the student be functions $g$ drawn from some distribution $\cG$. For example, in \Cref{ass:relu}, we gave that $\cG$ is given by $g(\vx) = \sigma( \langle \vu,\vx \rangle)$ where $\vu\sim \text{unif}(\mathbb S^{d-1})$. Let $\vw\in \R^m$ be the teacher's weights, and let the teacher be parametrized as $f_{\text{teacher}} = f_{w} = \sum_{i=1}^{m} w_i g_i$, where $\{g_i\}_{i=1}^{m}$ are teacher's units, i.e. features. We will refer to $\hat{f}_{w}$ the optimally trained teacher. Let the student's predictor at time $T$ be $f_T$. Let $\cH$ be the Hilbert space of square-integrable functions with respect to $\cD$. Let $\mPhi \in \R^{m \times m}$ be the Gram matrix of random units, {\it i.e.}, $\Phi_{ij} := \inne{\sigma_{i}}{\sigma_j}_{\cD} = \sum_{k \ge 1}\inne{\sigma_i}{e_k}_{\cD} \inne{\sigma_j}{e_k}_{\cD}$.

\subsection{Preliminary Lemmas}
\begin{lemma}\label{lem:eigen_func}
For all $k, k' \ge 1$,
\begin{align*}
    \E_{g \sim \cG}[\inne{g}{e_{k}}_{\cD} \inne{g}{e_{k'}}_{\cD}] = \begin{cases}
        \lambda_k & \quad \text{if } k = k', \\
        0 & \quad \text{otherwise}.
    \end{cases}
\end{align*}
\end{lemma}

\begin{lemma} \label{lm:cbDelta}
    For $c \in [0, 1)$,
    \begin{align*}
        ((1-c)(b+\Delta)-b)^2 \le \Delta^2 + \frac{c}{2-c}b^2.
    \end{align*}
\end{lemma}
\begin{proof}[Proof of \Cref{lm:cbDelta}]
    \begin{align*}
        ((1-c)(b+\Delta)-b)^2
        = ((1-c)\Delta - cb)^2 
        &= \left((1-c)^2 \cdot \frac{1}{1-c} \Delta + \left(2c - c^2\right) \cdot \frac{1}{2c-c^2} (-cb)\right)^2 \\
        &\le (1-c)^2 \cdot \frac{1}{(1-c)^2} \Delta^2 + \left(2c - c^2\right) \cdot \frac{b^2}{(2-c)^2} \\
        &= \Delta^2 + \frac{c}{2-c} b^2,
    \end{align*}
    where the inequality follows from Jensen's inequality.
\end{proof}

The following is a standard result on the generalized Rayleigh quotient.
\begin{lemma}\label{lm:max-rayleigh}
    If $\mX, \mY \in \R^{m \times m}$ are symmetric PSD matrices, then
    \begin{align*}
        \sup_{\vu \in \colspan(\mX)} \frac{\vu^\top \mY \vu}{\vu^\top \mX \vu}
        = \lambda_1\!\left((\sqrt{\mX})^{+} \mY (\sqrt{\mX})^{+}\right).
    \end{align*}
\end{lemma}
\begin{proof}
    Let $\vu \in \colspan(\mX) \setminus \{\vzero\}$. Then $\vu = \sqrt{\mX} \vz$ for some $\vz \in \R^m \setminus \{\vzero\}$.
    Thus,
    \begin{align*}
        \frac{\vu^\top \mY \vu}{\vu^\top \mX \vu}
        = \frac{\vz^\top (\sqrt{\mX})^{+} \mY (\sqrt{\mX})^{+} \vz}{\vz^\top \vz}
        \le \lambda_1\!\left((\sqrt{\mX})^{+} \mY (\sqrt{\mX})^{+}\right).
    \end{align*}
    The equality is achieved when $\vz$ is the top eigenvector of $(\sqrt{\mX})^{+} \mY (\sqrt{\mX})^{+}$.
\end{proof}

\subsection{Proof of Theorem \ref{thm:general-main}}

\begin{lemma}\label{lem:early_stop_loss}    Under~\Cref{ass:fstar}, for all $T > 0, S \ge K$ and $\vw \in \R^m$, we have
    \begin{align*}
        \cL(f_T) \le 
        \cL(\hat{f}_{\vw})
        + \frac{e^{-\lambda_K T}}{2 - e^{-\lambda_K T}} \norm{f^*}_{\cD}^2
        -(1 - \lambda_{S+1}^2 T^2) \sum_{k \ge S + 1} \inne{\hat{f}_{\vw}}{e_k}_{\cD}^2.
    \end{align*}
\end{lemma}
\begin{proof}[Proof of \Cref{lem:early_stop_loss}]
    By \eqref{eq:student-dynamics},
    \begin{align*}
        \cL(f_{T}) &=
        \sum_{k \ge 1} \underbrace{\left(\left(1 - e^{-\lambda_k T}\right) \inne{\hat{f}_{\vw}}{e_k}_{\cD} - \inne{f^*}{e_k}_{\cD}\right)^2}_{=:\,\ell_k}.
    \end{align*}
    We provide upper bounds for $\ell_k$ in three cases. First, for all $1 \le k \le K$, by~\Cref{lm:cbDelta} we have
    \begin{align*}
        \ell_k = \left(\left(1 - e^{-\lambda_k T}\right) \inne{\hat{f}_{\vw}}{e_k}_{\cD} - \inne{f^*}{e_k}_{\cD}\right)^2
        &\le \left(\inne{\hat{f}_{\vw}}{e_k}_{\cD}^2 - \inne{f^*}{e_k}_{\cD}\right)^2 + \frac{e^{-\lambda_k T}}{2 - e^{-\lambda_k T}} \inne{f^*}{e_k}_{\cD}^2 \\
        &\le \left(\inne{\hat{f}_{\vw}}{e_k}_{\cD}^2 - \inne{f^*}{e_k}_{\cD}\right)^2 + \frac{e^{-\lambda_K T}}{2 - e^{-\lambda_K T}} \inne{f^*}{e_k}_{\cD}^2,
    \end{align*}
    where we used the fact that $e^{-\lambda_k T} \le e^{-\lambda_K T}$ for $1 \le k \le K$.

    Then, for $K + 1 \le k \le S$, we have
    \begin{align*}
        \ell_k = \left(\left(1 - e^{-\lambda_k T}\right) \inne{\hat{f}_{\vw}}{e_k}_{\cD} - \inne{f^*}{e_k}_{\cD}\right)^2 = \left(1 - e^{-\lambda_k T}\right)^2 \inne{\hat{f}_{\vw}}{e_k}_{\cD}^2 \le \left(\inne{\hat{f}_{\vw}}{e_k}_{\cD} - \inne{f^*}{e_k}_{\cD}\right)^2.
    \end{align*}
    Finally, for $k \ge S+1$, we have $e^{-\lambda_k T} \ge e^{-\lambda_{S+1}T} \ge 1-\lambda_{S+1}T$, and thus $\left(1 - e^{-\lambda_k T}\right)^2 \le \lambda_{S+1}^2 T^2$. This implies the following bound for $\ell_k$:
    \begin{align*}
        \ell_k = \left(\left(1 - e^{-\lambda_k T}\right) \inne{\hat{f}_{\vw}}{e_k}_{\cD} - \inne{f^*}{e_k}_{\cD}\right)^2 = (1 - e^{-\lambda_k T})^2 \inne{\hat{f}_{\vw}}{e_k}_{\cD}^2 &\le \lambda_{S+1}^2 T^2 \inne{\hat{f}_{\vw}}{e_k}_{\cD}^2..
    \end{align*}
    Putting all these together proves the following:
    \begin{align*}
        \cL(f_T) &\le \sum_{k=1}^{S} \left(\inne{\hat{f}_{\vw}}{e_k}_{\cD}^2 - \inne{f^*}{e_k}_{\cD}\right)^2
        + \frac{e^{-\lambda_K T}}{2 - e^{-\lambda_K T}} \sum_{k=1}^{K} \inne{f^*}{e_k}_{\cD}^2
        + \lambda_{S+1}^2 T^2\sum_{k \ge S + 1}  \inne{\hat{f}_{\vw}}{e_k}_{\cD}^2 \\
        &= \sum_{k \ge 1} \left(\inne{\hat{f}_{\vw}}{e_k}_{\cD}^2 - \inne{f^*}{e_k}_{\cD}\right)^2
        + \frac{e^{-\lambda_K T}}{2 - e^{-\lambda_K T}} \sum_{k=1}^{K} \inne{f^*}{e_k}_{\cD}^2
        -(1 - \lambda_{S+1}^2 T^2) \sum_{k \ge S + 1} \inne{\hat{f}_{\vw}}{e_k}_{\cD}^2 \\
        &= \cL(\hat{f}_{\vw})
        + \frac{e^{-\lambda_K T}}{2 - e^{-\lambda_K T}} \norm{f^*}_{\cD}^2
        -(1 - \lambda_{S+1}^2 T^2) \sum_{k \ge S + 1} \inne{\hat{f}_{\vw}}{e_k}_{\cD}^2,
    \end{align*}
    which completes the proof.
\end{proof}

It remains to bound $\sum_{k \ge S + 1} \inne{\hat{f}_{\vw}}{e_k}_{\cD}^2$.
Let $G: \R^m \to \gH$ be the linear operator that maps $\vu \in \R^m$ to $\sum_{i=1}^{m} u_i g_i \in \gH$.
Let $Pf := \sum_{k=1}^{S} \inne{f}{e_k}e_k$ be the projection operator onto the top-$K$ eigenspace of $\gK$,
and $Qf$ be the projection onto the span of $\{g_1, \dots, g_m\}$.
Then
$G^* G = \mPhi$,
$G^* P G = \mA$ and $Q = G^* \mPhi^+ G$.

In the following, we give a characterization of $\kappa$.
\begin{lemma}\label{lm:rho-equals-kappa}
    If $\mA \ne \vzero$, then $R := QPQ \ne 0$ and $\kappa$ equals the smallest non-zero eigenvalue of $R$.
\end{lemma}
\begin{proof}
    If $\mA \ne \vzero$, then there exist $1 \le i < j \le m$ such that $\inne{Pg_i}{Pg_j} \ne 0$. This implies $R \ne 0$ because $\inne{g_i}{Rg_j} = \inne{PQg_i}{PQg_j} \ne 0$.

    Let $\rho_{\min}$ be the smallest non-zero eigenvalue of $R$.
    To see why $\kappa = \rho_{\min}$, we first express $\rho_{\min}$ as the minimum of a Rayleigh quotient:
    \begin{align}\label{eq:rho-as-rayleigh}
        \rho_{\min} = \inf_{\alpha \in \range(R)} \frac{\inne{\alpha}{R \alpha}}{\inne{\alpha}{\alpha}}.
    \end{align}
    Then we have
    \begin{align*}
        \range(R) = \range(QPQ)
        = \{ QPG\vv \mid \vv \in \R^m \}
        &= \{ GG^*QPG\vv \mid \vv \in \R^m \} \\
        &= \{ GG^*PG\vv \mid \vv \in \R^m \}.
    \end{align*}
    Since $\mA = G^*PG$, we further have
    \begin{align*}
        \range(R) 
        = \{ G \mA \vv \mid \vv \in \R^m \}
        = \{ G \vu \mid \vu \in \colspan(\mA) \}.
    \end{align*}
    Plugging this into \eqref{eq:rho-as-rayleigh}, we have
    \begin{align*}
        \rho_{\min} = \inf_{\vu \in \colspan(\mA)} \frac{\inne{G\vu}{R G \vu}}{\inne{G\vu}{G\vu}}  = \inf_{\vu \in \colspan(\mA)} \frac{\inne{\vu}{G^* P G \vu}}{\inne{\vu}{G^*G\vu}}
        &= \inf_{\vu \in \colspan(\mA)} \frac{\vu^\top \mA \vu}{\vu^\top \mPhi \vu} \\
        &= \left( \sup_{\vu \in \colspan(\mA)} \frac{\vu^\top \mPhi \vu}{\vu^\top \mA \vu} \right)^{-1}.
    \end{align*}
    Since $\mPhi = \mA + \mB$, we further have
    \begin{align*}
        \rho_{\min}
        = \left( \sup_{\vu \in \colspan(\mA)} \frac{\vu^\top (\mA + \mB) \vu}{\vu^\top \mA \vu} \right)^{-1}
        = \left( 1 + \sup_{\vu \in \colspan(\mA)} \frac{\vu^\top \mB \vu}{\vu^\top \mA \vu} \right)^{-1}.
    \end{align*}
    By~\Cref{lm:max-rayleigh}, the supermum here equals $\lambda_1\!\left((\sqrt{\mA})^+ \mB (\sqrt{\mA})^+\right)$. Therefore,
    \begin{align*}
        \rho_{\min}
        = \frac{1}{1 + \lambda_1\left((\sqrt{\mA})^+ \mB (\sqrt{\mA})^+\right)} = \kappa,
    \end{align*}
    which completes the proof.
\end{proof}

Now we give a lower bound for $\sum_{k \ge S + 1} \inne{\hat{f}_{\vw}}{e_k}_{\cD}^2$ in terms of $\cL(\hat{f}_{\vw})$.
\begin{lemma}\label{lm:lower_bound_improvment}
    Under~\Cref{ass:fstar,ass:rankA},
    if $\gL(\hat{f}_{\vw}) = \min_{\vw' \in \R^m} \gL(\hat{f}_{\vw'})$, then
    \begin{align}\label{eq:lower_bound_improvement}
        \sum_{k \ge S + 1} \inne{\hat{f}_{\vw}}{e_k}_{\cD}^2
        \ge \kappa \cdot \cL(\hat{f}_{\vw}),
    \end{align}
    where
    the equality is attained iff $f^* = Ph^*$
    for some $h^* \in \gH$ that is in the span of the eigenvectors of $R := QPQ$ corresponding to the eigenvalue $1$ (if exists) and the smallest non-zero eigenvalue, which is $\kappa$.
\end{lemma}
\begin{proof}[Proof of \Cref{lm:lower_bound_improvment}]
    By~\Cref{ass:rankA}, $f^* \in \vspan\{Pg_1, \dots, Pg_m\} = \range(PG) = \range(PQ)$, which implies that there exists $h \in \gH$
    such that $f^* = PQh$.
    As $\vw$ minimizes $\gL(\hat{f}_{\vw})$,
    $\hat{f}_{\vw}$ must be the least square solution
    and $\hat{f}_{\vw} = Qf^* = QPQh = Rh$.
    So we have the following identities:
    \begin{align*}
        &\sum_{k \ge S + 1} \inne{\hat{f}_{\vw}}{e_k}_{\cD}^2
        = \norm{(I-P)\hat{f}_{\vw}}_{\gD}^2
        = \norm{(I - P) QPQh}_{\gD}^2
        = \inner{h}{QPQ(I-P)QPQh}_{\gD}, \\
        &\cL(\hat{f}_{\vw})
        = \norm{f^* - \hat{f}_{\vw}}_{\gD}^2
        = \norm{PQh - QPQh}^2
        = \lnorm{(I - Q)PQh}^2
        = \inner{h}{QP(I-Q)PQ h}_{\gD}.
    \end{align*}
    We can express the linear operators $QPQ(I-P)QPQ$ and $QP(I-Q)PQ$ in terms of $R$ as follows.
    \begin{align*}
        QPQ(I-P)QPQ &= (QPQ)(QPQ) - (QPQ) (QPQ) (QPQ) = R(R-R^2), \\ 
        QP(I-Q)PQ &= QPQ - (QPQ) (QPQ) = R - R^2. 
    \end{align*}
    Note that $0 \preceq R \preceq I$ since $\inne{f}{QPQf}_{\gD} = \inne{Qf}{P(Qf)}_{\gD}$
    and $0 \le \inne{Qf}{P(Qf)}_{\gD} \le \inne{Qf}{Qf}_{\gD} = \inne{f}{f}_{\gD}$.
    Let $R f = \sum_{i=1}^{r} \rho_i \inne{v_i}{f} v_i$ be an eigendecomposition of $R$, where $1 \ge \rho_1 \ge \rho_2 \ge \cdots \ge \rho_r > 0$ and $\{v_1, \dots, v_r\}$ are the corresponding eigenvectors.

    By~\Cref{ass:rankA}, $\mA \ne \vzero$. So by~\Cref{lm:rho-equals-kappa}, $R \ne 0$, $r \ge 1$ and $\rho_r = \kappa$.
    Then we have
    \begin{align*}
        \sum_{k \ge S + 1} \inne{\hat{f}_{\vw}}{e_k}_{\cD}^2
        = \inner{h}{(R(R-R^2))h}_{\gD} 
        &= \sum_{i=1}^{r} \rho_i \cdot (\rho_i - \rho_i^2) \inne{h}{v_i}_{\gD}^2 \\
        &\ge \sum_{i=1}^{r} \kappa \cdot (\rho_i - \rho_i^2) \inne{h}{v_i}_{\gD}^2 
        = \kappa \cdot \inner{h}{(R-R^2)h}_{\gD} = \kappa \cdot \cL(\hat{f}_{\vw}),
    \end{align*}
    which proves \eqref{eq:lower_bound_improvement}.

    To understand when the equality is attained, we note that the equality holds iff the inequality in the second line is an equality.
    This means that for all $1 \le i \le r$, either $\rho_i \in \{1, \kappa\}$ or $\inne{h}{v_i}_{\gD} = 0$. Equivalently, $h$ can be expressed as $h = \mu_1 h_1 + \mu_2 h_2$ for some $\mu_1, \mu_2 \in \R$, where $h_1$ is an eigenvector of $R$ corresponding to eigenvalue $1$ and $h_2$ is an eigenvector of $R$ corresponding to eigenvalue $\kappa$.
    
    In this case, $f^* = PQh = \mu_1 PQh_1 + \mu_2 PQh_2$. This can be further simplified to $f^* = \mu_1 Ph_1 + \mu_2 Ph_2 = Ph$ since $h_1, h_2 \in \range(R) \subseteq \range(Q)$.
\end{proof}

\begin{proof}[Proof of~\Cref{thm:general-main}]
    Putting \Cref{lem:early_stop_loss}, \Cref{lm:lower_bound_improvment} and \Cref{lm:rho-equals-kappa} together, we have
    \begin{align*}
        \cL(f_T)
        &\le \cL(\hat{f}_{\vw})
                + \frac{e^{-\lambda_K T}}{2 - e^{-\lambda_K T}} \norm{f^*}_{\cD}^2
                -(1 - \lambda_{S+1}^2 T^2) \kappa \cdot \cL(\hat{f}_{\vw}) \\
        &= \left(1 - (1 - \lambda_{S+1}^2 T^2) \kappa\right) \cL(\hat{f}_{\vw}) + \frac{e^{-\lambda_K T}}{2 - e^{-\lambda_K T}} \norm{f^*}_{\cD}^2,
    \end{align*}
    which completes the proof.
\end{proof}

\subsection{General Lower Bound on $\kappa_S$}\label{app:kappa-bound}

We will bound 
\begin{align*}
    \kappa_S := 
    \frac{1}{1+\lambda_1\!\left((\sqrt{\mA_S})^+ \mB_S (\sqrt{\mA_S})^+\right)}.
\end{align*}
in the general case.

Let $F_A$ and $F_B$ be defined as follows:
\begin{align*}
    F_A &: \text{span}(e_1,\dots,e_J) \to \R^m \\
    F_A &: h \mapsto \begin{pmatrix} \vdots \\
        \langle h,g_j \rangle_{\cD}\\
        \vdots \\
    \end{pmatrix} \\
    F_B &: \text{span}(e_J, e_{J+1},\dots) \to \R^m\\
    F_B &: h \mapsto \begin{pmatrix} \vdots \\
        \langle h , g_j \rangle_{\cD} \\
        \vdots \\
    \end{pmatrix}.
\end{align*}
We will denote $F_A^{\dagger}$ and $F_B^{\dagger}$ their adjoint operators. We can rewrite the largest eigenvalue of $(\sqrt{\mA_S})^+ \mB_S (\sqrt{\mA_S})^+$ as 
\begin{align*}
    \lambda_1\!\left((\sqrt{\mA_S})^+ \mB_S (\sqrt{\mA_S})^+\right) = \sup_{v \in  \text{span}(e_1,\dots,e_J)} \frac{ v^T F_A^{\dagger}F_B F_B^{\dagger}F_A v}{v^T F_A^{\dagger}F_A F_A^{\dagger}F_Av}
\end{align*}
Let $\cH$ be a Hilbert space defined as 
\begin{align*}
    \cH = \{ \text{linear mappings from } \text{span}(e_1,\dots,e_J) \text{ to } \text{span}(e_J, e_{J+1},\dots) \}
\end{align*}
with the inner product defined as 
\begin{align*}
    \langle f,f' \rangle_{\cH} = \sum_{j=1}^{J} \langle f(e_i), f'(e_i) \rangle_{\cD}.
\end{align*}
We will show the following lower bound first. 
\begin{proposition}\label{prop:app:falower}
For $v\in  \text{span}(e_1,\dots,e_J) $ it holds that
\begin{align*}
    v^T F_A^{\dagger}F_A F_A^{\dagger}F_Av \ge \lambda_J(\mA)^2 \| v\|_2^2.
\end{align*}
    
\end{proposition}
\begin{proof}
    Let $\sum_{i=1}^{J} u_i e_i  = F_A^{\dagger}F_A v$.  Then we have that 
   \begin{align*}
       v^T F_A^{\dagger}F_A F_A^{\dagger}F_Av \ge  \langle \sum_{i=1}^{J} u_i e_i, \sum_{i=1}^{J} u_i e_i \rangle_{\cD} \ge \| u \|_2^2 \ge \lambda_J(\mA)^2 \| v \|_2^2.
   \end{align*}
\end{proof}

For the upper bound, we will use the concentration of $\| F_B^{\dagger} F_A v \|_{\cD}^{2}$.

\begin{proposition}\label{prop:app:fbfaupper}
    For $v = \sum_{j=1}^{J} v_i e_i$ it holds that 
    \begin{align*}
        \| F_B^{\dagger} F_A v \|_{\cD}^2 \le \left( \sum_{j=1}^{J} \|F_B^{\dagger} F_A e_i \|_{\cD}^{2} \right) \left( \sum_{j=1}^{J} v_i^2 \right).                   
    \end{align*}
\end{proposition}

\begin{proof}
    This follows immediately from Cauchy-Schwarz inequality.
\end{proof}

\begin{proposition}\label{app:normy}
Let $Y\in \cH$ be such that $\langle e_j, Y(v) \rangle = \langle e_j,g \rangle_{\cD} \langle g,v \rangle_{\cD}$
\begin{align*}
    \| Y \|_{H}^2 = \sum_{j=1}^{J} \sum_{k=J+1}^{\infty} \langle e_j, g \rangle_{\cD} ^2 \langle g,e_k \rangle_{\cD}^2 
\end{align*}
    so $\| Y \|_{H}^2= \| P_{\le J} g\|^2 \| P_{\ge J+1} g\|^2$, that is $\| Y \|_{H}= \| P_{\le J} g\|_{\cD} \| P_{\ge J+1} g\|_{\cD}$. 
    
    Furthermore, let $Y_i$ be the r.v. corresponding to the draw of the $i$-th feature $g_i$. Then 
    \begin{align*}
        F_B^{\dagger}F_A = \sum_{i=1}^{m} Y_i.
    \end{align*}
\end{proposition}

We have the following concentration bound in a Hilbert space.

\begin{theorem}[\cite{pinelis1994}]\label{thrm:hilberconcentration}
    Let $Y_i$ be independent $\cH$ valued random variable with $\E(Y_i)=0$, where $\cH$ is a seperable Hilbert space. Assume that for every $q\ge 2, q\in \mathbb N$ we have that $\E \left( \| Y_i \|_{\cH}^{q}\right)\le \frac{1}{2} q! B^2 L^{q-2}$. Then, for all $n\in \mathbb N$ and $\epsilon >0$ we have 
    \begin{align*}
        P\left(\lnorm{\frac{1}{n}\sum_{i=1}^{n} Y_i}_{\cH} > \epsilon\right) \le 2\exp \left( - \frac{n\epsilon^2}{B^2 + L \epsilon + B\sqrt{B^2+2L\epsilon}} \right).
    \end{align*}
\end{theorem}

Putting these together in the bound for $\lambda_1\!\left((\sqrt{\mA_S})^+ \mB_S (\sqrt{\mA_S})^+\right)$, we get the following.

\begin{proposition}\label{app:prop:kappageneral}
    The largest eigenvalue of $(\sqrt{\mA_S})^+ \mB_S (\sqrt{\mA_S})^+$ is bounded by 
    \begin{align*}
        \lambda_1\!\left((\sqrt{\mA_S})^+ \mB_S (\sqrt{\mA_S})^+\right) \le \frac{\sum_{j=1}^{J} \|F_B^+F_Ae_j\|_{\cD}^2}{\lambda_J(\mA)^2}
    \end{align*}
\end{proposition}

\begin{proof}[Proof of \Cref{app:prop:kappageneral}]
This follows directly from \Cref{prop:app:fbfaupper} and \Cref{prop:app:falower}.
\end{proof}

We can now use this to bound the norm of $F_B^{\dagger}F_A$.

\begin{proposition}[Hilbert Space Concentration]\label{prop:app:hilbertbound}
    With probability at least $1-2\delta$
    \begin{align*}
        \|F_B^{\dagger}F_A\|_{\cH}^2 \le m B \left(\log \frac{1}{\delta}\right)^2
    \end{align*}
    where $B$ is as defined in \Cref{thrm:hilberconcentration}.
\end{proposition}

\begin{proof}[Proof of \Cref{prop:app:hilbertbound}]
    In \Cref{thrm:hilberconcentration}, taking $m^2\eps^2 = m B^2 \log^2 \frac{1}{\delta}$, we get the desired result.
\end{proof}

We can lower bound the eigenvalues of $\mA$ using matrix concentration results.


\begin{theorem}\label{thrm:app:Aeigenvaluelowerbound}
    Let $\va=\begin{pmatrix} \vdots \\
        \langle g,e_k \rangle \\ \vdots
    \end{pmatrix}_{k=1}^{J}$ as $g\sim \cG$ be a row of $\mA$.
    Depending on whether the rows of $\mA$ are subgaussian or bounded a.s. we have the following two bounds:
    \begin{enumerate}
        \item Under \Cref{ass:app:daigfeat}, i.e. with $D_g = N(0,\Lambda)$ where $\Lambda = \text{diag}(\lambda_1,\lambda_2,\dots)$, then with probability $1-2\exp(-ct_A^2)$ it holds that
    \begin{align*}
        \lambda_{J}(\mA)\ge \lambda_{J}\left( \sqrt{m} -C\sqrt{J}-t_A\right)^2,
    \end{align*}
    if $\sqrt{m}\ge C\sqrt{J}+t_A$, where $C=\Theta(L_{g}^2)$ and $c= \Theta(1/L_{g}^{4})$ where $L_{g}$ is the subgaussian norm of the vector $\va$. In this case $L_{g}\le C_g \lambda_1$, where $C_g$ is an absolute constant.

    \item Under \Cref{ass:relu}, i.e. if $g= \sigma(\langle w,x \rangle) $ with $w\sim \text{unif}(S^{d-1})$ and $N_0+N_1+\dots+N_{s-1}+1\le J \le N_0+N_1+\dots+N_{s}$ and $s$ is $1$ or even, we have that with probability $1-2J \exp(-ct_A^2)$
    \begin{align*}
        \lambda_{J}(\mA) \ge \sigma_{s}^2 \left( \sqrt{m}-t_A \sqrt{J}\right)^2,
    \end{align*}
    if $\sqrt{m}\ge t_A\sqrt {J}$, where $c$ is an absolute constant.
    \end{enumerate}
    \item Under \Cref{ass:2layer}, with probability $1-2J\exp(-ct_A^2)$ we have that 
    \begin{align*}
        \lambda_{J}(\mA) \ge \sigma_{s}^2 \left( \sqrt{m}-t_A \sqrt{J}\right)^2,
    \end{align*}
    
\end{theorem}

This computation allows us to lower bound $\kappa_S$ in the cases of \Cref{ass:relu} in \Cref{app:2layer} and \Cref{ass:linearnetwork} in \Cref{ass:app:daigfeat}.

\begin{proof}[Proof of \Cref{thrm:app:Aeigenvaluelowerbound}]
    We will use Theorems $5.39$ and $5.41$ in \citet{vershynin2011}. These theorems say that if the rows of a matrix are either bounded a.s.~or sub-gaussian, then the singular values of this matrix are upper and lower bounded. Note first that in both cases, $\mA$ is a PSD matrix, so it's sufficient to consider the singular values of $\mA$. In both cases, note that in order to apply the theorems from \citet{vershynin2011} to a matrix $\mX$, we need to have $\Sigma(\mX)=\mI$, i.e. the rows of $\mX$ have to be isotropic random variables. Note that in both cases
    \begin{align*}
        \Sigma(\mA)_{ij} = \E_{\cG}\left(\langle g,e_i\rangle \langle g,e_j\rangle \right).
    \end{align*} 
    If we let $\Sigma = \Sigma(\mA)$, we can apply these theorems to $\Sigma^{-\frac{1}{2}}\mG$, where $\mG$ is such that $\mA = \mG^T \mG$. In that case, the theorems say that with corresponding probabilities that
    \begin{align*}
        s_J(\Sigma^{-\frac{1}{2}}\mG) &\ge (\sqrt{m}-c\sqrt{J}-t_A)^2 \\
        s_J(\Sigma^{-\frac{1}{2}}\mG) &\ge (\sqrt{m}-t_A\sqrt{J})^2,
    \end{align*}
    respectively, where $s_J$ is the $J$-th singular value of a matrix.
    Note that we have that  
    \begin{align*}
        s_J(\mG) &\ge s_{\min}(\Sigma^{\frac{1}{2}}) s_J(\Sigma^{-\frac{1}{2}} \mG).
    \end{align*}
    Therefore, we get that with corresponding probabilities 
    \begin{align*}
        \lambda_J(\mA) &\ge \lambda_{J}(\Sigma) \left(s_J(\Sigma^{-\frac{1}{2}} \mG)\right)^2.
    \end{align*}
    Here, we need lower bounds on $m$ so that we can square the inequality and go from singular values to eigenvalues.
    
    In the case of \Cref{ass:linearnetwork}, we have that 
    \begin{align*}
        \Sigma(\mA)_{ij} = \Lambda.
    \end{align*}
    Therefore, we have that $\lambda_J(\Sigma) = \lambda_J$. Note that we are only left with verifying the conditions of the theorem. First is that the rows of $\Sigma^{-\frac{1}{2}}\mA$ are isotropic which is true by design. Second, we need that the rows of $\Sigma^{-\frac{1}{2}}\mA$ are subgaussian with a subgaussian constant $L_g$. Note note that if $X\sim N(0,\sigma)$ then $\|X\|_{\psi_2}\le C\sigma$ for some absolute constant $C$. Note also that $\| \va\|_{\psi_2}=\sup_{x\in S^{J-1}} \| x^T\va \|_{\psi_2}\le \sqrt{C_1 \sum_{i=1}^{J} x_i^2 \|\va_i^2\|_{\psi_2}^2} \le \sqrt{C_1C \sum_{i=1}^{J} x_i^2 \lambda_i^2}\le\sqrt{C_1C\lambda_1^2 \sum_{i=1}^{J} x_i^2 }=\sqrt{C_1C\lambda_1^2} = C_g \lambda_1$. Note that therefore under \Cref{ass:linearnetwork}, the conditions of the Theorem $5.39$ in \cite{vershynin2011} are satisfied, so we get that with probability $1-2\exp(-ct_A^2)$
    \begin{align*}
        \lambda_J(\mA)\ge \lambda_J \left( \sqrt{m} -C\sqrt{J}-t_A\right)^2,
    \end{align*}
    as long as $\sqrt{m}\ge c\sqrt{J} +t_A$. Here $C=\Theta(L_g^2)$ and $c=\Theta(\frac{1}{L_g^4})$ in the sense that both $C$ and $c$ are upper and lower bounded by an absolute constant times the appropriate expression of $L_g$.

    In the case of \Cref{ass:relu}, note that since the target \Cref{ass:fstar} will not have anything in the directions with zero $\sigma_k$, we can just ignore and reindex the basis functions. So, for now we will just write as if they are nonzero, but we will keep in mind that we skip the zero $\sigma_k$. Therefore, we have that 
    \begin{align*}
        \Sigma(\mA)_{i,j} = \text{diag}(\sigma_1^2,\sigma_1^2,\dots,\sigma_2^2,\dots,\sigma_4^2,\dots),
    \end{align*}
    where each $\sigma_i$ repeats with the multiplicity $N_i$ for $i=1$ or $i$ even (otherwise it's $0$). Furthermore, note that $\| \va \|_2^2 = J$, so we can indeed apply Theorem $5.41$ from \citep{vershynin2011}.
    Therefore, we have that with probability at least $1-2J\exp(-ct_A^2)$
    \begin{align*}
        \lambda_J(\mA)\ge \sigma_s^2(\sqrt{m}-t_A \sqrt{J})^2
    \end{align*}
    where $c$ is an absolute constant.
\end{proof}

\section{Weak-to-Strong Bound Applied to the Case of Linear Networks}\label{app:diagfeat}

\begin{proof}[Proof of \Cref{thrm:main:diagfeatcov}]
Follows from \Cref{thrm:app:diagfeatcov}.
\end{proof}

\begin{proof}[Proof of \Cref{thrm:main:diagfeatgeneral}]
    This follows directly from \Cref{app:prop:modelequiv} and the general bound \Cref{thrm:app:diagfeat}.
\end{proof}

\begin{theorem}[Weak-to-Strong Generalization with Gaussian Feature Distribution]\label{thrm:app:diagfeat}
    Under \Cref{ass:fstar}, if \Cref{ass:app:daigfeat} holds for some $K$, and if the teacher attains minimum loss $\TL=\hat{\cL}_{\min}$, then for the student trained until time $T$ we have with probability at least $1-\frac{4}{m}$ that
    \begin{align*}
        \SL &\le 
\inf_{S \ge K} \Biggl\{
    \frac{\bigl(\sum_{i=1}^{S} \lambda_i\bigr)\bigl(\sum_{i=S+1}^{\infty} \lambda_i\bigr)}{\lambda_S^2}
    \,\frac{(\log m)^2}{m}\,\TL
    + \lambda_{S+1} \, T^2 \,\TL\Biggr\}
    + \frac{e^{-\lambda_K T}}{2 - e^{-\lambda_K T}} \,\bigl\|f^*\bigr\|_{\cD}^2.
    \end{align*}
\end{theorem}
\begin{proof}[Proof of \Cref{thrm:app:diagfeat}]
    This follows from \Cref{thm:general-main} using \Cref{thrm:app:helpingupperbound} and noting that we can replace $\lambda_{S+1}T^2(1-\kappa_S)\le \lambda_{S+1}T^2$. 
\end{proof}


\subsection{Proofs Related to The Case of Linear Networks}

We first aim to show that the $k$-th moments of $\| Y \|_{\cH} = \|P_{\le J} g\|_{\cD}\|P_{\ge J+1} g\|_{\cD} $ are bounded.

\begin{proposition}
    Let $X=\sum_{i=1}^{n} a_i^2 X_i^2$ where $X_i$ are i.i.d. standard Gaussian. Then for $k\ge 2 $
    \begin{align*}
         \E\left( X^k\right) \le k! \left( \sum_{i=1}^{m} a_i^2\right)^{k}.
    \end{align*}
\end{proposition}

\begin{proof} \removed{\marko{could try to imrpove the constant $2$ here.}}
Expanding the expression for $X^k=\left(\sum_{i=1}^{n} a_i^2 X_i^2\right)^k$ and using linearity of expectation, we have that
    \begin{align*}
        \E\left( X^k\right) &= \sum_{|\alpha| =k}\frac{k!}{\alpha!}\prod_{i=1}^{m} \left((a_i)^2\right)^{\alpha_i} \E\left(\left(X_i^2\right)^{\alpha_i}\right).
    \end{align*}
    Note that $\E(X_i^{2\alpha_i}) = (2\alpha_i-1)!!\le 2^{\alpha_i} (\alpha_i)!$, so we have that 
    \begin{align*}
        \E\left( X^k\right) &= \sum_{|\alpha| =k}\frac{k!}{\alpha!}\prod_{i=1}^{m} \left((a_i)^2\right)^{\alpha_i} \E\left(\left(X_i^2\right)^{\alpha_i}\right) \le \sum_{|\alpha| =k}\frac{k!}{\alpha!}\prod_{i=1}^{m} \left((a_i)^2\right)^{\alpha_i}2^{\alpha_i} (\alpha_i)! \\
        & = \sum_{|\alpha| =k}\frac{k!}{\alpha!}\alpha! \prod_{i=1}^{m} \left(2(a_i)^2\right)^{\alpha_i}  = k! \sum_{|\alpha|=k}\frac{\alpha!}{\alpha!} \prod_{i=1}^{m}\left(2(a_i)^2\right)^{\alpha_i}\\
        &\le k! \sum_{|\alpha|=k} \frac{k!}{\alpha!} \prod_{i=1}^{m} \left(2(a_i)^2\right)^{\alpha_i}
    \end{align*}
    where the last inequality is true because $\alpha! \le k!$. Note that $\sum_{|\alpha|=k} \frac{k!}{\alpha!} \prod_{i=1}^{m} \left(2(a_i)^2\right)^{\alpha_i}=\left(\sum_{i=1}^{m} 2 a_i^2 \right)^k$, so the final bound is 
    \begin{align*}
        \E\left( X^k\right) &\le k! \sum_{|\alpha|=k} \frac{k!}{\alpha!} \prod_{i=1}^{m} \left(2(a_i)^2\right)^{\alpha_i} = k! \left(\sum_{i=1}^{m} 2 a_i^2 \right)^k.
    \end{align*}
\end{proof}

This implies that $\| Y \|_{\cH}^k$ is bounded for all $k\ge 2$ with an expression of the required form.
\begin{proposition}\label{prop:app:Bcompgauss}
    Let $g\sim \cG$ with $\langle g,e_i \rangle \sim N(0,\Lambda)$ with $\Lambda = \text{diag}(\lambda_1,\lambda_2,\dots)$. Then for $Y$ it holds that for all $k\ge 2$
    \begin{align*}
        \E\left(\| Y \|_{\cH}^k\right) \le k! \left(\sqrt{\left(2\sum_{i=1}^{J}\lambda_i\right)\left(2\sum_{i=J+1}^{\infty}\lambda_i\right)}\right)^k.
    \end{align*}
    Therefore, in this case we can take $B=\sqrt{2\left(2\sum_{i=1}^{J}\lambda_i\right)\left(2\sum_{i=J+1}^{\infty}\lambda_i\right)}$ and $L=\sqrt{\left(2\sum_{i=1}^{J}\lambda_i\right)\left(2\sum_{i=J+1}^{\infty}\lambda_i\right)}$.
\end{proposition}
Note that \Cref{prop:app:hilbertbound} applies in this case as well. Additionally, \Cref{thrm:app:Aeigenvaluelowerbound} applies in this case. 
\begin{theorem}\label{thrm:app:helpingupperbound}
     Let $g\sim \cG$ with $\langle g,e_i \rangle \sim N(0,\Lambda)$ with $\Lambda = \text{diag}(\lambda_1,\lambda_2,\dots)$. Then with probability at least $1-2\delta-2\exp(-c_1\frac{1}{\lambda_1^4}t_A^2 )$ we have that 
     \begin{align*}
          \lambda_1\!\left((\sqrt{\mA_S})^+ \mB_S (\sqrt{\mA_S})^+\right) \le \frac{\|F_B^{\dagger}F_A\|_{\cH}^2}{\lambda_S(\mA)^2} \le \frac{8m\left(\sum_{i=1}^{S}\lambda_i\right)\left(\sum_{i=S+1}^{\infty}\lambda_i\right)\left(\log \frac{1}{\delta}\right)^2}{\lambda_S ^2\left( \sqrt{m}-C\sqrt{S}-t_A\right)^4}
     \end{align*}
     where $c_1$ and $C$ are absolute constants.
     Setting $t_A = \lambda_1^2 \log \frac{1}{\delta_1}$, $\delta=\frac{1}{m}, \delta_1 =\frac{1}{m} $, we get that with probability at least $1-\frac{4}{m}$ we have 
     \begin{align*}
          \lambda_1\!\left((\sqrt{\mA_S})^+ \mB_S (\sqrt{\mA_S})^+\right) \le  \frac{8m\left(\sum_{i=1}^{S}\lambda_i\right)\left(\sum_{i=S+1}^{\infty}\lambda_i\right)\left(\log m\right)^2}{\lambda_S^2 \left( \sqrt{m}-C\sqrt{S}-\lambda_1^2\log m\right)^4}.
     \end{align*}
     If additionally $m=\omega(S)$ and $m=\omega((\lambda_1 \log m)^2)$ then with probaility $1-\frac{4}{m}$
     \begin{align*}
    \lambda_1\!\left((\sqrt{\mA_S})^+ \mB_S (\sqrt{\mA_S})^+\right) \le  8\frac{\left(\sum_{i=1}^{S}\lambda_i\right)\left(\sum_{i=S+1}^{\infty}\lambda_i\right)}{\lambda_S^2 }\frac{\left(\log m\right)^2}{m}.
     \end{align*}
\end{theorem}
\begin{proof}[Proof of \Cref{thrm:app:helpingupperbound}]
    This theorems follows directly by taking the approproate lower bound from \Cref{thrm:app:Aeigenvaluelowerbound} and combining that with the Hilbert concentration bound \Cref{prop:app:hilbertbound}.
\end{proof}

\subsection{Additional Proofs for Linear Networks}\label{app:additionalproofslin}



\begin{proof}[Proof of \Cref{thrm:main:thetaone}]
    This follows from \Cref{thrm:app:diagfeatlowerbound} by taking $\epsilon_{d}=0.01 $, $\epsilon_{m} = 0.01$, $m\ge \Theta(\alpha^{\frac{2}{1-\epsilon_{d}}})$ and $\delta = 0.01$. Note that since $m= \sqrt{\frac{d-1}{\alpha}}$ we can turn this inequality into an inequality with $\alpha$ and $d$, i.e $\alpha\ge \Theta(\frac{1}{d^{0.49}})$. The last claim also follows from \Cref{thrm:app:diagfeatlowerbound}, since as $m \to \infty$ the lower bound converges to $0.99\frac{\alpha}{1+\alpha}$ and the probability to $1$. 
\end{proof}

\begin{theorem}[Weak-to-Strong Generalization with Linear Network]\label{thrm:app:diagfeatcov}
    Consider \Cref{ass:linearnetwork} when $f^*$ is supported by the first $k$ coordinates, i.e. $f^* = \beta \vx$, where $\beta_i=0$ for $i>k$ and take $f^*$ to be norm $1$. Take $\Psi_m = (1,\dots,1,\frac{1}{m},\dots,\frac{1}{m})$, where $1$ is repeated $k$ times and $\frac{1}{m}$ is repeated $N_1(m) = \alpha m^2$ times, with $\alpha = m^{-\eta}$ for any $\eta \in [0,\frac{1}{2}]$ (so $d=k+N_1(m)$). Then, for all $m>C\sqrt{k}$ \footnote{here $C$ is an absolute constant given by \Cref{thrm:app:helpingupperbound}}, if the student is trained until time $T=\log m +\log \frac{1}{\TL}$ with probability at least $0.99$,
    \begin{align*}
        \TL&\ge \frac{\frac{1}{k}m^{-\eta}}{1+\frac{1}{k}m^{-\eta}} \\
        \SL &\le k \frac{\log^2 m}{m^{\eta}} \TL + \frac{1}{m} (\log m + \log \frac{1}{\TL})^2 \TL + \frac{1}{m} \TL 
    \end{align*}

    So, for all $m$ we have with probability at least $0.99$, 
    \begin{align*}
        \SL\le \tilde{O}(k\TL^2). 
    \end{align*}
    In particular $\PGR\to 1$ as $\MT \to \infty$. 
\end{theorem}

\begin{proof}[Proof of \Cref{thrm:app:diagfeatcov}]

    For $k=1$ this follows from \Cref{thrm:app:diagfeatlowerbound}. Consider $k$ copies of \Cref{thrm:app:diagfeatlowerbound} with $N(m) = \frac{\alpha}{k} m^{2}$ each, $\Psi_{m,1},\dots, \Psi_{m,k}$. For example, $\Psi_{m,1} = (1,0,\dots,0, \frac{1}{m},\dots,\frac{1}{m},0,\dots,0)$, were we have $N(m)=\frac{N_1(m)}{k}$ repeating $\frac{1}{m}$. Note that a set of weights $\vw \in \R^m$ that the teacher chooses to optimize over all $k$ problems is worse than choosing a separate set of weights for each of the problems, so the loss will be greater than the sum of the losses in each of the subproblems. So if $\TL^i$ is the loss in the $i$-th instance, we have that $\TL\ge \sum_i \TL^i$. Let $B_{j}^{2}$ be the norm of $f^*$ in the directions of the basis functions corrresponding to the $j$-th subset of $\Psi_m$, i.e. $B_{j}^{2} = \sum_{i \in \Psi_{m,j} }\beta_i^2$. Note that by the normalization of $f^*$, $\sum_{j=1}^{k} B_{j}^2 = 1$. Furthermore, note that our setup is scale homogeneous, i.e. if we scale $\|f^*\|_{\cD}^2$ by $c^2$ then the loss will scale by $c^2$ as well. Therefore, we can apply \Cref{thrm:app:diagfeatlowerbound} to each of the $k$ subprolems to get with probability at least $1-2\exp(-\frac{1}{4}(\alpha m^2)^{\epsilon_d})-2\exp(-\frac{1}{4}m^{\epsilon_m})$ that
    \begin{align*}
        \TL^{j} \ge B_j^2\left(0.99\frac{\alpha}{k} \frac{1}{\frac{\alpha}{k}+1} -2m^{-1/2+\epsilon_m}-(\alpha m^2)^{-1/2+\epsilon_d} \alpha\right) \ge 0.98 B_j^2 \frac{\frac{m^{-\eta}}{k}}{\frac{m^{-\eta}}{k}+1}.
    \end{align*}

    Therefore, we have that 
    $\TL \ge \sum_{j=1}^{k}0.98 B_j^2 \frac{\frac{m^{-\eta}}{k}}{\frac{m^{-\eta}}{k}+1} =0.98 \frac{\frac{m^{-\eta}}{k}}{\frac{m^{-\eta}}{k}+1}$. 
    
    Now consider \Cref{thrm:app:diagfeat} in this case. Set $S=k$. Then we get for $\delta_T = \frac{1}{m}\TL$ the upper bound since the sums of $\psi_i$ are $\sum_{i=1}^{k} \psi_i = k$ and $\sum_{i=k+1}^{d} \psi_i = \alpha m $ respectively with probability at least $1-\frac{1}{100}\frac{1}{m^4}-2\exp(-c_1 m^{1/4})$
    \begin{align*}
        \SL &\le 40 k \alpha\log^2 m \TL + \frac{1}{m} (\log m + \log \frac{1}{\TL})^2 \TL + \frac{1}{m} \TL 
    \end{align*}
    where the constant $40$ comes from the constant $8$ from the upper bound and the constant from $\log \frac{1}{\delta}$ by setting $\delta = 100m^4$.
    Note now that since $\TL$ is lower bounded by $\frac{m^{-\eta}}{2k}$, we can combine $\log \frac{1}{\TL}$ and $\log m$ terms into 
    \begin{align*}
        \SL\le 50 k \frac{\log^2 m}{m^{\eta}}\TL.
    \end{align*}
    Here we bounded $(\log m + \log 2k)^2\le 2\log m$.
    This exactly gives with probability at least $1-\frac{1}{10}\frac{1}{m^4}-2\exp(-c_1 m^{1/4})-2\exp(-\frac{1}{4}(\alpha m^2)^{\epsilon_d})-2\exp(-\frac{1}{4}m^{\epsilon_m})$
    \begin{align*}
        \SL \le \tilde{O}(k \TL^2). 
    \end{align*}
    By setting $\epsilon_d<\frac{1}{8}$ and $\epsilon_m<\frac{1}{8}$, we get that by union bound over all $m$, with probability at least $0.99$ for all $m$
    \begin{align*}
        \SL \le \tilde{O}(k \TL^2). 
    \end{align*}
\end{proof}

In \Cref{thrm:app:diagfeatlowerbound} we will establish a case where we can provably lower bound the teacher loss.

\begin{theorem}[Lower bound for Teacher Loss]\label{thrm:app:diagfeatlowerbound}
    In \cref{ass:fstar} take $f^*=e_1$ and in \Cref{ass:linearnetwork} take $\langle g,e_i \rangle \sim N(0,\Lambda_m)$, where $\Lambda_m = \text{diag}(1,\frac{1}{m},\frac{1}{m},\dots,\frac{1}{m})$, where $\frac{1}{m}$ is repeated $N(m)$ times. If $N(m)= \alpha m^2$ then for all $\epsilon_d, \epsilon_m \in [0,\frac{1}{2}]$ with probability $1-2\exp(-\frac{1}{4}(\alpha m)^{\epsilon_d})-2\exp(-\frac{1}{4}m^{\epsilon_m})$ we have $\TL\ge 0.99 \frac{\alpha}{1+\alpha} - 2m^{-\frac{1}{2}+\epsilon_m} - (\alpha m)^{-\frac{1}{2}+\epsilon_d}\alpha$. Furthermore, as long as $\alpha m \left(\log \frac{1}{\delta}\right)^2 = \Omega(m^{-2+\epsilon_s})$ for some $\epsilon_s >0$, $1-\delta-2\exp(-c_1 m^{\frac{1}{4}})$, we have that 
    $\SL \le 8\alpha m (\log( \frac{1}{\delta}))^2 \TL \left(1 + O(m^{-\epsilon_s}) \right)$.
\end{theorem}

\begin{lemma}\label{lemma:app:highconcentrationlowerbound}
Let $\mG_{ij}=\sum_{i=2}^{\infty}\langle g_j, e_i \rangle \langle g_l, e_i \rangle$, let $\mH_{i,k} = \langle g_i , e_k \rangle \in \R^{m\times d}$, and let $\vg = \begin{pmatrix} \vdots \\
        \langle g_j,e_1 \rangle \\ \vdots
    \end{pmatrix}_{j=1}^{m} \sim N(0,\psi \mI_{m})$ be the rows of $\mH$, i.e. $\Lambda = \psi \mI_m$. 
Then with probability at least $1-2\exp(- \frac{d\delta^2}{4})$ for all $\vv \in \R^{m}$
\begin{align*}
    \vv^T\mG\vv \ge (1-\delta)tr(\Lambda) \vv^T \vv.
\end{align*}
\end{lemma}

\begin{proof}[Proof of \Cref{lemma:app:highconcentrationlowerbound}]

Note that $\mG=\mH\mH^T$ so we have that $\vv^T \mG \vv = \| \mH^T \vv||_2^2$. Note further that 
\begin{align*}
    \mH^T \vv  = \begin{pmatrix} \vdots \\
       \sum_{i=1}^{m} v_i \langle g_i,e_j \rangle \\ \vdots
    \end{pmatrix}_{j=1}^{d} \sim N(0, \psi \mI_{d}\| v\|_2^2 ),
\end{align*}
since the entries $\sum_{i=1}^{m} v_i \langle g_i,e_j \rangle$ are independent and each has variance $\psi \| v\|_2^2$.

Note now that 
\begin{align*}
    \| \mH^T \vv \|_2^2 = \sum_{j=1}^{d} \left( \sum_{i=1}^{m} v_i \langle g_i,e_j \rangle\right)^2 \sim \psi \|\vv\|_2^2 \left( Z_1^2+\dots + Z_d^2\right),
\end{align*}
where $Z_i \sim N(0,1)$ are iid, i.e. $\| \mH^T \vv \|_2^2\sim \psi \|\vv\|_2^2 X$ where $X\sim \chi^2_d$. By concentation of $\chi_d^2$ distribution, we have that with probability at least $1-2\exp(-\frac{d\delta^2}{4})$ that 
\begin{align*}
    \vv^T \mG\vv = \| \mH^T \vv\|_2^2 \ge (1-\delta) \psi d \vv^T\vv=(1-\delta) tr(\Lambda) \vv^T\vv.
\end{align*}
\end{proof}

\begin{proof}[Proof of \Cref{thrm:app:diagfeatlowerbound}]

Let $\hat{f}_{\hat \vw}$ be the teacher predictor.
Note that we can rewrite this as 
\begin{align*}
    \E_{\cD}\left(f^*-\hat{f}_{\vw} \right)^2 &= \E_{\cD} \left( e_1(x)^2 \left( 1-\sum_{j=1}^{m}w_j \langle g_j, e_1 \rangle \right)^2\right) + \E_{\cD} \left( \left(\sum_{j=1}^{m} w_j \sum_{i=2}^{\infty} \langle g_j,e_i \rangle e_i(x) \right)^2\right)\\
    &=1-2\sum_{j=1}^{m} w_j \langle g_j,e_1\rangle + \left( \sum_{j=1}^{m} w_j \langle g_j,e_1\rangle \right)^2 +\E_{\cD} \left( \left(\sum_{i=1}^{\infty} \sum_{j=1}^{m} w_j  \langle g_j,e_i \rangle e_i(x) \right)^2\right) \\
    &= 1-2\sum_{j=1}^{m} w_j \langle g_j,e_1\rangle + \left( \sum_{j=1}^{m} w_j \langle g_j,e_1\rangle \right)^2 +\sum_{i=2}^{\infty} \E_{\cD} \left( \left( \sum_{j=1}^{m} w_j  \langle g_j,e_i \rangle \right)^2\right) \\
    &= 1-2\sum_{j=1}^{m} w_j \langle g_j,e_1\rangle + \left( \sum_{j=1}^{m} w_j \langle g_j,e_1\rangle \right)^2 +\sum_{i=2}^{\infty}  \left( \sum_{j=1}^{m} w_j  \langle g_j,e_i \rangle \right)^2 \\
\end{align*}
Let $[\mG]_{jl}=\sum_{i=2}^{\infty} \langle g_j,e_i\rangle \langle g_l,e_i\rangle$ and $\vg = \begin{pmatrix} \vdots \\
        \langle g_j,e_1 \rangle \\ \vdots
    \end{pmatrix}_{j=1}^{m}$. 

Now, we can rewrite the above equation as follows.   
    \begin{align*}
    \cL(\hat{f}_{\vw}) &= 1-2\sum_{j=1}^{m} w_j \langle g_j,e_1\rangle + \left( \sum_{j=1}^{m} w_j \langle g_j,e_1\rangle \right)^2 +\sum_{i=2}^{\infty}  \left( \sum_{j=1}^{m} w_j  \langle g_j,e_i \rangle \right)^2 \\
    &= 1-2 \vw^T \vg +(\vw^T \vg)^2+ \vw^T \mG \vw .
    \end{align*}
    By \Cref{lemma:app:highconcentrationlowerbound}, we have that with probability at least $1-2\exp(- \frac{d\delta^2}{4})$ that
    \begin{align*}
         \vw^T \mG \vw \ge (1-\delta) tr(\Lambda) \vw^T \vw,
    \end{align*}
    where $d=N(m)$. Therefore, we have that 
\begin{align*}
    \cL(\hat{f}_{\vw}) &= 1-2 \vw^T \vg +(\vw^T \vg)^2+ \vw^T \mG \vw \ge 1-2\vw^T\vg+(\vw^T\vg)^2+(1-\delta)tr(\Lambda) \vw^T\vw.
    \end{align*}
    
 The last expression is minimized for $\vw = \frac{1}{\|\vg\|_2^2+(1-\delta)tr(\Lambda)} \vg$. Note that we have $tr(\Lambda) = \alpha m$.
 Note further that $\|\vg\|_{2}^2 \sim \chi_m^2$, so it concetrates around $m$, i.e. with probability at least $1-2\exp(-\frac{\delta_m^2 m}{4})$ we have 
 \begin{align*}
    (1+\delta_m) m\ge  \| \vg \|_2^2 \ge (1-\delta_m) m.
 \end{align*}
 So overall, we have that with probability at least $1-2\exp(- \frac{d\delta^2}{4})-2\exp(- \frac{m\delta_m^2}{4})$
\begin{align*}
     \cL(\hat{f}_{\vw}) &\ge \left( 1-\frac{\| \vg \|_2^2}{\|\vg\|_2^2+(1-\delta)tr(\Lambda)}   \right)^2  + (1-\delta)tr(\Lambda)\left(\frac{1}{\|\vg\|_2^2+(1-\delta)tr(\Lambda)}\right)^2 \|\vg \|_2^2 \\
     &\ge \left(1-\frac{m(1+\delta_m)}{m(1+\delta_m)+(1-\delta)\alpha m}\right)^2 + (1-\delta) \alpha \left(\frac{1}{(1+\delta_m)+(1-\delta)\alpha }\right)^2(1-\delta_m)\\
     &= \frac{(1-\delta)\alpha}{1+\delta_m+(1-\delta)\alpha}-\frac{2\delta_m}{1+\delta_m + (1-\delta) \alpha}.
\end{align*}

\Cref{thrm:app:helpingupperbound} in this case says that for $J=1$ we have with probability $1-\delta-2\exp(-c_1 m^{\frac{1}{4}})$ 
\begin{align*}
    1-\kappa \le 8\frac{1\cdot \frac{1}{m} N(m)}{m} (\log( \frac{1}{\delta}))^2  = 8\alpha m (\log( \frac{1}{\delta}))^2 .
\end{align*}

Therefore \Cref{thm:general-main} implies that with probability at least $1-\delta-2\exp(-c_1 m^{\frac{1}{4}})$ with $T= \log ( \frac{1}{\delta_T})$
\begin{align*}
    \SL \le 8\alpha m (\log( \frac{1}{\delta}))^2 \TL + \frac{1}{m^2} (\log( \frac{1}{\delta_T}))^2 \TL+  \frac{\delta_T}{2-\delta_T} \| f^* \|_2^2.
\end{align*}
For $\delta_T = \frac{1}{m^2} \TL$, i.e. for stopping time $T=2 \log m +\log \frac{1}{\TL}$ 
\begin{align*}
    \SL \le 8\alpha m (\log( \frac{1}{\delta}))^2 \TL \left(1 + \frac{2\log^2 m + \log^2\frac{1}{\TL}}{m^2 \cdot \alpha m  (\left(\log \frac{1}{\delta}\right)^2} \right).
\end{align*}
As long as $\alpha m \left(\log \frac{1}{\delta}\right)^2 = \Omega(m^{-2+\epsilon_{s}})$ for some $\epsilon >0$ this can be simplified to
\begin{align*}
    \SL \le 8\alpha m (\log( \frac{1}{\delta}))^2 \TL \left(1 + O(m^{-\epsilon_s}) \right).
\end{align*}
Note that we can simplify the lower bound for $\TL$, i.e. we have that with probabilit at least $1-2\exp(- \frac{d\delta^2}{4})-2\exp(- \frac{m\delta_m^2}{4})$
\begin{align*}
    \TL \ge \frac{\alpha}{1+\alpha+\delta_m-\delta\alpha} - 2\delta_m - \delta\alpha.
\end{align*}
As long as $\frac{1}{100} \ge \delta_m -\delta \alpha$, this can be further simplified to 
\begin{align*}
    \TL \ge 0.99\frac{\alpha}{1+\alpha} - 2\delta_m - \delta\alpha.
\end{align*}
Therefore, taking $\delta = \frac{1}{d^{\frac{1}{2}-\epsilon_d}}$ and $\delta_m = \frac{1}{m^{\frac{1}{2}-\epsilon_m}}$, we have that with probability at least $1-2\exp(-\frac{1}{4}(\alpha m^2)^{\epsilon_d})-2\exp(-\frac{1}{4}m^{\epsilon_m})$ that 
\begin{align*}
    \TL \ge 0.99 \frac{\alpha}{1+\alpha} - 2m^{-\frac{1}{2}+\epsilon_m} - (\alpha m^2)^{-\frac{1}{2}+\epsilon_d}\alpha.
\end{align*}
\end{proof}

\section{Weak-to-Strong Bound Applied to the Case of $2$-layer NN}\label{app:2layer}

This section deals with the cases of \Cref{ass:relu} and \Cref{ass:2layer}. 

\begin{proof}[Proof of \Cref{thrm:main:2layerrelu}]

    In \Cref{thrm:app:2layerReLUgeneral}, if we select $\delta_T = \frac{1}{\sqrt{\MT}} \TL$, then with stopping time $T=\frac{1}{\sigma_k^2} (\log \MT+ \log \frac{1}{\TL})$ we have that with probability $1-\frac{2}{\MT}-K\exp(-c\frac{\MT}{K})$ 
    \begin{align*}
        \SL&\le O_{d,k}\left(\frac{1}{\sqrt{\MT}} \left(\log \MT \right)^2 \right) \TL \\
        &+O_{d,k}\left(\frac{1}{\sigma_k^4}\frac{1}{\sqrt{\MT}}\right)\left(\log\MT - \log \TL \right)^2\TL+\frac{1}{\sqrt{\MT}}\TL\|f^*\|_{\cD}^2.
    \end{align*}
    This will hold if $\MT / \log \MT > K = N_k+\dots+N_0$ so it suffices to have $\MT \ge (N_0+\dots+N_k)^2$. For our setup, this is $(N_0+\dots+N_k)^2=\Theta(d^{2k})$.
    From this we have that 
    \begin{align*}
        \SL&\le O_{d,k}\left(\frac{1}{\sqrt{\MT}} \left(\log \MT \right)^2 \right) \TL \\
        &+O_{d,k}\left(\frac{1}{\sigma_k^4}\frac{1}{\sqrt{\MT}}\right)\left(\log^2\MT + \log^2 \TL \right)\TL+\frac{1}{\sqrt{\MT}}\TL
    \end{align*}
    where we used the fact that the target is nomalized.
    Note now that from \Cref{lemma:app:sizeofsigmak}, we have that $\sigma_k$ can be absorbed into the $O_{d,k}$. So we have that 
    \begin{align*}
        \SL&\le O_{d,k}\left(\frac{1}{\sqrt{\MT}} \left(\log \MT \right)^2 \right) \TL \\
        &+O_{d,k}\left(\frac{1}{\sqrt{\MT}}\right)\log^2 \TL \TL.
    \end{align*}
    This can be written as 
    \begin{align*}
        \SL\le  O_{d,k}\left( \frac{\log^2 \MT+\log^2 \TL}{\sqrt{\MT}} \, \TL \right).
    \end{align*}
    Now, to get the right exponent, we need a sharp lower bound on $\TL$. We can get that under the Gaussian Universality Ansatz. Under Gaussian Universality Ansatz, the teacher's test risk $\TL$ behaves like the deterministic equivalent \Cref{eq:app:det-eq}. Therefore, \Cref{thrm:app:diagfeatlowerbound} applies, which tells us that 
    \begin{align*}
        \TL \ge \Theta\left( \MT^{-\alpha} d^{2K} \|f\|_{\cD}^2\right).
    \end{align*}
    In other words, $\MT^{-1} \le O_{d,k}(\TL^{\frac{1}{\alpha}})$.
    
    Note that under Gaussian Universality, since $\TL$ is lower bounded polyonomially in $\MT$, we can absorb $\log^2\TL$ into $\log^2 \MT$. Therefore, we get 
    \begin{align*}
        \SL&\le O_{d,k}\left(\frac{1}{\sqrt{\MT}} \left(\log \MT \right)^2 \right) \TL. \\
    \end{align*}
    So we have that 
    \begin{align*}
        \SL\le O_{d,k}\left(\frac{\log^2 \MT}{\sqrt{\MT}}\right) \TL \le \tilde{O}_{d,k}\left(\TL^{1+\frac{1}{2\alpha}} \right).
    \end{align*}
    Since we can take $\alpha$ to be close to $\frac{d+2}{d-2}$, for $d>200$ we will have this be $1.02$ so the final exponent is $1.49$. Finally, note that here $K$ is the number of nonzero $\sigma_k$ up to index $k$, so by \Cref{cor:sizeofsigmaknk}, it is of order $\Theta(d^k)$. since $c$ is an absolute constant, for $\MT>\Theta(d^{2k})$,  $K\exp(-c\frac{\MT}{d^k})<\frac{1}{\MT}$ and otherwise we can have the constant in $O_{d,k}$.
\end{proof}

\begin{proof}[Proof of \Cref{thrm:main:2layerNNgeneral}]
    This is given by \Cref{thrm:app:2layerNNMaster} for $t_A = \frac{1}{2} \sqrt{\frac{\MT}{K}}$. Note that since $\MT\ge K$, \Cref{ass:rankA} holds with probability one. 
\end{proof}

\begin{theorem}[Master Bound for Weak-to-Strong Generalization with 2-layer NN]\label{thrm:app:2layerNNMaster}
    Under \Cref{ass:relu}, if \Cref{ass:fstar} is satisfied for $f^*$ and $K$. If $\sqrt{m}\ge t_A \sqrt{K}$ and the teacher attains minimum loss $\TL=\hat{\cL}_{\min}$, then the student trained to early stopping time $T=\frac{1}{\sigma_K^2} \log \frac{1}{\delta_T}$ will have with probability at least $1- 2\delta-2K\exp(-ct_A^2)$ 
    \begin{align*}
        \SL 
        &\le 
\inf_{S\ge K} \Biggl\{
    \frac{ m (\sum_{k=1}^{S} \sigma_k^2)(\sum_{k=S+1}^{\infty} \sigma_k^2 )  \log^2 m}{\sigma_{S}^4 (\sqrt{m}-t_A \sqrt{K})^2}\,\TL
    + \frac{\sigma_{S+1}^4}{\sigma_k^4} \left(\log \frac{1}{\delta_T}\right)^2 \,\TL\Biggr\}
    + \frac{\delta_T}{2-\delta_T} \,\bigl\|f^*\bigr\|_{\cD}^2.
    \end{align*}

\end{theorem}

\begin{proof}[Proof of \Cref{thrm:app:2layerNNMaster}]
Follows directly by instantiating \Cref{thm:general-main} with the lower bound on $\kappa_{S}$ given by \Cref{thrm:app:kappalowerbound2layer} and the lower bound on $\lambda_J(\mA)$ given by \Cref{thrm:app:Aeigenvaluelowerbound}. Finally, we just choose $T=\frac{1}{\sigma_K^2} \log \frac{1}{\delta_T}$.
\end{proof}
\subsection{Weak-to-Strong Bound Applied to the Case of $2$-layer ReLU NN}\label{app:sub:2layerReLU}

\begin{proof}[Proof of \Cref{thrm:main:2layerReLUgeneral}]
    This is given by \Cref{thrm:app:2layerReLUgeneral}.
\end{proof}

\begin{theorem}[Weak-to-Strong Generalization with 2-layer ReLU NN]\label{thrm:app:2layerReLUgeneral}
    Under \Cref{ass:relu}, if \Cref{ass:fstar} is satisfied for $f^*$ and $K$, let $k$ be the unique number such that $N_0+\dots+N_{k-1}+1\le K \le N_0+\dots+N_k$. If $\MT/\log \MT \ge K$ and the teacher  attains minimum loss $\TL=\hat{\cL}_{\min}$, then the student trained to early stopping time $T=\frac{1}{\sigma_k^2} \log \frac{1}{\delta_T}$ will have with probability at least $1-\frac{2}{\MT}-K\exp(-c\frac{\MT}{k})$ 
    \begin{align*}
        \SL&\le O_{d,k}\left(\frac{1}{\sqrt{\MT}} \left(\log \MT \right)^2 \right) \TL \\
        &+O_{d,k}\left(\frac{1}{\sigma_k^4}\frac{1}{\sqrt{\MT}}\right)\left(\log \frac{1}{\delta_T} \right)^2\TL+\frac{\delta_T}{2-\delta_T}\|f^*\|_{\cD}^2.
    \end{align*}
    Here $c$ is an absolute constant.
\end{theorem}
\begin{proof}[Proof of \Cref{thrm:app:2layerReLUgeneral}]
We will apply \Cref{thrm:app:2layerNNMaster}. Note that the eigenvalues of the activation functions $\sigma$ in this case are $\sigma_k$ and note that for odd $k>1$, we have $\sigma_k=0$. Note that we can just ignore the zero eigenvalues. So in this case, $\sigma_S$ will actually be equal to $\sigma_s$ and $\sigma_{S+1}$ to $\sigma_{s+2}$ if we select $S=N_0+\dots+N_s$. Take $s$ from \Cref{prop:app:optimals}. Then the proposition says that we have 
    \begin{align*}
        \frac{ m (\sum_{k=1}^{s} \sigma_k^2N_k)(\sum_{k=s}^{\infty} \sigma_k^2 N_k)  \left(\log \frac{1}{\delta}\right)^2}{\sigma_{s}^4 \left( \sqrt{m}-t_A \sqrt{K}\right)^4} &\le \Theta_d\left( \frac{ m \sqrt{m} \left(\log \frac{1}{\delta}\right)^2}{ \left( \sqrt{m}-t_A \sqrt{K}\right)^4} \right) \\
        \lambda_{S+1}^2T^2 = \frac{\sigma_{s+2}^4}{\sigma_k^4} \left(\log \frac{1}{\delta_T}\right)^2 &\le \Theta_d\left(\frac{1}{\sqrt{m}} \frac{1}{\sigma_k^4} \left(\log \frac{1}{\delta_T}\right)^2\right).
    \end{align*}
    Therefore, from \Cref{thrm:app:2layerNNMaster}, we have that with probability at least $1- 2\delta-2K\exp(-ct_A^2)$
    \begin{align*}
        \SL&\le \Theta_d\left( \frac{ m \sqrt{m} \left(\log \frac{1}{\delta}\right)^2}{ \left( \sqrt{m}-t_A \sqrt{K}\right)^4} \right) \TL \\
        &+\Theta\left(\frac{1}{\sqrt{m}} \frac{1}{\sigma_k^2} \left(\log \frac{1}{\delta_T}\right)^2\right)\left(\log \frac{1}{\delta_T} \right)^2\TL+\frac{\delta_T}{2-\delta_T}\|f^*\|_{\cD}^2.
    \end{align*}
    Take $\delta = \frac{1}{m}$ and take $t_a = m^{\frac{1}{2}}\frac{1}{2\sqrt{K}}$ so that $\sqrt{m}-t_A \sqrt{K}\ge\sqrt{m}/2$. Then $2K\exp(-ct_A^2)\le \frac{1}{m}$ if $\frac{m}{\log m}\ge K$. Note that therefore with probability at least $1-\frac{4}{m}$, we have 
    \begin{align*}
        \SL&\le \Theta_d\left(\frac{1}{\sqrt{\MT}} \left(\log \MT \right)^2 \right) \TL \\
        &+\Theta\left(\frac{1}{\sigma_k^4}\frac{1}{\sqrt{\MT}}\right)\left(\log \frac{1}{\delta_T} \right)^2\TL+\frac{\delta_T}{2-\delta_T}\|f^*\|_{\cD}^2,
    \end{align*}
    which completes the proof.
\end{proof}

\subsection{Proofs Related to $2$-layer NN}

\begin{proposition}[Optimal choice of $s$]\label{prop:app:optimals}
    We can take $s=m^{\frac{1}{4(d-4)}}+o_m(1)$ ($s$ is closest even integer to $m^{\frac{1}{4(d-4)}}$) so that 
    \begin{align*}
        \frac{ m (\sum_{k=1}^{s} \sigma_k^2N_k)(\sum_{k=s}^{\infty} \sigma_k^2 N_k)  \left(\log \frac{1}{\delta}\right)^2}{\sigma_{s}^4 \left( \sqrt{m}-t_A \sqrt{K}\right)^4} &\le \Theta_d\left( \frac{ m \sqrt{m} \left(\log \frac{1}{\delta}\right)^2}{ \left( \sqrt{m}-t_A \sqrt{K}\right)^4} \right) \\
        \lambda_{S+1}^2T^2 = \frac{\sigma_{s+2}^4}{\sigma_k^4} \left(\log \frac{1}{\delta_T}\right)^2 &\le \Theta_d\left(\frac{1}{\sqrt{m}} \frac{1}{\sigma_k^4} \left(\log \frac{1}{\delta_T}\right)^2\right).
    \end{align*}
    Note that this $\Theta$ hides absolute constants and $d$ dependence.
\end{proposition}
\begin{proof}
    Note that for $k\gg d$ we have that $N_k = \Theta(k^{d-2})$. This follows directly from the closed form $N_k =  \frac{(2k+d-2)(k+d-3)!}{k!(d-2)!}$. Furthermore, by \Cref{lemma:app:sizeofsigmak}, we have that $\sigma_k^2 = \Theta(k^{-(d+2)})$ for even $k\gg d$. Furthermore, $\sum_{k=1}^{\infty} \sigma_k^2N_k = \sigma(1)$. So, we have that 
    \begin{align*}
        \frac{ m (\sum_{k=1}^{s} \sigma_k^2N_k)(\sum_{k=s}^{\infty} \sigma_k^2 N_k)  \left(\log \frac{1}{\delta}\right)^2}{\sigma_{s}^4 \left( \sqrt{m}-t_A \sqrt{K}\right)^4} &\le \frac{1}{\sigma_s^4}\frac{ m \sigma(1)^2 \left(\log \frac{1}{\delta}\right)^2}{ \left( \sqrt{m}-t_A \sqrt{K}\right)^4} .\\
    \end{align*}
    Therefore, taking $s=m^{\frac{1}{4(d+2)}}+o_m(1)$, where $o_m(1)$ is taken so that $s$ is the closest integer to $m^{\frac{1}{4(d+2)}}$. Then since $\sigma_s^2 =\Theta(s^{-(d+2)}) $, we have that $\sigma_s^2 = \Theta(m^{-\frac{1}{4}})$, so $\frac{1}{\sigma_{s}^4}=\Theta(\sqrt{m})$. Note that $\sigma_{s+2}^2 = \Theta((m^{\frac{1}{4(d+2)}}+2)^{-(d+2)}) = \Theta( m^{-\frac{1}{4}})$. Therefore $\sigma_{s+2}^4 = \Theta(\frac{1}{\sqrt{m}})$.
\end{proof}


\begin{theorem}\label{thrm:app:kappalowerbound2layer}
    Under \Cref{ass:relu}, if \Cref{ass:fstar} is satisfied for $f^*$ and $K$, let $s$ be such that $N_0+\dots+N_{s-1}+1\le K \le N_0+\dots+N_s$. Then with probability at least $1-2\delta - 2K \exp(-ct_A^2)$
     \begin{align*}
          \frac{1}{\kappa_S}-1\le \lambda_1\!\left((\sqrt{\mA_S})^+ \mB_S (\sqrt{\mA_S})^+\right) \le \frac{\|F_B^{\dagger}F_A\|_{\cH}^2}{\lambda_S(\mA)^2} \le \frac{ m (\sum_{k=1}^{s} \sigma_k^2N_k)(\sum_{k=s}^{\infty} \sigma_k^2 N_k)  \left(\log \frac{1}{\delta}\right)^2}{\sigma_{s}^4 \left( \sqrt{m}-t_A \sqrt{K}\right)^4}
     \end{align*}
     where $c$ is an absolute constant and $\delta \in (0,1)$. For $s\ge 2$ then withe probability at least $1-2\delta - 2K \exp(-ct_A^2)$
     \begin{align*}
          \lambda_1\!\left((\sqrt{\mA_S})^+ \mB_S (\sqrt{\mA_S})^+\right) \le \frac{ m  \sigma(1)^2\left(\log \frac{1}{\delta}\right)^2}{\sigma_{s}^4 \left( \sqrt{m}-t_A \sqrt{K}\right)^4}.
     \end{align*}
\end{theorem}

\begin{proof}
    The first bound follows from \Cref{thrm:app:Aeigenvaluelowerbound} applied to $\lambda_{J}(\mA)$ and \Cref{prop:app:hilbertbound} applied with \Cref{prop:app:BcompReLUsecond}. This gives that with probability at least $1-2\delta - 2K \exp(-ct_A^2)$
    \begin{align*}
          \lambda_1\!\left((\sqrt{\mA_S})^+ \mB_S (\sqrt{\mA_S})^+\right) \le \frac{\|F_B^{\dagger}F_A\|_{\cH}^2}{\lambda_S(\mA)^2} \le \frac{ m (\sum_{k=1}^{s} \sigma_k^2N_k)(\sum_{k=s}^{\infty} \sigma_k^2 N_k)  \left(\log \frac{1}{\delta}\right)^2}{\sigma_{s}^4 \left( \sqrt{m}-t_A \sqrt{K}\right)^4}.
     \end{align*}
    
    The second bound follows by noting that $\sum_{k=1}^{\infty} \sigma_k^2N_k =\sigma(\langle \vx,\vx \rangle) =\sigma(1)$.
\end{proof}

In this case, the basis of eigenfunctions can be taken to be the spherical harmonics $\{ \{\phi_{k,i}\}_{i=1}^{N_k} \}_{k=1}^{\infty}$ where $\phi_{k,i}$ is the $i$-th spherical harmonic of order $k$ in $d$ dimensions and $N_k =  \frac{(2k+d-2)(k+d-3)!}{k!(d-2)!}$ is the dimensionality of spherical harmonics of order $k$ in $d$ dimensions.


\begin{proposition}[Eigendecomposition of the kernel induced by random features in the case of $2$-layer NN]\label{app:prop:eigendecomp2layer}
    Under \Cref{ass:relu}, the kernel induced by the random features is given by 
    \begin{align*}
        \cK(\vx, \vx') = \sum_{k=1}^{\infty} \sigma_k^2 \langle \phi_{k}(\vx), \phi_{k}(\vx') \rangle.
    \end{align*}
    Here, $\phi_k : \R^{d} \to \R^{N_k}$ is a vector of spherical harmonics of order $k$ in $d$ dimension, $N_k =  \frac{(2k+d-2)(k+d-3)!}{k!(d-2)!}$, and $\sigma_k$ is the $k$-th term in the decomposition of the ReLU functions $\sigma$ to the basis of $\{\phi_k\}_{i=1}^{\infty}$.
\end{proposition}

We provide the estimates for the size of $\sigma_k$ in \Cref{app:2layereigenval}.

Finally, we find $B$ and $L$ in the spherical case.
\begin{proposition}\label{prop:app:BcompReLU}
Under \Cref{ass:relu}, if $J = N_0+N_1+\dots+N_{s}$ then
    \begin{align*}
     \|P_{\le J} g\|_{\cD}^2 &= \sum_{k=1}^{s} \sigma_k^2N_k\\
     \|P_{\ge J+1} g\|_{\cD}^2 &= \sum_{k=s+1}^{\infty} \sigma_k^2N_k.
    \end{align*}
    So $L=B=\sqrt{\left(\sum_{k=1}^{s} \sigma_k^2N_k\right) \left( \sum_{k=s+1}^{\infty} \sigma_k^2N_k\right)}$.
\end{proposition}
\begin{proof}
    We have that 
    \begin{align*}
        P_{\le J}g &= \sum_{k=1}^{s} \sigma_k \phi_k(\vw)^T \phi_k(\vx)\\
        \|P_{\le J}g\|_{\cD}^2 &= \langle \sum_{k=1}^{s} \sigma_k \phi_k(\vw)^T \phi_k(\vx), \sum_{l=1}^{s} \sigma_l \phi_l(\vw)^T \phi_l(\vx) \rangle_{\cD} = \sum_{k=1}^{s} \sum_{l=1}^{s}\sigma_k \sigma_l \langle \phi_k(\vw)^T \phi_k(\vx), \phi_l(\vw)^T \phi_l(\vx)\rangle_{\cD}. \\
    \end{align*}
    Now since $\phi_k(\vw)^T \phi_k(\vx) = \sum_{j=1}^{N_k} \phi_{k,j}(\vw) \phi_{k,j}(\vx)$ and since $\{\phi_{k,j}\}$ are orthogonal w.r.t $\cD$, only the terms of the form $\phi_{k,j}(\vw)\phi_{k,j}(\vx)\phi_{k,j}(\vw)\phi_{k,j}(\vx)$ remain. For these, we have that $\E_{\cD}\left( \phi_{k,j}(\vw)\phi_{k,j}(\vx)\phi_{k,j}(\vw)\phi_{k,j}(\vx)\right) = \phi_{k,j}(\vw)^2 \cdot 1$. So we have that 
    \begin{align*}
        \|P_{\le J}g\|_{\cD}^2 &=  \sum_{k=1}^{s} \sigma_k^2  \langle \phi_k(\vw)^T \phi_k(\vx), \phi_k(\vw)^T \phi_k(\vx)\rangle_{\cD} \\
        &= \sum_{k=1}^{s} \sigma_k^2 \phi_{k,j}(\vw)^2= \sum_{k=1}^{s} \sigma_k^2 N_k.
    \end{align*}
    The last step holds because $\phi_k(\vw)^T\phi_k(\vw)=N_kP_k(\langle \vw, \vw \rangle)=N_k P_k(1) = N_k$. 

    The same argument shows that $\|P_{J+1}g\|_{\cD}^2 = \sum_{k=s+1}^{\infty}\sigma_k^2 N_k$.
\end{proof}

\begin{corollary}\label{prop:app:BcompReLUsecond}
    Under \Cref{ass:relu}, if $N_0+\dots+N_{s-1}+1\le J\le N_0+\dots+N_{s}$ then 
    \begin{align*}
        L = B =\sqrt{(\sum_{k=1}^{s} \sigma_k^2N_k)(\sum_{k=s}^{\infty} \sigma_k^2 N_k)}.
    \end{align*}
    This simplifies to $L=B\le \Theta(\frac{1}{d})\sqrt{\frac{2^s}{s!}}$.
\end{corollary}
\begin{proof}
    Note that if $J=N_0+N_1+\dots+N_{s-1}+p$ for some $1\le p\le N_s$ then we have that 
    \begin{align*}
        \|P_{\le J} g\|_{\cD}^2 &= \sum_{k=1}^{s-1} \sigma_k^2N_k+p\sigma_{s}^2\\
     \|P_{\ge J+1} g\|_{\cD}^2 &= (N_k-p)\sigma_s^2+\sum_{k=s+1}^{\infty} \sigma_k^2N_k.
    \end{align*}
\end{proof}

\subsection{Computation of the Kernel Eigenvalues for the case $2$-layer NN}\label{app:2layereigenval}

\begin{proof} of \Cref{app:prop:eigendecomp2layer}: For ReLU activation, we can do the following decomposition:
\begin{align*}
    \sigma(\inne{\vw}{\vx}) = \sum_{k \ge 0} N_k \sigma_k P_{d,k}(\inne{\vw}{\vx}) = \sum_{k \ge 0} \sigma_k \inne{\vphi_{d,k}(\vw)}{\vphi_{d,k}(\vx)},
\end{align*}
Here $\sigma_k$ is given by
\begin{align*}
    \sigma_k
    = \frac{\abs{\sphS^{d-2}}}{\abs{\sphS^{d-1}}} \int_0^{1} s P_{k,d}(s) (1 - s^2)^{\frac{d-3}{2}} \dd s.
\end{align*}
Combining this with $\abs{\sphS^{d-1}} = \frac{2\pi^{d/2}}{\Gamma(\frac{d}{2})}$,
we have
\begin{align*}
    \sigma_k
    = \frac{\Gamma(\frac{d}{2})}{\sqrt{\pi}\Gamma(\frac{d-1}{2})} \int_0^{1} s P_{k,d}(s) (1 - s^2)^{\frac{d-3}{2}} \dd s.
\end{align*}

The eigendecomposition of the kernel in this case is given by 
\begin{align*}
    \cK(\vx,\vx')=\E_{\vw}[\sigma(\inne{\vw}{\vx})\sigma(\inne{\vw}{\vx'})]
    = \E_{\vw}\left[ \sum_{k \ge 0} \sigma_k \inne{\vphi_k(\vw)}{\vphi_k(\vx)}  \cdot \sigma_k \inne{\vphi_k(\vw)}{\vphi_k(\vx')} \right]
    = \sum_{k \ge 0} \sigma_k^2 \inne{\vphi_k(\vx)}{\vphi_k(\vx')}.
\end{align*}

\end{proof}

Now, we find the size of $\sigma_k$.

\begin{lemma}[Size of $\sigma_k$]\label{lemma:app:sizeofsigmak}
    Depending on the size of $d$ and $k$, the following holds for the size of coefficients of ReLU activation in the basis of Legendre polynomials $\sigma_k = \frac{\Gamma(\frac{d}{2})}{\sqrt{\pi} \Gamma(\frac{d-1}{2})} \cdot \frac{1}{2^k} \frac{1}{(\frac{k}{2}+\frac{d-1}{2})\dots (\frac{d-1}{2})}$ for even $k$, $\sigma_1=\frac{1}{2d}$, and $\sigma_{k}=0$ for odd $k>1$. Asymptotically, this means
    \begin{enumerate}
        \item $\sigma_0 = \frac{1}{\sqrt{2\pi d}}+\Theta(d^{-\frac{3}{2}})$, $\sigma_1 = \frac{1}{2d}$
        \item If $k=\Theta(1)$, then for even $k$, $\sigma_k = \Theta(d^{-\frac{k+1}{2}})$.
        \item If $k\gg d$, then for even $k$, $\sigma_k = \Theta(k^{-\frac{d-1}{2}-\frac{3}{2}})$.
    \end{enumerate}
\end{lemma}

\begin{proof}
    The third case is shown in \cite{petrini2022}. To prove the general claim, we use the formula derived in \cite{petrini2022}
    \begin{align*}
    \sigma_k = \frac{\Gamma(\frac{d}{2})}{\sqrt{\pi} \Gamma(\frac{d-1}{2})} \cdot \frac{\Gamma(\frac{d-1}{2})}{ \Gamma(k+\frac{d-1}{2})} \left( - \left( -\frac{1}{2}\right)^k \right) \frac{d^{k-2}}{dt^{k-2}}(1-t^2)^{k+\frac{d-3}{2}}\Big|_{0}^{1}.
    \end{align*}
    Note that all terms will be multiplied by $(1-t^2)^{2+\frac{d-3}{2}}$ which is $0$ at $t=1$, so we only need to consider $t=0$. Note that if a term has a nonzero power of $t$, it will be evaluated to $0$. When taking $j$-th ($j\le k-2$) derivative of the expression above, all the terms will be in the form of $(1-t^2)^{l+\frac{d-3}{2}}t^s$ times some constant that depends on $d$ for some integers $l$ and $s$. So if out of $k-2$ derivatives we take $a$ on the term with $(1-t^2)$ and $k-2-a$ on $t^s$ (note that the ordering doesn't matter as we only add constants), in order to have term with no $t$ at the end, we need to have $k-2-a=s$, but $s=a$ so we have that $a=\frac{k-2}{2}$. Therefore, the evaluated derivate equals exactly 
    \begin{align*}
       \frac{d^{k-2}}{dt^{k-2}}(1-t^2)^{k+\frac{d-3}{2}}\Big|_{0}^{1} =  \left(k+\frac{d-3}{2}\right)\dots \left(k-(\frac{k-2}{2}-1) + \frac{d-3}{2}\right) = \left(k+\frac{d-3}{2}\right)\dots \left(\frac{k+4}{2} + \frac{d-3}{2}\right).
    \end{align*}
    Note that $\frac{\Gamma(\frac{d-1}{2})}{ \Gamma(k+\frac{d-1}{2})} = \frac{1}{(k-1+\frac{d-1}{2})\dots (\frac{d-1}{2})}$, therefore we can compute that 
    \begin{align*}
        \sigma_k &= \frac{1}{2^k}\frac{\Gamma(\frac{d}{2})}{\sqrt{\pi} \Gamma(\frac{d-1}{2})} \cdot \frac{1}{(k-1+\frac{d-1}{2})\dots (\frac{d-1}{2})}\left(k+\frac{d-3}{2}\right)\dots \left(\frac{k+4}{2} + \frac{d-3}{2}\right)\\
        \sigma_k &= \frac{\Gamma(\frac{d}{2})}{\sqrt{\pi} \Gamma(\frac{d-1}{2})} \cdot \frac{1}{2^k}\frac{1}{(\frac{k}{2}+\frac{d-1}{2})\dots (\frac{d-1}{2})}.\\
        \sigma_k &= \frac{\Gamma(\frac{d}{2})}{\sqrt{\pi} \Gamma(\frac{d-1}{2})} \cdot \frac{1}{2}\frac{1}{(k+d-1)((k-2)+d-1)((k-4)+d-1)\dots (2+d-1)(d-1)}.\\
    \end{align*}
    So, when $k = \Theta(1)$ we have that this is $\sigma_k = \Theta(d^{\frac{1}{2}} d^{-\frac{k+2}{2}}) = \Theta(d^{-\frac{k+1}{2}})$, since the first term is $\frac{\Gamma(\frac{d}{2})}{\sqrt{\pi} \Gamma(\frac{d-1}{2})}=\Theta(d^{\frac{1}{2}})$.
\end{proof}

\begin{corollary}[Size of $\sigma_k^2N_k$]\label{cor:sizeofsigmaknk}
    The following bound holds for $\sigma_k^2N_k$ for $k$ even or $k=1$
    If $k\gg d $ then $\sigma_k^2N_k=\Theta(k^{-4})$. If $k\ll d$ then $\sigma_k^2N_k = \Theta(d^{-1})$. If $k>1$ is odd, then $\sigma_k^2N_k=0$.
\end{corollary}

\begin{proof}
    We have for $\sigma_k^2N_k$ that 
    \begin{align*}
        \sigma_k^2 N_k &= \frac{1}{4}\left(\frac{\Gamma(\frac{d}{2})}{\sqrt{\pi} \Gamma(\frac{d-1}{2})}\right)^2 \cdot  \frac{(d-1)(d-1+1)\dots (d-1+k-2)}{(k+d-1)^2\dots (d-1+2)^2 (d-1)^2} (2k+d-2) \\
        & = \frac{1}{4}\left(\frac{\Gamma(\frac{d}{2})}{\sqrt{\pi} \Gamma(\frac{d-1}{2})}\right)^2 \cdot  (2k+d-2) \frac{(d-1+1)(d-1+3)\dots (d-1+k-3)}{(d-1)(d-1+2)\dots (d-1+k-4)(d-1+k-2)} \frac{1}{(d-1+k)^2}.\\
    \end{align*}
    When $k\ll d$, we can just count the terms with $d$ in the numerator and denominator and note that $\left(\frac{\Gamma(\frac{d}{2})}{\sqrt{\pi} \Gamma(\frac{d-1}{2})}\right)^2 = \Theta(d)$. For $k\gg d$, we just need to count the terms with $k$ in the numerator and denominator. The results follow.

\end{proof}

\removed{
\marko{Does this assume that the student is smaller in every direction?}
\marko{Merge with F.1}
\begin{theorem}[Limitation of Weak-to-Strong Generalization with a too weak teacher]\label{thm:app:limitationw2s}
    Under \Cref{defi:shrinkage_optimality} and \Cref{ass:studentbounded}, the following inequality holds 
    \begin{align*}
        \TL \le \SL + 2\sqrt{\SL }\sqrt{1-\TL}.
    \end{align*}
\end{theorem}}

\section{Proofs of Limitation of Weak-to-Strong Generalization}\label{app:limitation}

\begin{proof}[Proof of \Cref{thrm:main:lowerbound}]
This follows from \Cref{thrm:main:generallimit} by noting that from \Cref{lemma:gf_shrinkage_optimal} the teacher satisfies \Cref{defi:shrinkage_optimality} with respect to the target and the student satisfies \Cref{defi:shrinkage_optimality} with respect to the teacher.
\end{proof}

\begin{proof}[Proof of \Cref{lemma:gf_shrinkage_optimal}] We rewrite the claim of \Cref{lemma:gf_shrinkage_optimal} as follows.
    \begin{lemma}[Shrinkage Optimality of Gradient Flow Solutions]
    For any positive semi-definite kernel $\mathcal{K}$, time $T > 0$, and ground truth $f^*$, the gradient flow solution $f_T = \mathcal{T}^{\mathcal{K}}_T(f^*)$ is shrinkage optimal with respect to $f^*$ in the sense of \Cref{defi:shrinkage_optimality}.
\end{lemma}
    From \Cref{eq:student-dynamics}, the gradient flow solution is $f_T = \sum_{k \ge 1} (1 - e^{-\lambda_k T}) \langle f^*,e_k\rangle_{\mathcal{D}}e_k$. For shrinkage optimality, we need to show that $\alpha = 1$ minimizes $\mathbb{E}_{\mathbf{x}}[(\alpha f_T(\mathbf{x})-f^*(\mathbf{x}))^2]$ for $0 \leq \alpha \leq 1$.
    
    Expanding this expression in the eigenbasis:
    \begin{align}
        \mathbb{E}_{\mathbf{x}}[(\alpha f_T(\mathbf{x})-f^*(\mathbf{x}))^2] &= \left\|\alpha\sum_{k \ge 1} (1 - e^{-\lambda_k T}) \langle f^*,e_k\rangle_{\mathcal{D}}e_k - \sum_{k \ge 1} \langle f^*,e_k\rangle_{\mathcal{D}}e_k\right\|^2_{\mathcal{D}} \\
        &= \sum_{k \ge 1} (\alpha(1 - e^{-\lambda_k T})-1)^2 \langle f^*,e_k\rangle^2_{\mathcal{D}}
    \end{align}
    The proof is immediate by noting that $1-e^{-\lambda_kT}\ge 0$ and $\alpha(1 - e^{-\lambda_k T})\le 1$ for all $k$ and $0\le \alpha \le 1$.
\end{proof}

\begin{proof}[Proof of \Cref{thrm:main:generallimit}]

The proof is a straightforward geometric argument in the function space. Note first that \Cref{defi:shrinkage_optimality} gives a restiction of where the teacher predictor $f_{\text{teacher}}$ can be in the function space with respect to the target $f^*$ and the origin.
Note that $\| f_{\text{teacher}} - f^*\|_{\cD}^2 \le \| \alpha f_{\text{teacher}} - f^*\|_{\cD}^2$ for all $0\le \alpha \le 1$ holds if and only if $\inne{f_{\text{teacher}}}{f_{\text{teacher}}-f^*}\le 0$. 
Similarlry, \Cref{defi:shrinkage_optimality} holds if and only if $\inne{f_{\text{student}}}{f_{\text{teacher}}-f_{\text{student}}}\le 0$. 
This implies that the teacher predictor $f_{\text{teacher}}$ is inside the set $f_{\text{teacher}}\in \{ f \mid \| f-\frac{f^*}{2}\|_{\cD}\le \frac{1}{2}\|f^*\|_{\cD}\}$. 
Similarly, we have that $f_{\text{student}} \in \{ f \mid \| f-\frac{f_{\text{teacher}}}{2}\|_{\cD}\le \frac{1}{2}\|f_{\text{teacher}}\|_{\cD}\}$.
Both of these sets are spheres in the function space w.r.t. $\cD$-norm. 
Let $\beta\ge1$ be such that $\tilde{f}_{\text{teacher}}=\beta f_{\text{teacher}}$ and such that $\inne{\tilde{f}_{\text{teacher}}}{f^*-\tilde{f}_{\text{teacher}}}=0$. Then we have that 
\begin{align*}
    \TL = 1-\beta^2 f_{\text{teacher}}^2+(\beta-1)^2 f_{\text{teacher}}^2=1-(2\beta-1)f_{\text{teacher}}^2.
\end{align*}
Let $\tilde{f}_{\text{student}}$ the predictor inside the student set with the smallest risk, i.e. on the intersection of the line connecting $f^*$ and $\frac{f_{\text{teacher}}}{2}$ and the boundary of the allowed student set. Note then that 
\begin{align*}
    \cL(\tilde{f}_{\text{student}}) = \left(\| f^* - \frac{f_{\text{teacher}}}{2}\|_{\cD}-\|\frac{f_{\text{teacher}}}{2}\|_{\cD} \right)^2
\end{align*}
Note further that 
\begin{align*}
    \| f^* - \frac{f_{\text{teacher}}}{2}\|_{\cD}^2 = 1-\beta^2 f_{\text{teacher}}^2+(\beta-\frac{1}{2})^2 f_{\text{teacher}}^2 = 1-(\beta-\frac{1}{4})f_{\text{teacher}}^2.
\end{align*}
Therefore, we have that 
\begin{align*}
    f_{\text{teacher}}^2 = \frac{1-\TL}{2\beta-1}.
\end{align*}
From this we get that 
\begin{align*}
     \cL(\tilde{f}_{\text{student}}) &= \left(\sqrt{1-(\beta-\frac{1}{4})\frac{1-\TL}{2\beta-1}} - \frac{1}{2}\sqrt{\frac{1-\TL}{2\beta-1}} \right)^2\\
     &= \frac{1}{8\beta-4}\left(\sqrt{4\beta-3+(4\beta-1)\TL} - \sqrt{1-\TL} \right)^2
\end{align*}
Note that for $\beta \ge 1$ and $\TL\le 1$ we have that 
\begin{align*}
     \cL(\tilde{f}_{\text{student}}) &= \frac{1}{8\beta-4}\left(\sqrt{4\beta-3+(4\beta-1)\TL} - \sqrt{1-\TL} \right)^2\\
     &\ge \frac{1}{2} \left(1 + \TL - \sqrt{1 + 2\TL - 3\TL^2} \right).
\end{align*}
The last inequality holds by noticiting that the function is increasing in $\beta$, so it is minimized for $\beta= 1$ for $0<\TL<1$.

That is, since $\SL \ge \cL(\tilde{f}_{\text{student}})$ by design, we have that 
\begin{align*}
    \SL \ge \frac{1}{2} \left(1 + \TL - \sqrt{1 + 2\TL - 3\TL^2} \right) \ge \frac{3}{4} \TL^2.
\end{align*}
    The last inequality $\frac{1}{2} \left(1 + \TL - \sqrt{1 + 2\TL - 3\TL^2} \right)\ge \frac{3}{4} \TL^2$ holds for $0\le\TL\le 1$.
\end{proof}
\removed{For the proof of \Cref{thrm:main:boostinglimit}, we will show the case for $n=1$ student first.
\begin{proposition}[Limitation of Weak-to-Strong generalization with a bounded student]\label{thrm:app:onestudentlimit}
    Consider the weak-to-strong setup where the teacher satisfies \Cref{defi:shrinkage_optimality} and the student satisfies \Cref{ass:studentbounded}. Then we have that 
    \begin{align*}
        \SL \ge \frac{1}{4}\TL^2
    \end{align*}
\end{proposition}
\begin{proof}[Proof of \Cref{thrm:app:onestudentlimit}]
    \marko{Fill this.}
\end{proof}}

\begin{proof}[Proof of \Cref{thrm:main:normboundedlimit}]
    The proof is similar to the proof of \Cref{thrm:main:generallimit}. Similarly, we have that the teacher predictor is bounded to the set $f_{\text{teacher}}\in \{ f \mid \| f-\frac{f^*}{2}\|_{\cD}\le \frac{1}{2}\|f^*\|_{\cD}\}$. The student predictor is in this case bounded to the set $f_{\text{student}}\in \{f\mid \|f\|_{\cD}\le \| f_{\text{teacher}}\|_{\cD} \}$. Let $\beta = \inne{f^*-f_{\text{teacher}}}{f^*}$. Note that from the condition $f_{\text{teacher}}\in \{ f \mid \| f-\frac{f^*}{2}\|_{\cD}\le \frac{1}{2}\|f^*\|_{\cD}\}$ we have that $\inne{f_{\text{teacher}}}{f^*-f_{\text{teacher}}}\le 0$ which implies that there is $\beta \ge 1$ for which we have $\inne{\beta f_{\text{teacher}}}{f^*-f_{\text{teacher}}}$. Note that $\TL = \| f^*-f_{\text{teacher}}\|_{\cD}^2$. Therefore, we have that  since $\| f^*\|_{\cD}^2=1$
    \begin{align*}
        \TL = 1-\beta^2 f_{\text{teacher}}^2+(\beta-1)^2 f_{\text{teacher}}^2=1-(2\beta-1)f_{\text{teacher}}^2.
    \end{align*}
    This implies that 
    \begin{align*}
    f_{\text{teacher}}^2 = \frac{1-\TL}{2\beta-1}.
    \end{align*}
    Note that 
    \begin{align*}
        \SL \ge (\|f^*\|_{\cD}-\|f_{\text{teacher}}\|_{\cD})^2 = (1-\|f_{\text{teacher}}\|_{\cD})^2
    \end{align*} by the same argument as in \Cref{thrm:main:generallimit}. Therefore, we have that 
    \begin{align*}
        \SL \ge \left(1-\sqrt{\frac{1-\TL}{2\beta-1}} \right)^2
    \end{align*}
    Note that $1-\sqrt{\frac{1-\TL}{2\beta-1}} $ is positive and  increasing in $\beta$, so the above lower bound is minimized for $\beta=1$. Therefore
    \begin{align*}
        \SL \ge \left(1-\sqrt{1-\TL} \right)^2 = 2-\TL-2\sqrt{1-\TL}  \ge \frac{1}{4}\TL^2.
    \end{align*}
\end{proof}

\begin{proof}[Proof of \Cref{thrm:main:boostinglimit}]
    The proof is immediate from \Cref{thrm:main:normboundedlimit} and the fact that the student $f_{\text{student}}^{(i)}$ is shrinking-optimal w.r.t. $f_{\text{student}}^{(i-1)}$ implies that $\|f_{\text{student}}^{(i)}\|_\cD \le \|f_{\text{student}}^{(i-1)}\|_{cD}$, which further implies that $\|f_{\text{student}}^{(i)}\|_\cD \le \|f_{\text{teacher}}\|_{\cD}$.
\end{proof}

\section{Deterministic Equivalent of the Teacher Error}

Let $\tilde{\cL}(f^*,m)$ be the deterministic equivalent of the test risk of our model with $m$ random features trained on population loss with target $f^*$. Let $s_K$ be the eigenvalues of the activation function, i.e. the decompositon of the activation function in the eigenbasis $\{e_i\}_{i=1}^{\infty}$. Let $\beta^*$ be the coefficients of the target $f^*$ in the eigenbasis, $f^* = \sum_{i=1}^{\infty} \beta_i^* e_i$. To write down a closed form of $\tilde{\cL}(f^*,m)$, we introduce the following notation $\mS$ and $\nu$
\begin{align*}
    \mS &= \text{diag}(s_1,s_2,\dots)\\
    m&=\text{tr}(\mS(\mS+\nu)^{-1}).
\end{align*}
That is, $\nu$ is the unique solution to this equation. According to Corollary $3.5$ in \citet{defilippis2024}, the deterministic equivalent of test risk has the following closed form then 
\begin{align}\label{eq:app:det-eq}
    \tilde{\cL}(f^*,m) = \nu \langle \beta^*,(\mS+\nu)^{-1}\beta^* \rangle.
\end{align}

\subsection{Lower Bound for Teacher Error}

We show a lower bound on the deterministic equivalent of test risk of a Random Feature Model with $m$ features that we consider in our setup. Under the Gaussian Universality Ansatz, the true test error of the teacher's predictor $\TL$ will behave like its deterministic equivalent.

\begin{theorem}[Lower bound for the Error of a $2$-layer random feature ReLU]\label{thrm:app:detlowerbound}
    If $f^*$ satisfies \Cref{ass:fstar} for \Cref{ass:linearnetwork} with $K$ such that for some $k$, $N_0+\dots+N_{k-1}+1\le K \le N_0+\dots+N_k$, then for fixed $d$ there exists $\alpha>\frac{d+2}{d-2}$ such that for the deterministic equivalent of the test risk we have
    \begin{align*}
        \tilde{\cL}(f^*,m)\ge \Theta\left(\frac{1}{m^\alpha} d^{2} \| f^* \|_{\cD}^2\right).
    \end{align*}
    We can take $\alpha = \frac{d+2}{d-2}+o_d(1)$.
\end{theorem}

\begin{proof}[Proof of \Cref{thrm:app:detlowerbound}]
    Note that the eigenvalues of the activation function are $\sigma_s^2$ with multiplicity $N_s$.
    We want to lower bound $\nu$. Assume that for large $m$, $\nu\le \frac{1}{m^\alpha}$. Note that we have the following two things for $s\gg d$, 
    \begin{align*}
        \sigma_s^2 = \Theta( s^{-(d+2)} )\text{ and } N_s = \Theta(s^{d-2}).
    \end{align*}
    The first follows from \Cref{lemma:app:sizeofsigmak} and the second one follows immediately from the closed form of $N_s$.
    Note that in this case we have that $s_i = \sigma_i^2$.
    Note that then for $\sigma_s^2 > \nu$ we have that $\frac{\sigma_s^2}{\sigma_s^2+\nu}>\frac{1}{2}$. Note that if $\nu <\frac{1}{m^{\alpha}}$ then we have that for all $s<m^{\frac{\alpha}{d+2}}$ that $\sigma_s^2 > \nu$. Note that there is at least $N_{s_{\max}}$ of those, where $s_{\max}$ is the largest such $s$. We have that $s_{\max} \ge m^{\frac{\alpha}{d+2}}$ But we have that $N_{s_{\max}} = \Theta(s_{\max}^{d-2}) \ge \Theta( m^{\alpha\frac{d-2}{d+2}})$. Note also that with this choice of $s_{\max}$
    \begin{align*}
        m=\text{tr}(\mS(\mS+\nu)^{-1}) = \sum_{i=1}^{\infty} \frac{s_i}{s_i+\nu} = \sum_{s=1}^{\infty}\sum_{i=1}^{N_s} \frac{\sigma_{s,i}^2}{\sigma_{s,i}^2+\nu}\ge N_{s_{\max}} \frac{1}{2} \ge \Theta(m^{\alpha\frac{d-2}{d+2}}),
    \end{align*}
    where $\sigma_{s,i} = \sigma_s$.

    So if $\alpha=\frac{d+2}{d-2}+o_{d}(1)$ then we have a contradiction. Therefore $\nu \ge \frac{1}{m^{\alpha}}$ for this $\alpha=\frac{d+2}{d-2}+o_{d}(1)$. Plugging this back in we have 
    \begin{align*}
        \tilde{\cL}(f^*,m) =\nu \left( \sum_{i=1}^{\infty} \frac{1}{\sigma_i^2+\nu} \sum_{j=1}^{N_i} \beta_{i,j}^{*2} \right)\ge \frac{1}{2}\nu \frac{1}{\sigma_1^2+\frac{1}{m}} \| f^*\|_{\cD}^2.
    \end{align*}
    For the last inequality, note first that $\sigma_i\le \sigma_1$ so $\frac{1}{\sigma_i^2+\nu}\ge \frac{1}{\sigma_1^2+\nu}$. Also note that 
    \begin{align*}
        m = \sum_{i=1}^{\infty}\frac{s_i}{s_i+\nu} \le \frac{\sum_{i=1}^{\infty} s_i}{\nu},
    \end{align*}
    so $\nu \le \frac{\sum_{i=1}^{\infty} s_i}{m} $. Note that since $\sum_{s=1}^{\infty} \sigma_s^2 N_s = \sigma(1) = 1$ we have that $\sum_{i=1}^{\infty} s_i =1$. Therefore, it holds that $\nu \le \frac{1}{m}$. This finishes the proof.
\end{proof}

\subsection{ReLU Network Error Upper Bound}\label{app:gaussianthetaone}

\removed{\begin{proof}[Proof of \Cref{thrm:main:nothetaoneforrelu}]
    This follows from \Cref{thrm:app:upperboundreludet}. Note that under Gaussian Universality, this theorem holds for the test risk.
\end{proof}}

\begin{theorem}[Upper Bound on Teacher Error with ReLU Network]\label{thrm:app:upperboundreludet}
If $f^*$ is spanned by the spherical harmonics of order that is even and at mos $k$ or $1$, for \Cref{ass:relu} in fixed dimension $d$, we have the following bound on the deterministic equivalent of the test risk 
\begin{align*}
        \tilde{\cL}(f^*,m)\le \Theta\left(\frac{1}{m} d^{(k+1)} \| f^* \|_{\cD}^2\right)
    \end{align*}
\end{theorem}

\begin{proof}[Proof of \Cref{thrm:app:upperboundreludet}]
Note that 
\begin{align*}
     m=\text{tr}(\mS(\mS+\nu)^{-1}) = \sum_{i=1}^{\infty} \frac{s_i}{s_i+\nu} \le \frac{\sum_{i=1}^{\infty}s_i}{\nu} 
\end{align*}
so we have $\nu \le \frac{\sum_{i=1}^{\infty}s_i}{m}$. Since $\sum_{i=1}^{\infty} s_i = \sigma(1) = 1$, we have that $\nu\le \frac{1}{m}$.
Then we have that since $f^*$ is spanned by harmonics of order at most $k$
\begin{align*}
     \tilde{\cL}(f^*,m) =\nu \left( \sum_{i=1}^{\infty}\frac{1}{\sigma_i^2+\nu} \sum_{j=1}^{N_i} \beta_{i,j}^{*2} \right) = \nu \sum_{i=1}^{k} \frac{1}{\sigma_i^2+\nu} \sum_{j=1}^{N_s}\beta_{i,j}^{*2} \le \frac{\nu}{\sigma_k^2+\nu} \sum_{i=1}^{k}\sum_{j=1}^{N_i} \beta_i^{*2} = \frac{\nu}{\sigma_k^2+\nu} \| f^* \|_{\cD}^{2}.
\end{align*}
Finally, note that $\sigma_{k}^2 = \Theta(d^{-(k+1)})$ so we have $\frac{\nu}{\nu+\sigma_k^2} \le \frac{1}{m} \Theta(d^{(k+1)})$
\end{proof}
\begin{remark}
    The same proof shows that in the Gaussian Features \Cref{ass:app:daigfeat} if the covairance structure is fixed  we cannot have a $\Theta(1)$-error asymptotic for any function that is learnable by the model as $\MT\to \infty$ in fixed dimension. 
\end{remark}
\removed{
\marko{delete this for now}
\begin{theorem}[Lower bound on Teacher Error in ReLU with Increasing Dimension]
    Let $f^*$ be spanned by spherical harmonics of order at most $k$. If we take a sequence of ReLU models as \Cref{ass:relu} indexed by $d$ with $\MT = d^{2k}$ then there is $C_{T}>0$ such that for any $d$ we have
    \begin{align*}
        \TL \ge C_{T}.
    \end{align*}
\end{theorem}}

\section{Experiment Details}\label{sec:exp}
In this section, we provide experiment details for simulating weak-to-strong generalization for both 2-layer ReLU networks (\Cref{fig:exp:2layer-main}) and linear networks (\Cref{fig:exp:linear-main}). All experiments are conducted on one H100 GPU, and are finished within 24 GPU hours.

To avoid the effect of randomness, experiments are repeated $5$ times with different random seeds, 95\% confidence intervals are shown in \Cref{fig:exp:2layer-loss-ratio} and \Cref{fig:exp:linear-loss-ratio}. 

For curve fitting in \Cref{fig:exp:2layer-loss-fitting} and \Cref{fig:exp:linear-loss-fitting}, we use \texttt{numpy.polyfit} with degree $1$ after taking the log of teacher loss $\TL$ and student loss $\SL$. 

For linear network, we simulate the limit case where student size $\MS\to \infty$. The simulation follows \Cref{eq:student-dynamics} with $\lambda_1=1$, $\lambda_i=(d-1)^{-2/3}$ for $i>1$. Note this method is not viable for 2-layer ReLU network because $\lambda_i$ is long-tailed for 2-layer ReLU network. Therefore, we set $\MS=16384$ in \Cref{fig:exp:2layer-main}.

\section{Additional Experimental Results}\label{app:additiona-exp}

\subsection{2-Layer ReLU Network Experiments}
\begin{figure}[h]
    \centering
    \subfigure[$\SL/\TL$ v.s. $t$ at $d=16$]{\includegraphics[width=0.33\textwidth]{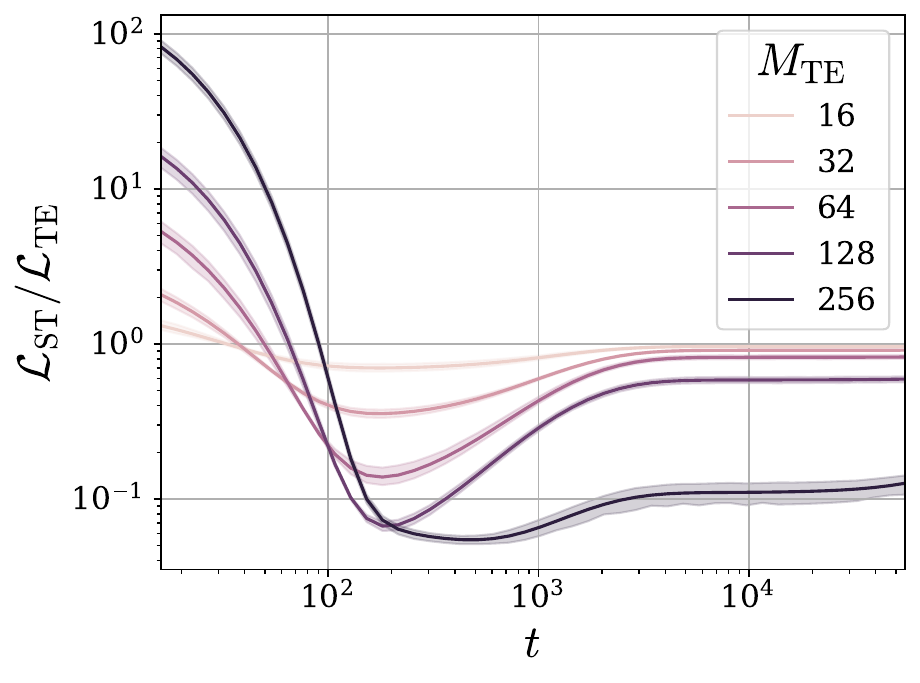}}
    \subfigure[$\SL/\TL$ v.s. $t$ at $d=32$]{\includegraphics[width=0.33\textwidth]{figures/gf_d_0=32_m=16384.pdf}}
    \subfigure[$\SL/\TL$ v.s. $t$ at $d=64$]{\includegraphics[width=0.33\textwidth]{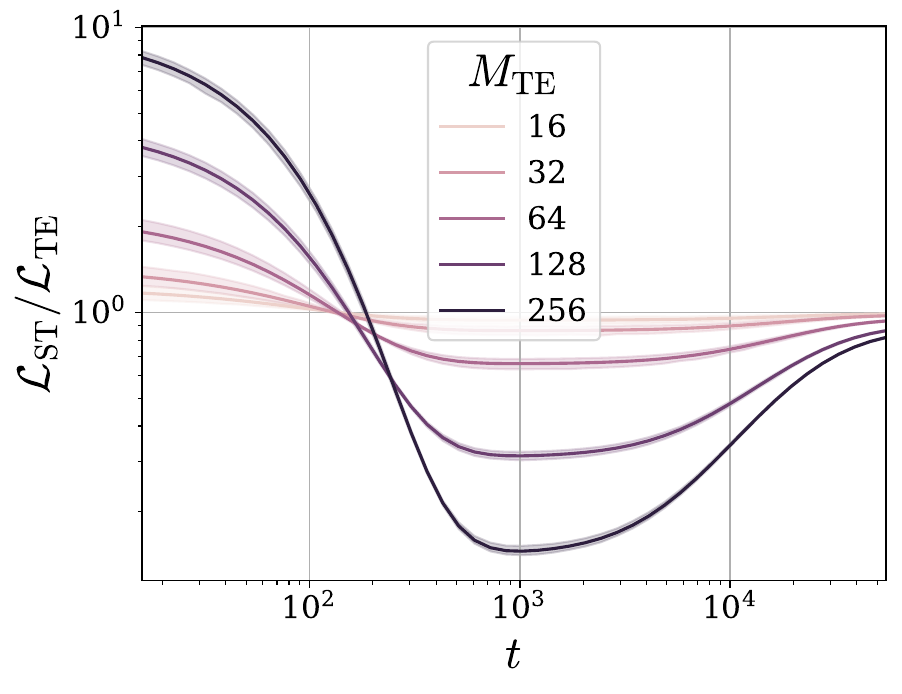}}
    \subfigure[$\SL$ v.s. $\TL$ at $d=16$]{\includegraphics[width=0.33\textwidth]{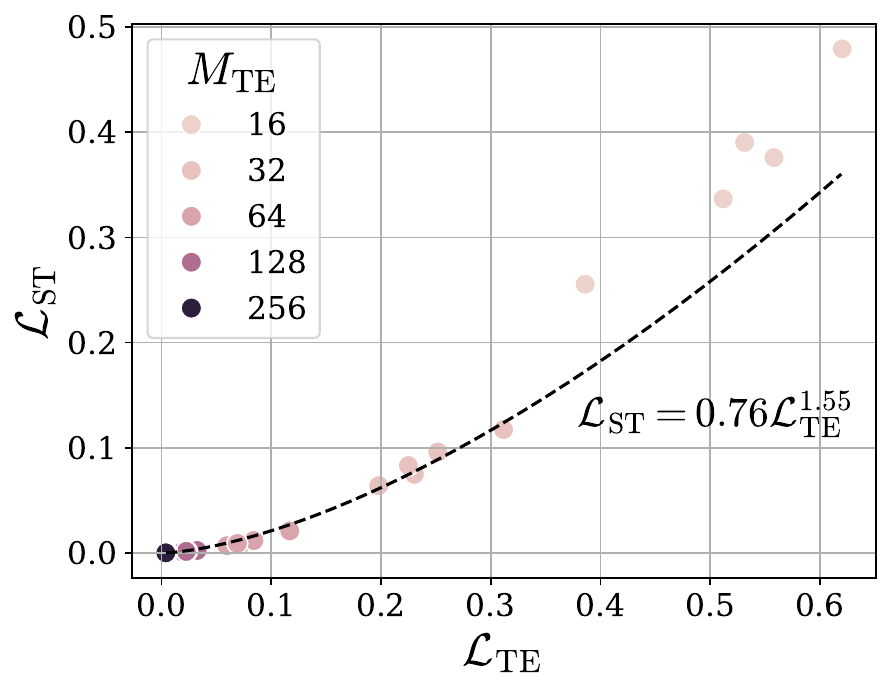}}
    \subfigure[$\SL$ v.s. $\TL$ at $d=32$]{\includegraphics[width=0.33\textwidth]{figures/gf_d_0=32_m=16384_lt_vs_ls.pdf}}
    \subfigure[$\SL$ v.s. $\TL$ at $d=64$]{\includegraphics[width=0.33\textwidth]{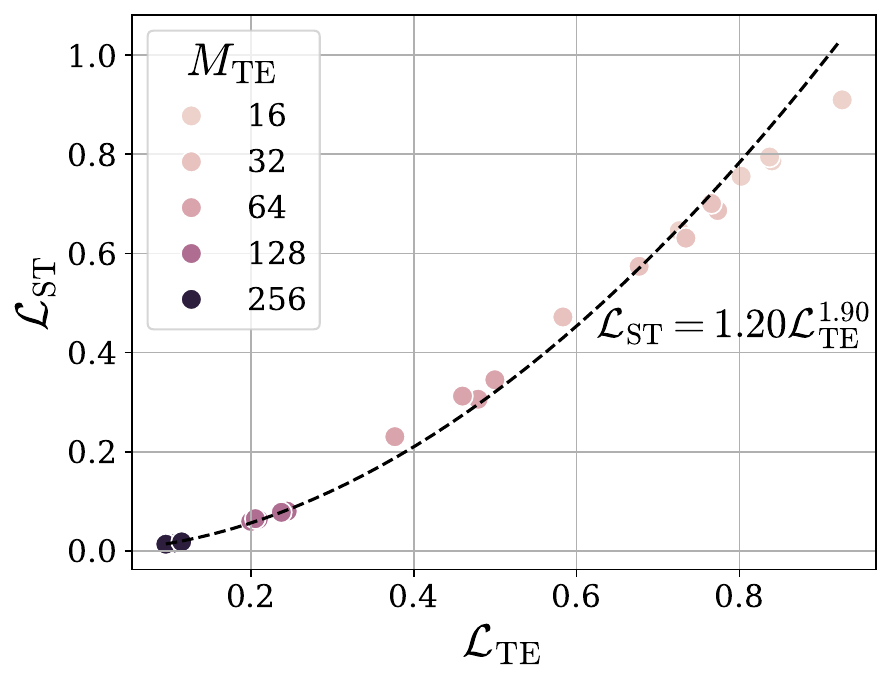}}
    \caption{Weak-to-strong generalization happens in 2-layer ReLU networks with input dimension $d=16,32,64$, student size $\MS=16384$, and teacher size $\MT\in\{16, \ldots, 256\}$. We consider target function $f^*$ be a linear function, i.e., $f^*= \langle \beta, \vx\rangle$ for some $\beta$ of unit norm.  The top figures plots the ratio between student loss $\SL$ and teacher loss $\TL$, with varying $\MT$ and gradient flow training time $t$. In bottom figures, we fit student loss $\SL$ as a power law function of $\TL$. The empirical observations align with \Cref{thrm:main:2layerrelu}.}
    \label{fig:exp:2layer-full}
\end{figure}

\Cref{fig:exp:2layer-full} shows full experimental results for \Cref{ass:relu}. It additionally includes $d=16$ and $64$ compared to \Cref{fig:exp:2layer-main}. We observe that when $d=64$, the student loss $\SL$ is polynomially smaller than the teacher loss $\TL$, with a higher estimated exponent compared to $d=32$. When $d=16$, the curve suggests a sudden change of optimal early stopping time, which we believe is caused by lack of student size $\MS$.

\begin{figure}[h]
    \centering
    \subfigure[$\MS=8192$, $d=16$]{\includegraphics[width=0.33\textwidth]{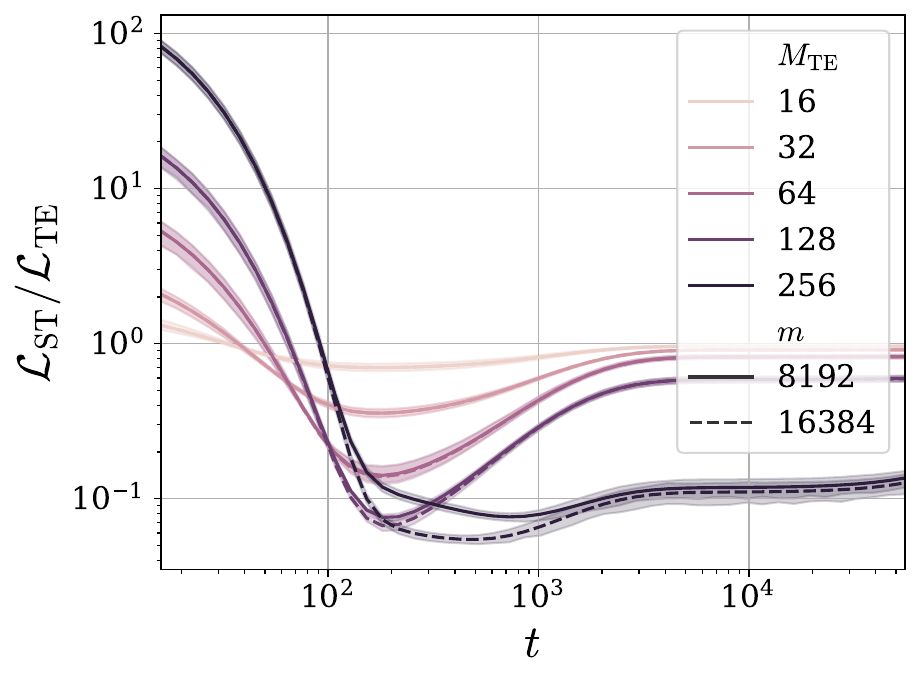}}
    \subfigure[$\MS=8192$, $d=32$]{\includegraphics[width=0.33\textwidth]{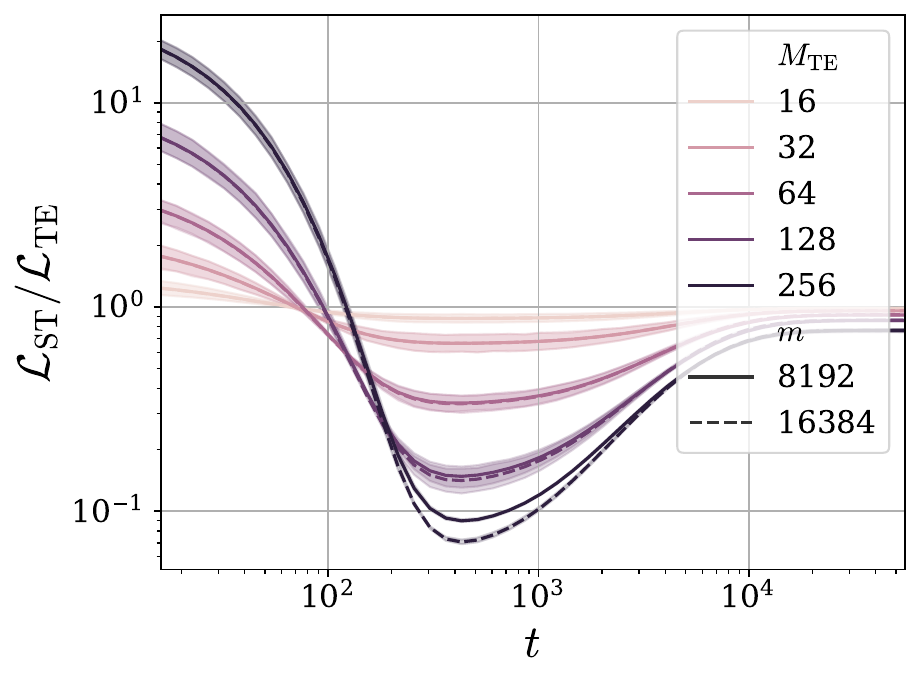}}
    \subfigure[$\MS=8192$, $d=64$]{\includegraphics[width=0.33\textwidth]{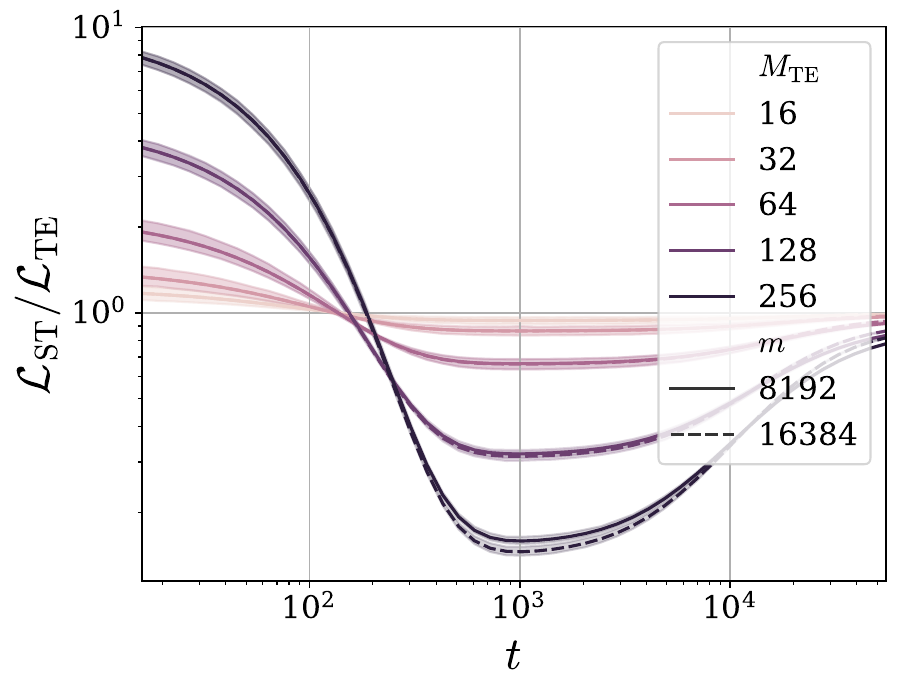}}
    \caption{Abalation of weak-to-strong generalization in 2-layer ReLU networks. We use same setting as \Cref{fig:exp:2layer-full} and compare the results with smaller student size $\MS$. }
    \label{fig:exp:2layer-ablation}
\end{figure}

However, due to GPU memory limit, we are not able to increase $\MS$. Thus we conduct ablation study with smaller $\MS$ to investigate the impact of $\MS$. In \Cref{fig:exp:2layer-ablation}, we compare the loss ratio curve between $\MS=8192$ and $\MS=16384$. We observe larger gap between the two when teacher loss $\TL$ is smaller, i.e., for small $d$ and large $\MT$. Therefore, it is not hard to believe that $\SL=\tilde{\Theta}(\TL^2)$ for this setting as well. We leave it as future work.

\subsection{Linear ReLU Network Experiments}
\begin{figure}[h]
    \centering
    \subfigure[$\SL/\TL^2$ v.s. $t$ at $k=1$]{\includegraphics[width=0.33\linewidth]{figures/delta=0.5_alpha=1_m=-1_gf.pdf}}
    \subfigure[$\SL/\TL^2$ v.s. $t$ at $k=10$]{\includegraphics[width=0.33\linewidth]{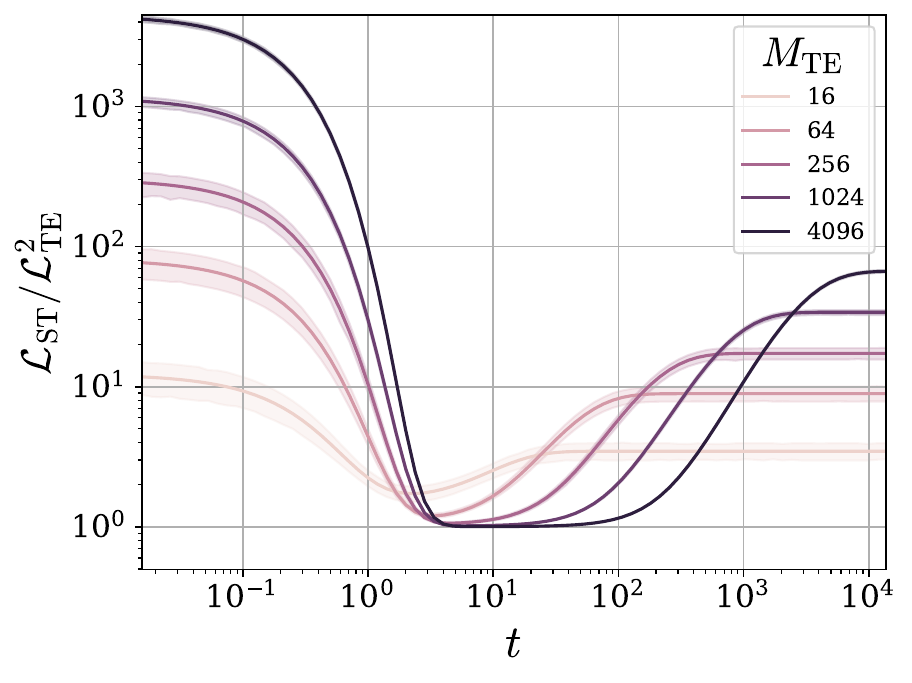}}
    \subfigure[$\SL/\TL^2$ v.s. $t$ at $k=100$]{\includegraphics[width=0.33\linewidth]{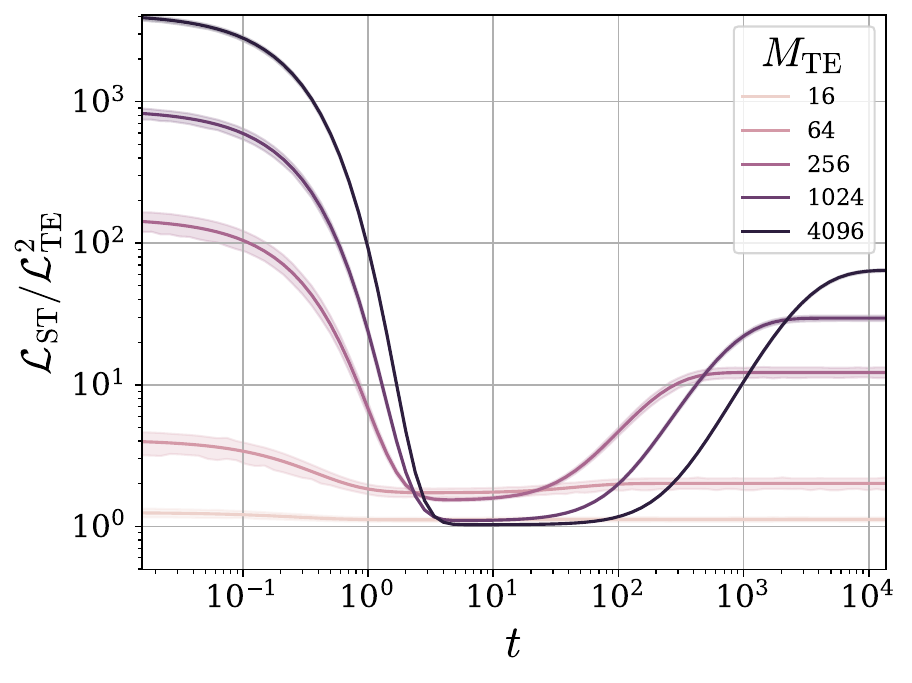}}
    \caption{Weak-to-Strong generalization happens in antitropic random linear feature networks (\Cref{ass:linearnetwork}). 
 Here we used an input distribution as in \Cref{thrm:main:diagfeatcov},  with $k=1, 10, 100$ and a target function $f^*=\langle \beta, \vx\rangle$ where $\beta$ is a unit norm vector with non-zero values for first $k$ coordinates. The figures the ratio between the student loss $\SL$ and squared teacher loss $\TL^2$, with varying teacher size $\MT$, and where the dimensionality $d=\MT^{3/2}+k$ as set as in the scaling of \Cref{thrm:main:diagfeatcov}, as a function of the gradient flow  time $t$. With proper early stopping time $\SL/\TL^2$ converges to approximately $1$ as $\MT$ grows, confirming that for large $\MT$ we have $\SL \propto \TL^2$ as in \Cref{thrm:main:diagfeatcov}.}
    \label{fig:exp:linear-full}
\end{figure}

\Cref{fig:exp:linear-full} shows full experimental results for \Cref{ass:linearnetwork}. It shows the same conclusions as \Cref{fig:exp:linear-main}, but with $k=10$ and $100$, providing additional verification of \Cref{thrm:main:diagfeatcov}.

\end{document}